\documentclass[hidelinks,onefignum,onetabnum]{siamart250211}
\usepackage{tikz}
\usepackage{subcaption}
\usepackage{graphicx}
\usepackage{booktabs}
\usepackage{multirow}
\usepackage{tcolorbox}
\usepackage{epstopdf}
\usepackage{adjustbox}
\usepackage[dvipsnames]{xcolor}

\usepackage{hyperref}
\hypersetup{
	colorlinks=true,
	linkcolor=blue,
	citecolor=blue,
	filecolor=magenta,
	urlcolor=blue,
}

\usepackage{lipsum}
\usepackage{amsfonts}
\usepackage{graphicx}
\usepackage{epstopdf}
\usepackage{algorithmic}
\ifpdf
  \DeclareGraphicsExtensions{.eps,.pdf,.png,.jpg}
\else
  \DeclareGraphicsExtensions{.eps}
\fi

\newsiamremark{remark}{Remark}
\newsiamremark{hypothesis}{Hypothesis}
\crefname{hypothesis}{Hypothesis}{Hypotheses}
\newsiamthm{claim}{Claim}
\newsiamremark{fact}{Fact}
\crefname{fact}{Fact}{Facts}

\headers{Total Normal Curvature Regularization}{Tianle Lu, Ke Chen and Yuping Duan}

\title{Total Normal Curvature Regularization and its Minimization for Surface and Image Smoothing\thanks{Submitted to the editors DATE.
}}

\author{Tianle Lu\thanks{School of Mathematical Sciences, Beijing Normal University 
  (\email{tianle0357@mail.bnu.edu.cn}).}
  \and Ke Chen\thanks{Department of Mathematics and Statistics, University of
Strathclyde (\email{k.chen@strath.ac.uk}).}
\and Yuping Duan\thanks{School of Mathematical Sciences, Beijing Normal University 
  (\email{doveduan@gmail.com}).}
}

\usepackage{amsopn}

\ifpdf
\hypersetup{
  pdftitle={Total Normal Curvature Regularization},
  pdfauthor={Tianle Lu, Ke Chen and Yuping Duan}
}
\fi

\begin{document}

\maketitle

\begin{abstract}
We introduce a novel formulation for curvature regularization by penalizing normal curvatures from multiple directions. This total normal curvature regularization is capable of producing solutions with sharp edges and precise isotropic properties. To tackle the resulting high-order nonlinear optimization problem, we reformulate it as the task of finding the steady-state solution of a time-dependent partial differential equation (PDE) system. Time discretization is achieved through operator splitting, where each subproblem at the fractional steps either has a closed-form solution or can be efficiently solved using advanced algorithms. Our method circumvents the need for complex parameter tuning and demonstrates robustness to parameter choices. The efficiency and effectiveness of our approach have been rigorously validated in the context of surface and image smoothing problems.
\end{abstract}



\section{Introduction}
Curvature plays a crucial role in image and surface smoothing \cite{shen2003euler, tai2011fast}. In image processing, it facilitates edge detection \cite{bredies2015convex, ciomaga2017image}, denoising \cite{chambolle2019total,he2021penalty}, smoothing \cite{deng2019new, liu2024fast}, and the enhancement of segmentation algorithms \cite{duan2014two, bae2017augmented,liu2023elastica}. Areas of high curvature typically indicate edges or feature points, enabling the removal of noise while preserving essential features. For surface smoothing, curvature aids in optimizing 3D model surfaces by eliminating sharp or discontinuous areas, leading to more natural appearances. In computer graphics, curvature supports geometric processing and mesh simplification, enhancing rendering efficiency and visual quality \cite{bensaid2024multi, chambolle2024minimizing}. Overall, curvature offers essential geometric insights for advanced image analysis and 3D model optimization.

Among curvature-based models, Euler’s elastica is a classical approach that regularizes both the length and curvature of level lines. For any image $v(x_{1},x_{2})$ defined on $\Omega \in \mathbb{R}^{2}$, its energy functional,
\begin{equation*}
    \int_{\Omega} (1+ a \kappa^{2})\vert \nabla v \vert \mathrm{d} \mathbf{x},
\end{equation*}
where $a \geq 0$ is the regularization parameter and $\vert \cdot \vert$ denotes the Euclidean norm of the gradient vector. 
This model effectively promotes smooth, continuous edges while preserving fine structures, and has been widely applied to denoising \cite{bae2010graph, liu2021color}, inpainting \cite{masnou1998level, shen2003euler, yashtini2016fast}, and segmentation \cite{tai2011fast, zhu2013image}. 
However, minimizing this energy is difficult due to its nonlinearity and nonconvexity. 
The existing numerical optimization methods for solving the curvature related energy minimization problems can be roughly divided into two categories: gradient descend and operator splitting method. 
Early approaches employed gradient flow methods involving complex PDEs \cite{ballester2001filling, shen2003euler}, which are computationally expensive. 
To alleviate these difficulties, various numerical techniques have been developed, including graph-cut methods \cite{bae2010graph}, multi-grid methods \cite{brito2008multigrid}, convex  relaxation \cite{bredies2015convex, chambolle2019total}.
Another major category of methods relies on operator splitting techniques, which aim to simplify the original nonconvex and nonlinear optimization problem by decomposing it into several more manageable subproblems. 
The key idea is to decouple the complex interactions between terms in the energy functional, allowing each subproblem to be solved efficiently. 
Several strategies exist for this purpose. 
One approach applies splitting schemes \cite{glowinski1989augmented} directly to the associated PDE system, such as Lie-splitting or fractional step methods \cite{deng2019new}. 
Alternatively, auxiliary variables can be introduced to reformulate the problem as a constrained minimization, where iterative updates are performed on each variable separately; typical examples include the augmented Lagrangian method (ALM) and alternating direction methods \cite{tai2009augmented, tai2011fast, duan2013fast, liu2023elastica}. 
Notably, many of these algorithms, originally developed for Euler’s elastica, have also been successfully adapted to the minimization of other curvature-based regularization models, such as those involving mean curvature, Gaussian curvature, and total curvature energies.

Mean curvature is defined as the arithmetic mean of the two principal curvatures at a given point on a surface \cite{do2016differential}. As an extrinsic geometric quantity, mean curvature depends on the specific embedding of the surface in space. It provides information about the local bending trend of the surface at that point. Mean curvature regularization has been used for image smoothing \cite{zhu2012image,yang2012homotopy, myllykoski2015new, zhong2021image,gong2018mean} and surface smoothing \cite{ohtake2000polyhedral,shen2004fuzzy}. 
Gaussian curvature is the product of the two principal curvatures at a given point on a surface. As an intrinsic geometric quantity, the Gaussian curvature does not depend on the specific embedding of the surface in space but is determined by the intrinsic geometric properties of the surface. It provides information on the bending nature of the surface at that point and can reveal local shape characteristics of the surface. For example, positive Gaussian curvature indicates that the surface is elliptic at that point, negative Gaussian curvature indicates that the surface is hyperbolic, and zero Gaussian curvature indicates that the surface is parabolic. Gaussian curvature has been used for image smoothing \cite{brito2016image, liu2022operator, zhang2023fast} and surface smoothing \cite{zhao2006triangular, elsey2009analogue}. 
However, both mean curvature and Gaussian curvature are scalar quantities and therefore do not contain detailed information about how the surface bends in specific directions.

While mean and Gaussian curvatures offer valuable scalar summaries of surface geometry \cite{zhu2012image, brito2016image}, they originate from a more fundamental concept: normal curvature \cite{do2016differential}. At each point on a surface, normal curvature characterizes the bending of the surface in a specific tangent direction. Mean curvature is computed as the average of normal curvatures across all directions, whereas Gaussian curvature is the product of the maximal and minimal normal curvatures. Thus, normal curvature serves as a more primitive, direction-sensitive quantity that underpins both mean and Gaussian curvatures. It is particularly useful for analyzing the local geometric properties of surfaces, especially in applications that demand precise control and optimization of surface shapes \cite{gong2017curvature, zhong2020minimizing}. By examining normal curvature across various directions, one can gain deeper insights into the local bending behavior of the surface. Building on this concept, we introduce the total normal curvature (TNC), defined as the integral of the absolute values of normal curvatures over all possible directions.

In this work, we propose a variational model for image and surface smoothing based on total normal curvature. To address the nonlinearities inherent in the model, we introduce two auxiliary fields—one matrix-valued and one vector-valued—subject to appropriate constraints. By enforcing these constraints through indicator functionals, we derive the first-order optimality conditions for the augmented problem. These conditions are interpreted as the steady-state of an associated time-dependent PDE system. We then employ an operator-splitting discretization in time to ensure that each substep of our scheme either has a closed-form solution or can be solved with high efficiency. We validate the proposed method on surface-smoothing and image-denoising benchmarks, which demonstrate its robust optimization of the normal-curvature-based model with minimal sensitivity to parameter choices.

The remainder of this paper is organized as follows. 
In \cref{sec:main}, we introduce the necessary notations and definitions. 
The reformulation of the total normal curvature minimization model is described in \cref{sec: model}.
We describe the operator-splitting method based on the Lie scheme in \cref{sec:alg}.
The proposed operator-splitting method and solvers for each subproblem are presented in \cref{sec: sub}.
The proposed method is spatially discretized in \cref{sec:dis}.
In \cref{sec:experiments}, we apply our methodology to the solution of surface and image smoothing problems.
Finally, we summarize and conclude our paper in \cref{sec:conclusions}.

\section{Notations and Definitions}
\label{sec:main}
Let $f: \Omega \rightarrow \mathbb{R}$ be an image defined on the domain $\Omega \subset \mathbb{R}^2$, and $v: \Omega \rightarrow \mathbb{R}$ be the desired clean image. Our goal is to estimate a smooth surface $\mathcal{S}=(x,y,v(x,y))$ from the given image. 
Let $\phi(x,y,z) = v(x,y) - z$ be the level set function, where $v(x,y)$ represents the surface height. The mean curvature $\kappa_{M}$ over the zero-level set $z=v(x,y)$ is given by (cf.  \cite{zhu2012image})
\begin{equation}
\label{eq:mean_curvature}
\begin{aligned}
\kappa_{M}&=\frac12\nabla\cdot\left(\frac{\nabla\phi}{|\nabla\phi|}\right) \\
&=\frac{1}{2}\nabla\cdot\left(\frac{\nabla v}{\sqrt{1+|\nabla v|^{2}}}\right)\\
&=\frac{\left(1+v_{x}^{2}\right) v_{y y}-2 v_{x} v_{y} v_{x y}+\left(1+v_{y}^{2}\right) v_{x x}}{2\left(1+v_{x}^{2}+v_{y}^{2}\right)^{3/2}},
\end{aligned}
\end{equation}
where $\nabla$ and $\nabla\cdot$ denote the gradient and divergence operators, respectively. 
The Gaussian curvature $\kappa_{G}$ of the surface implicitly defined by the zero level set of $\phi$ is given by (cf. \cite{brito2016image}): 
\begin{equation}
\begin{aligned}
        \kappa_{G} &= \frac{\nabla \phi \mathbf{H}^{*}(\phi)\nabla \phi^{\top}}{\vert \nabla \phi \vert^{4}}\\&= \frac{\mathrm{det}(\mathbf{H}(v))}{(\vert\nabla v\vert^{2} +1)^{2}}\\&= \frac{v_{xx}v_{yy}-v_{xy}v_{yx}}{(v_{x}^{2}+v_{y}^{2}+1)^{2}},
\end{aligned}
\label{eq:gaussian_curvature}
\end{equation}
where $\mathbf{H}(\phi)$ is the Hessian matrix and $\mathbf{H}^{*}(\phi)$ is the adjoint matrix. 

The tangent plane $T_p\mathcal{S}$ at a point $p = (x_0, y_0, v(x_0, y_0))$ on the surface $\mathcal{S}$ is the vector space of all tangent vectors at $p$, defined as the set of velocity vectors $\gamma'(0)$ of smooth curves $\gamma(t)$ on $\mathcal{S}$ satisfying $\gamma(0) = p$. Parameterizing $T_p \mathcal{S}$ by an angle $\theta$, the normal curvature $\kappa_n(\theta)$ along the tangent direction $\mathbf{t} = (\cos\theta, \sin\theta)^\top$ is defined as the ratio of the second fundamental form $\mathbf{II}$ to the first fundamental form $\mathbf{I}$:
\begin{equation}
\begin{aligned}
        \kappa_n(\theta) 
        &=\frac{\mathbf{II}(\mathbf{t}, \mathbf{t})}{\mathbf{I}(\mathbf{t}, \mathbf{t})}\\
        &=\frac{\mathbf{t}^{\top} \mathbf{H} \mathbf{t}}{\sqrt{1 + \vert \nabla v \vert^{2}} \cdot \left( 1 + (\nabla v : \mathbf{t})^2 \right)}   \\
        & = \frac{v_{xx} \cos^2\theta + 2v_{xy} \cos\theta \sin\theta + v_{yy} \sin^2\theta}{\sqrt{1 + v_x^2 + v_y^2} \cdot \left[1 + (v_x \cos\theta + v_y \sin\theta)^2 \right]},
\end{aligned}
\label{eq:normal_curvature}
\end{equation}
where $\nabla v : \mathbf{t} = v_{1}t_{1}+ v_{2}t_{2}$. A more detailed definition of normal curvature is presented in Appendix \ref{appendix:A}.
As $\mathbf{t}$ varies over all unit vectors in $T_{p} \mathcal{S}$, the continuous function $\kappa_n(\theta)$ attains its maximum and minimum values
\[\kappa_1 = \max_{\theta \in [0,2\pi)} \kappa_n(\theta), \quad \kappa_2 = \min_{\theta \in [0,2\pi)} \kappa_n(\theta),\] 
known as the principal curvatures. The mean curvature, defined as the average of the principal curvatures, captures the surface’s overall bending intensity, whereas the Gaussian curvature, given by their product, characterizes its intrinsic bending properties.

The comparative analysis of various discrete Total Variation (TV) definitions such as anisotropic, isotropic, and upwind TV, provided in \cite{condat2017discrete}, examined how different gradient norm definitions influence numerical accuracy in image processing tasks. Thus, we investigate the second-order geometric priors via curvature regularization to improve the precision of curvature definitions.  
The central question we seek to answer is: At a given point $p$ on a regular surface $\mathcal{S}$, if we consider the normal curvatures $\kappa_n(\theta)$ along all unit tangent directions $\mathbf{t} \in T_p\mathcal{S} $, can we obtain a more accurate local geometric characterization than relying on the principle curvatures? Since curvature computation involves the discretization of second-order differential operators, a precise definition that rigorously accounts for directional variations is essential for ensuring the numerical accuracy and stability of discrete curvature operators.

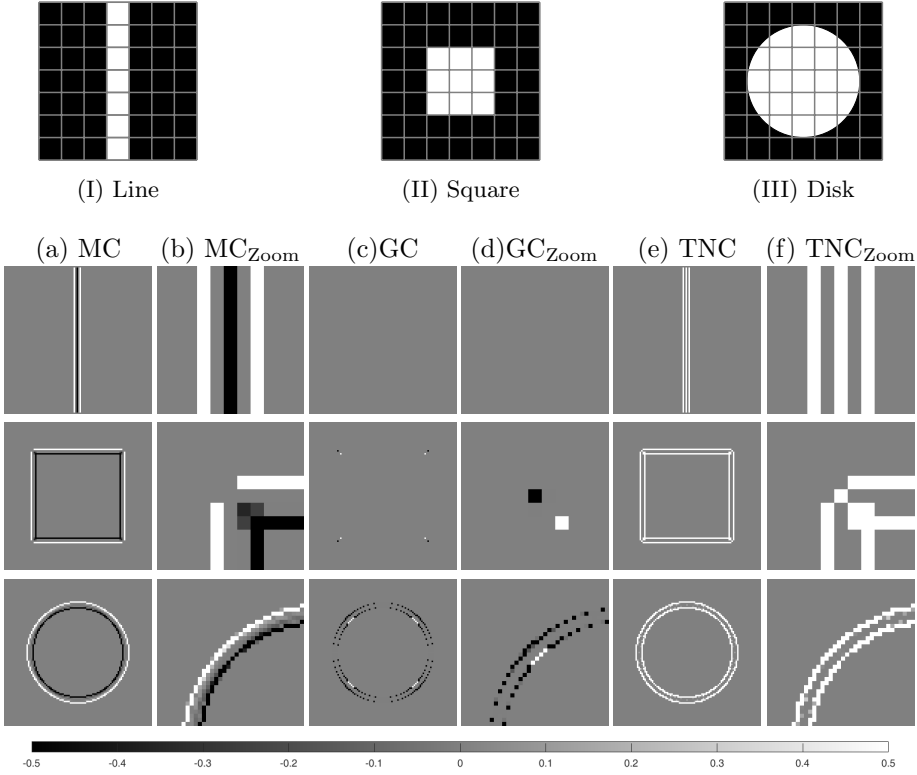
\begin{figure}[t]
  \centering
  \begin{subfigure}[b]{0.3\textwidth}
    \centering
    \begin{tikzpicture}[scale=0.3]
      \draw[step=1cm,gray,line width=0.2mm] (0,0) grid (7,7);
      \fill[black] (0,7) rectangle (3,0);
      \fill[black] (4,7) rectangle (7,0);
      \draw[step=1cm,gray,line width=0.2mm] (0,0) grid (7,7);
    \end{tikzpicture}
    \caption*{(I) Line}
    \label{sub:line}
  \end{subfigure}
  \hfill
  \begin{subfigure}[b]{0.3\textwidth}
    \centering
    \begin{tikzpicture}[scale=0.3]
      \draw[step=1cm,gray,line width=0.2mm] (0,0) grid (7,7);
      \fill[black] (0,7) rectangle (7,0);
      \fill[white] (2,2) rectangle (5,5);
      \draw[step=1cm,gray,line width=0.2mm] (0,0) grid (7,7);
    \end{tikzpicture}
    \caption*{(II) Square}
    \label{sub:corner}
  \end{subfigure}
  \hfill
  \begin{subfigure}[b]{0.3\textwidth}
    \centering
    \begin{tikzpicture}[scale=0.3]
      \draw[step=1cm,gray,line width=0.2mm] (0,0) grid (7,7);
      \fill[black] (0,7) rectangle (7,0);
      \fill[white] (3.5,3.5) circle (2.5);
      \draw[black] (3.5,3.5) circle (2.5);
      \draw[step=1cm,gray,line width=0.2mm] (0,0) grid (7,7);
    \end{tikzpicture}
    \caption*{(III) Disk} 
    \label{sub:disk}
  \end{subfigure}
	\begin{tabular}{c@{\hspace{2pt}}c@{\hspace{2pt}}c@{\hspace{2pt}}c@{\hspace{2pt}}c@{\hspace{2pt}}c}
	(a) $\mathrm{MC}$ & (b) 
    $\mathrm{MC}_{\mathrm{Zoom}}$
     & (c)$\mathrm{GC}$ & (d)$\mathrm{GC}_{\mathrm{Zoom}}$  & (e) $\mathrm{TNC}$ & (f) $\mathrm{TNC}_{\mathrm{Zoom}}$\\          \includegraphics[width=0.15\textwidth]   {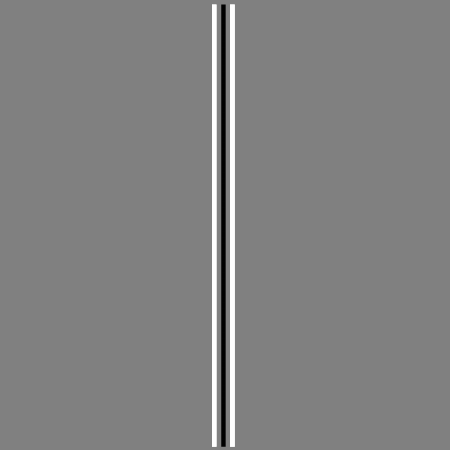}&
        \includegraphics[width=0.15\textwidth]{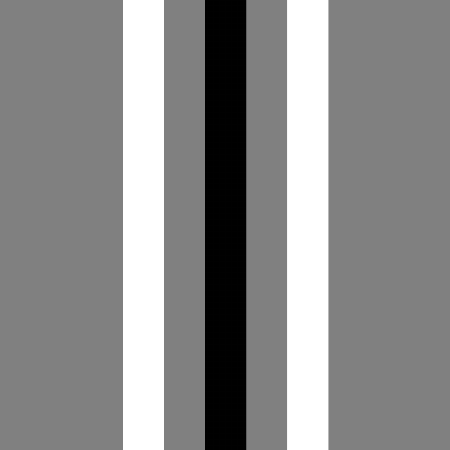}&
        \includegraphics[width=0.15\textwidth]{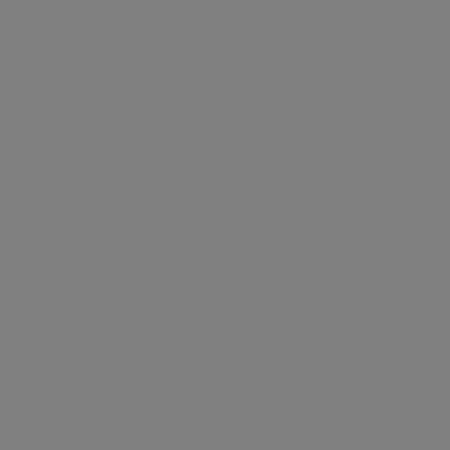}&
        \includegraphics[width=0.15\textwidth]{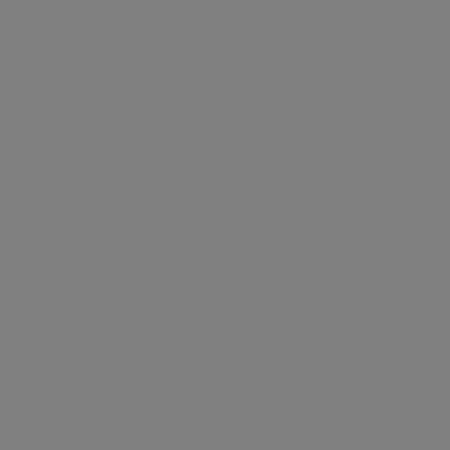}&
        \includegraphics[width=0.15\textwidth]{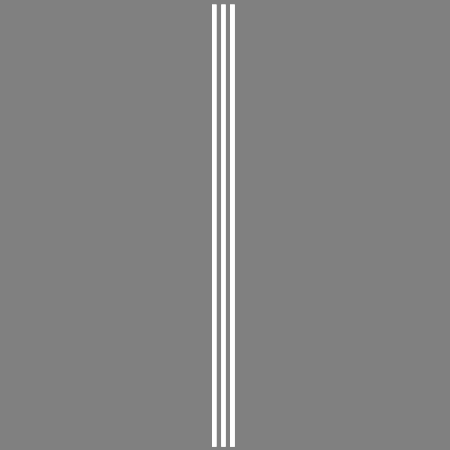}&
        \includegraphics[width=0.15\textwidth]{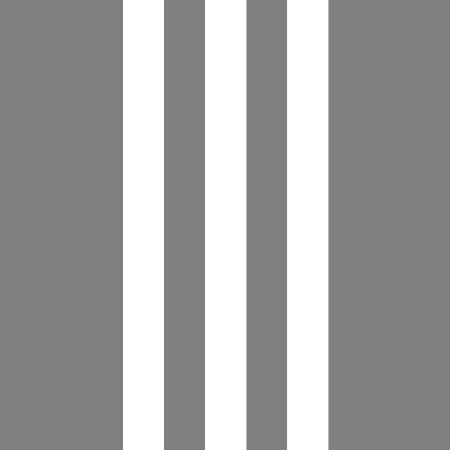}
        \\
        \includegraphics[width=0.15\textwidth]   {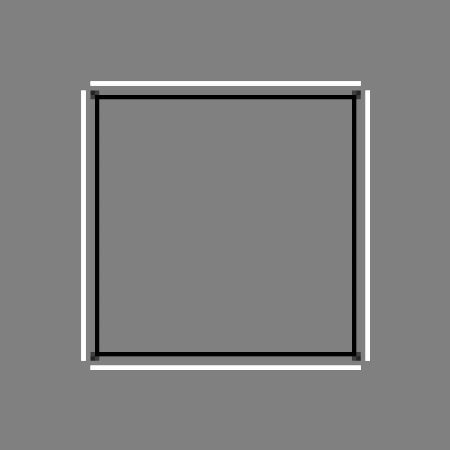}&
        \includegraphics[width=0.15\textwidth]{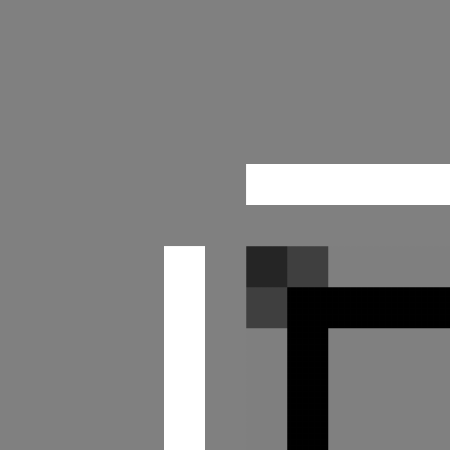}&
        \includegraphics[width=0.15\textwidth]{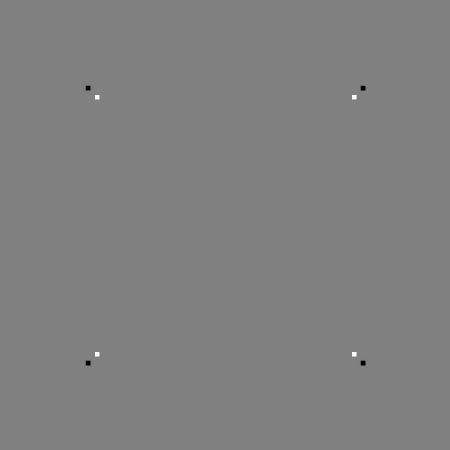}&
        \includegraphics[width=0.15\textwidth]{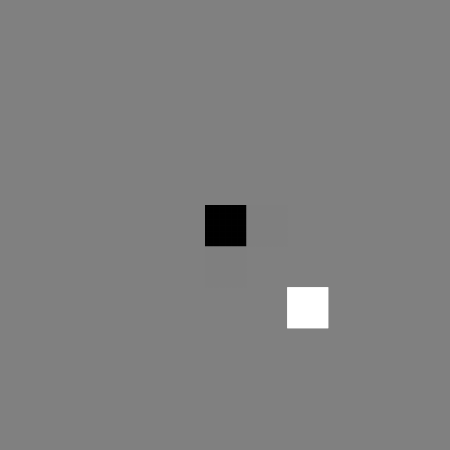}&
        \includegraphics[width=0.15\textwidth]{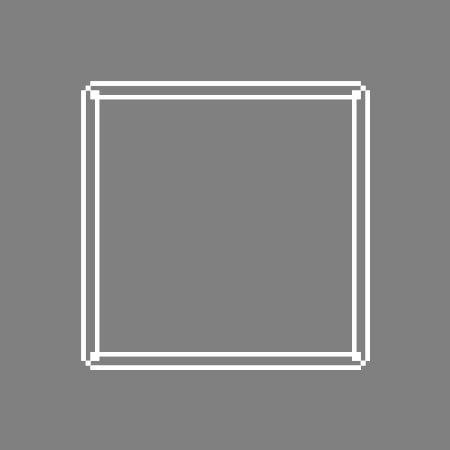}&
        \includegraphics[width=0.15\textwidth]{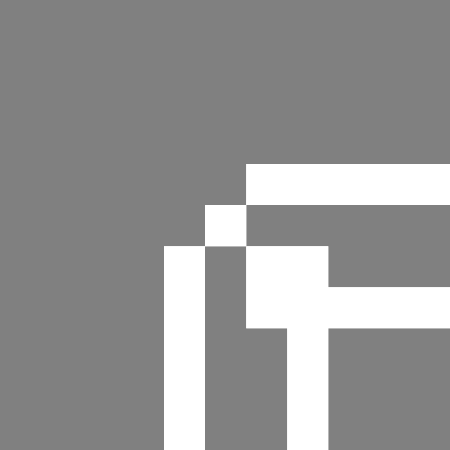}\\
        \includegraphics[width=0.15\textwidth]   {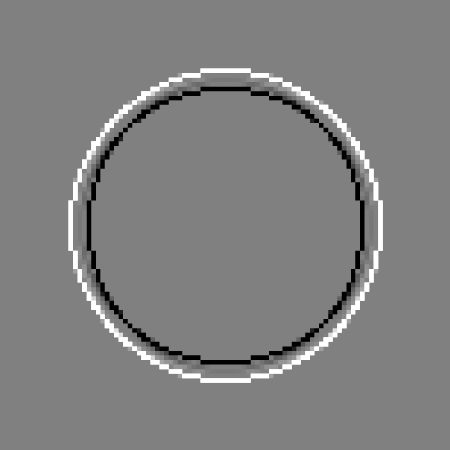}&
        \includegraphics[width=0.15\textwidth]{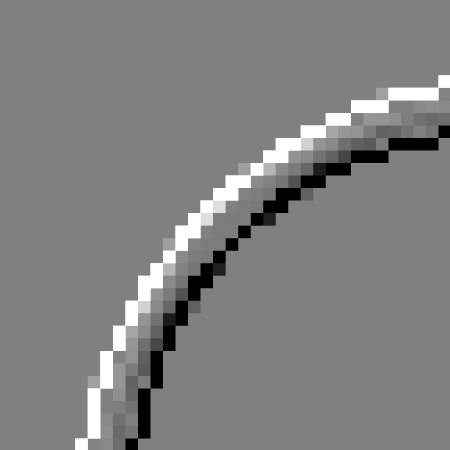}&
        \includegraphics[width=0.15\textwidth]{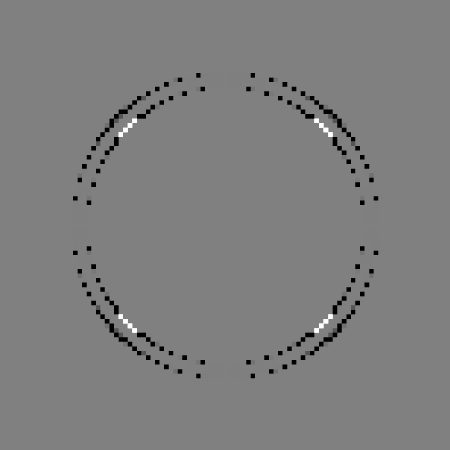}&
        \includegraphics[width=0.15\textwidth]{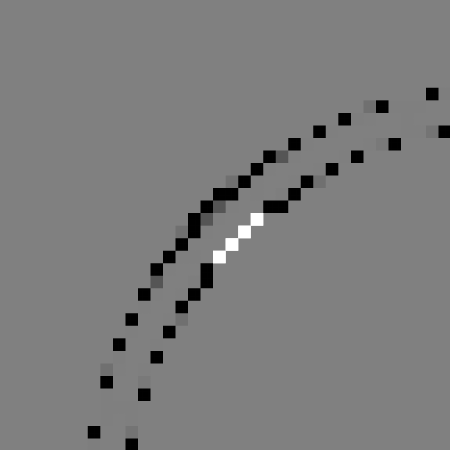}&
        \includegraphics[width=0.15\textwidth]{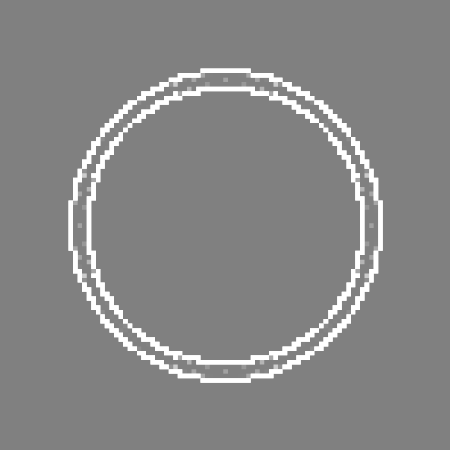}&
        \includegraphics[width=0.15\textwidth]{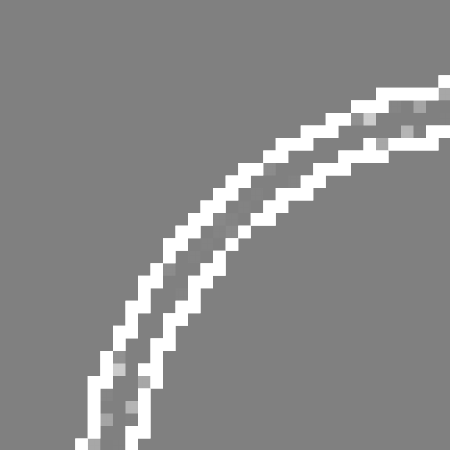}
        \\
        \multicolumn{6}{c}{\includegraphics[width=0.9\textwidth]{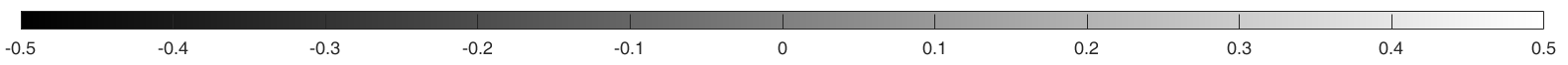}}
        \end{tabular}
   \caption{Numerical comparison among different curvatures on three typical image patterns, where (I)-(III) are test patterns; (a)-(f) are different curvatures and their zoomed-in regions.}
   \label{fig:dif_curv_patterns}
\end{figure}

In this work, we define the total normal curvature (TNC), based on the normal curvature given in (\ref{eq:normal_curvature}), as follows
\begin{equation}
\label{eq:TNC}
\begin{aligned}
\kappa_{T}&=\int_{0}^{2\pi}\vert\kappa_{n}(\theta)\vert \mathrm{d}\theta\\
&=\int_{0}^{2\pi} \frac{\vert \mathbf{t}^{\top} \mathbf{H} \mathbf{t} \vert }{\sqrt{1 + \vert \nabla v \vert^{2}} \cdot \left( 1 + (\nabla v : \mathbf{t})^2 \right)} \mathrm{d}\theta  \\
& = \int_{0}^{2\pi}\frac{\vert v_{xx} \cos^2\theta + 2v_{xy} \cos\theta \sin\theta + v_{yy} \sin^2\theta \vert}{\sqrt{1 + v_x^2 + v_y^2} \cdot \left[1 + (v_x \cos\theta + v_y \sin\theta)^2 \right]}\mathrm{d}\theta.
\end{aligned}
\end{equation}
In \cref{fig:dif_curv_patterns}, we evaluate three curvature measures: Mean Curvature (MC, \eqref{eq:mean_curvature}), Gaussian Curvature (GC, \eqref{eq:gaussian_curvature}), and Total Normal Curvature (TNC, \eqref{eq:TNC}), on three fundamental geometric patterns: Line, Square, and Disk. We assume an image size of $N \times N$ with $N$ sufficiently large to minimize boundary effects. 
For the line pattern, MC and TNC yield nearly identical contours, whereas GC returns zero curvature ($\kappa_G = 0$) due to the conditions $u_{yy} = 0$ and $u_{xy} = u_{yx} = 0$. In the square pattern, the curvature measures exhibit distinct behaviors in corner characterization: MC shows missing values at the corners; GC, though theoretically zero, displays numerical artifacts near the corners; and TNC consistently preserves the corner structure. Similarly, in the disk pattern, GC is affected by numerous error points, while TNC demonstrates significantly fewer artifacts in the region between the two rings.

\begin{remark}
    Whereas a continuous surface exhibits infinitely many normal curvatures as direction varies, we discretize the orientation space using a 3×3 local window.
    This standard approach uniformly samples the $[0, 2\pi)$ interval into eight distinct directions, defined by $\theta \in \{0, \pi/4,$$ \dots, 7\pi/4\}$. 
    These sampled directions effectively characterize the local geometry around the center pixel.
\end{remark}

\section{Total Normal Curvature Model}
\label{sec: model}
As explained in the previous section and
motivated by the good results of the TNC,
we propose the following TNC regularization model for surface and image smoothing problems
\begin{equation}
    \min_{v\in\mathcal{H}^{2}(\Omega)} \frac{\alpha}{2}\int_{\Omega} \int_{0}^{2\pi} \frac{\vert \mathbf{t}^{\top} \mathbf{H} \mathbf{t} \vert }{\sqrt{1 + \vert \nabla v \vert^{2}} \cdot \left( 1 + (\nabla v : \mathbf{t})^2 \right)}  \, \mathrm{d} \theta \, \mathrm{d} \mathbf{s} + \beta \int_{\Omega} \vert \nabla v\vert \mathrm{d}\mathbf{x} + \frac{\gamma}{2}\int_{\Omega} \vert f-v \vert ^{2} \mathrm{d} \mathbf{x},
    \label{model1}
\end{equation}
where $\alpha > 0$, $\beta \geq 0$ and $\gamma>0$ are weighting parameters balancing these terms, $\mathbf{t} = (\cos\theta, \sin \theta)^{\top}$ is a tangent vector and $\mathcal{H}^{2}(\Omega)$ is the Sobolev space. Here $\mathrm{d}\mathbf{s}$ denotes the surface area element, and $\mathrm{d}\mathbf{x}= \mathrm{d} x_{1}\mathrm{d}x_{2}$ represents the planar area element in $\Omega$ and $\mathbf{H}$ denotes the Hessian of $v$, given by:
\begin{equation*}
\mathbf{H}=\begin{pmatrix}
\frac{\partial^{2}v}{\partial x_{1}^{2}} & \frac{\partial^{2}v}{\partial x_{1}\partial x_{2}} \\
\frac{\partial^{2}v}{\partial x_{1}\partial x_{2}} & \frac{\partial^{2}v}{\partial x_{2}^{2}}
\end{pmatrix}.
\end{equation*}
Observe that when $\alpha = 0$ in the curvature regularization, the proposed model reduces to the classical TV model, which can remove noise and preserve sharp jumps, but suffers from staircase effects. For $\alpha>0$, the curvature regularization term utilizes its edge-preserving and corner-preserving properties, which is helpful to handle the drawbacks of classical TV regularization. 

By substituting the surface area element $\mathrm{d}\mathbf{s}$ by $\sqrt{1+ \vert \nabla v\vert^{2}} \mathrm{d}\mathbf{x}$, we obtain the following equivalent formulation:
\begin{equation}
\label{org_problem}
    \min_{v\in\mathcal{H}^{2}(\Omega)} \frac{\alpha}{2}\int_{\Omega} \int_{0}^{2\pi} \frac{\vert \mathbf{t}^{\top} \mathbf{H} \mathbf{t} \vert }{1 + (\nabla v : \mathbf{t})^2 }  \, \mathrm{d} \theta \, \mathrm{d} \mathbf{x} + \beta \int_{\Omega} \vert \nabla v\vert \mathrm{d}\mathbf{x} + \frac{\gamma}{2}\int_{\Omega} \vert f-v \vert ^{2} \mathrm{d} \mathbf{x}.
\end{equation}
The minimization problem (\ref{org_problem}) is highly nonlinear and non-smooth, primarily due to the total absolute normal curvature term. Thus, we introduce two auxiliary variables, i.e., a vector-valued function \(\mathbf{q}\) and a matrix-valued function \(\mathbf{G}\), to decouple the nonlinearities from the differential operators, where \(\mathbf{q}\) and \(\mathbf{G}\) are defined as follows:
\[
\mathbf{q} = \nabla v = \begin{bmatrix}
q_1 \\
q_2
\end{bmatrix} \in (\mathcal{H}^1(\Omega))^2, \quad \text{and} \quad \mathbf{G} = \nabla \mathbf{q} = \begin{pmatrix}
G_{11} & G_{12} \\
G_{21} & G_{22}
\end{pmatrix} \in (\mathcal{L}^2(\Omega))^{2 \times 2}.
\]
To further relax these constraints, we employ indicator functionals and define the sets \(\Sigma\) and \(S\) as follows:
\[
\Sigma = \left\{ \mathbf{q} \mid \mathbf{q} \in (\mathcal{L}^2(\Omega))^2, \exists v \in \mathcal{H}^1(\Omega) \text{ such that } \mathbf{q} = \nabla v, \int_{\Omega} v \, \mathrm{d}\mathbf{x} = \int_{\Omega} f \, \mathrm{d}\mathbf{x} \right\},
\]
and
\[
S = \left\{ (\mathbf{q}, \mathbf{G}) \mid (\mathbf{q}, \mathbf{G}) \in (\mathcal{H}^1(\Omega))^2 \times (\mathcal{L}^2(\Omega))^{2 \times 2} \text{ such that } \mathbf{G} = \nabla \mathbf{q} \right\}.
\]
By defining these sets, we effectively incorporate the constraints into the optimization framework, allowing for a more tractable approach to solving the original problem.
Based on the above notation, the problem (\ref{org_problem}) can be reformulated as the following equivalent unconstrained optimization problem for
$(\mathbf{q}, \mathbf{G})$, where its solution corresponds to $v$ for solving (\ref{org_problem}):
\begin{multline}
    \min_{\substack{ \mathbf{q} \in (\mathcal{H}^1(\Omega))^2, \\ \mathbf{G} \in (\mathcal{L}^2(\Omega))^{2 \times 2}}} J(\mathbf{q}, \mathbf{G}) = 
    \frac{\alpha}{2}\int_{\Omega} \int_{0}^{2\pi} \frac{\vert\mathbf{t}^{\top} \mathbf{G} \mathbf{t}\vert}{1 + (\mathbf{q} : \mathbf{t})^2}  \, \mathrm{d} \theta \, \mathrm{d} \mathbf{x} 
    + \beta \int_{\Omega} \vert \mathbf{q}\vert  \mathrm{d}\mathbf{x} \\ 
    + \frac{\gamma}{2}\int_{\Omega} \vert f - v_{\mathbf{q}} \vert^{2} \mathrm{d} \mathbf{x} 
    + I_{\Sigma}(\mathbf{q}) 
    + I_{S}(\mathbf{q}, \mathbf{G}),
    \label{model3}
\end{multline}
with $I_{\Sigma}$ and $I_{S}$ being the indicator functionals defined by
\begin{equation*}
I_{\Sigma}(\mathbf{q}) = \begin{cases} 
0 & \text{if } \mathbf{q} \in \Sigma, \\ 
+\infty & \text{otherwise,} 
\end{cases} \quad \mbox{and}\quad
I_{S}(\mathbf{q}, \mathbf{G}) = \begin{cases} 
0 & \text{if } (\mathbf{q}, \mathbf{G}) \in S, \\ 
+\infty & \text{otherwise.} 
\end{cases}
\end{equation*}
Then $v_{\mathbf{q}}$ becomes the solution of the following system
\begin{equation}
\begin{cases}
\nabla^2 v_{\mathbf{q}} = \nabla \cdot \mathbf{q} & \text{in } \Omega, \\
 (\nabla v_{\mathbf{q}})\cdot \mathbf{n} =0  & \text{on}~\partial\Omega, \\
 \int_{\Omega} v_{\mathbf{q}} \, \mathrm{d}\mathbf{x} =  \int_{\Omega} f \, \mathrm{d}\mathbf{x},
\end{cases}
\label{boundcondition}
\end{equation}
where $\nabla^2$ denotes the Laplace operator and $\mathbf{n}$ is the unit outward normal vector on the boundary $\partial\Omega$. Although Neumann boundary conditions are used in (\ref{boundcondition}), it is readily applicable to periodic boundaries, which enable efficient solution to the subproblems via the fast Fourier transform and thus accelerate the numerical computation; see Section \ref{subsec:perbound}.

\begin{remark}
The choice of the admissible space $\mathcal{H}^2(\Omega)$ in \eqref{model1} was motivated primarily by mathematical convenience, as it ensures that the second-order derivative term in our model is well-defined. We acknowledge, however, that the lack of a rigorous theoretical analysis regarding its edge-preserving properties remains a limitation. Consider the one-dimensional case where the surface reduces to a planar curve $v(x)$. Using the arc length parameterization $\mathrm{d}s = \sqrt{1+(v'(x))^2}\mathrm{d}x$, the total absolute curvature satisfies:
\begin{equation*}
 \int |\kappa|\,\mathrm{d}s 
= \int \frac{|v''(x)|}{1 + (v'(x))^2}\,\mathrm{d}x 
= \int \left| \frac{\mathrm{d}}{\mathrm{d}x} \left( \arctan v'(x) \right) \right| \mathrm{d}x 
= \operatorname{TV}(\arctan v'(x)) \le \pi.  
\end{equation*}
It shows that the total curvature equals the total variation of the tangent angle $\theta(x) = \arctan v'(x)$. 
As a result, the total curvature across a sharp transition remains bounded by $\pi$, even as the transition width tends to zero, which reveals the intrinsic edge-preserving property of the curvature regularization. However, a rigorous analysis to the $BV(\Omega)$ setting for our model remains an open problem for future investigation.
\end{remark}

\section{Operator splitting algorithm}
\label{sec:alg}
In this section, we propose an operator-splitting method for solving the minimization problem (\ref{model3}). By decomposing coupled operators into a sequence of simpler, computationally tractable subproblems, it provides an effective algorithmic framework for complex systems and has been widely applied in nonlinear partial differential equations \cite{glowinski2019finite}, inverse problems \cite{glowinski2015penalization}, and various imaging models \cite{deng2019new, liu2022operator, duan2022fast}. 

\subsection{The optimality condition associated with (\ref{model3})}
Let us define the functionals $J_{1}$, $J_{2}$ and $J_{3}$ by
\begin{equation} \label{eq:split}
\left\{
\begin{aligned}
&J_{1}(\mathbf{q}, \mathbf{G}) = \frac{\alpha}{2}\int_{\Omega} \int_{0}^{2\pi} \frac{\vert\mathbf{t}^{\top} \mathbf{G} \mathbf{t}\vert }{ 1 + (\mathbf{q} : \mathbf{t})^2 }  \, \mathrm{d} \theta \, \mathrm{d} \mathbf{x}, \\
&J_{2}(\mathbf{q})  =\beta \int_{\Omega} \vert \mathbf{q}\vert \mathrm{d}\mathbf{x},\\
&J_{3}(\mathbf{q}) = \frac{\gamma}{2}\int_{\Omega} \vert f-v_{\mathbf{q}} \vert ^{2} \mathrm{d} \mathbf{x}.
\end{aligned}
\right.
\end{equation}
Suppose that $(\mathbf{p}, \mathbf{H})$ is a minimizer of the functional in (\ref{model3}). We then have $u =v_{\mathbf{p}}$ and
the following system of (necessary) optimality conditions holds:
\begin{equation}
\begin{cases}
D_{\mathbf{q}} J_1(\mathbf{p}, \mathbf{H}) +  \partial_{\mathbf{q}}J_{2}(\mathbf{p}) +D_{\mathbf{q}} J_3(\mathbf{p})  + \partial_{\mathbf{q}} I_S(\mathbf{p}, \mathbf{H}) + \partial_{\mathbf{q}} I_{\Sigma}(\mathbf{p}) \ni \mathbf{0}, \\
\partial_{\mathbf{G}} J_1(\mathbf{p}, \mathbf{H}) + \partial_{\mathbf{G}} I_S(\mathbf{p}, \mathbf{H}) \ni \mathbf{0},
\end{cases}
\label{optcondition}
\end{equation}
where $D_{\mathbf{q}}$ (resp., $\partial_{\mathbf{q}}$) denotes the partial derivative (resp., subdifferential) for differentiable (resp., non-smooth) functionals with respect to $\mathbf{q}$. The operator $\partial_{\mathbf{G}}$ is defined analogously.

We associate the optimality conditions in \eqref{optcondition} with the following initial value problem (dynamical flow):
\begin{equation}
\begin{cases}
\eta\frac{\partial\mathbf{p}}{\partial t} +D_{\mathbf{q}} J_1(\mathbf{p}, \mathbf{H}) +  \partial_{\mathbf{q}} J_2(\mathbf{p})  + D_{\mathbf{q}}J_{3}(\mathbf{p})+\partial_{\mathbf{q}} I_S(\mathbf{p}, \mathbf{H}) + \partial_{\mathbf{q}} I_{\Sigma}(\mathbf{p}) \ni \mathbf{0}, \\
\frac{\partial\mathbf{H}}{\partial t}+\partial_{\mathbf{G}} J_1(\mathbf{p}, \mathbf{H}) + \partial_{\mathbf{G}} I_S(\mathbf{p}, \mathbf{H}) \ni \mathbf{0},\\
(\mathbf{p}(0),\mathbf{H}(0) ) = (\mathbf{p}_{0},\mathbf{H}_{0}),
\end{cases}
\label{dflow1}
\end{equation}
where $\eta$ is a positive constant controlling the evolution speed of $\mathbf{p}$. 
Since the steady-state of \eqref{dflow1} satisfies the optimality condition \eqref{optcondition}, we compute it by applying an operator-splitting based time discretization method to \eqref{dflow1} (see section \ref{subsec: Alg}), with the choice of initial condition $(\mathbf p_0,\mathbf H_0)$ discussed in section \ref{subsec: Initial condition}.

\subsection{An operator-splitting method for the dynamical-flow system (\ref{dflow1})}
\label{subsec: Alg}
We employ an operator-splitting method based on the Lie scheme \cite{glowinski2017splitting, glowinski2017some} for the temporal discretization of the problem \eqref{dflow1}. Let $n$ denote the iteration index and $\tau > 0$ the time step size, such that $t^{n} = n\tau$.  Starting from an initial condition $(\mathbf{p}_{0}, \mathbf{H}_{0})$, the approximate solution $(\mathbf{p}^{n}, \mathbf{H}^{n})$ is updated at each time step through the following four fractional steps:

\noindent\underline{\emph{Initialization}}:
\begin{equation}
    (\mathbf{p}^{0},\mathbf{H}^{0} ) = (\mathbf{p}_{0},\mathbf{H}_{0}).
    \label{FS0}
\end{equation}
\noindent\underline{\emph{Fractional Step 1}}:

\begin{equation}
\begin{cases}
    \begin{cases}
    \eta \frac{\partial \mathbf{p}}{\partial t} + D_{\mathbf{q}} J_1(\mathbf{p}, \mathbf{H}) = \boldsymbol{0}, \\
    \frac{\partial \mathbf{H}}{\partial t} + \partial_{\mathbf{G}} J_1(\mathbf{p}, \mathbf{H}) \ni \boldsymbol{0},
    \end{cases}
    \text{ in } \Omega \times (t^n, t^{n+1}),\\
    (\mathbf{p}(t^{n}), \mathbf{H}(t^{n})) = (\mathbf{p}^{n}, \mathbf{H}^{n}),
\end{cases}
\label{FS1}
\end{equation}
and set
\begin{equation}
    (\mathbf{p}^{n+1/4}, \mathbf{H}^{n+1/4}) = (\mathbf{p}(t^{n+1}), \mathbf{H}(t^{n+1})).
\end{equation}

\noindent\underline{\emph{Fractional Step 2}}:
\begin{equation}
\begin{cases}
    \begin{cases}
    \eta \frac{\partial \mathbf{p}}{\partial t} + \partial _{\mathbf{q}} J_{2}(\mathbf{p})\ni \boldsymbol{0}, \\
    \frac{\partial \mathbf{H}}{\partial t} = \boldsymbol{0},
    \end{cases}
    \text{ in } \Omega \times (t^n, t^{n+1}), \\
    (\mathbf{p}(t^{n}), \mathbf{H}(t^{n})) = (\mathbf{p}^{n+1/4}, \mathbf{H}^{n+1/4}),
\end{cases}
\label{FS2}
\end{equation}
and set
\begin{equation}
\label{FS3.2}
    (\mathbf{p}^{n+2/4}, \mathbf{H}^{n+2/4}) = (\mathbf{p}(t^{n+1}), \mathbf{H}(t^{n+1})). 
\end{equation}

\noindent\underline{\emph{Fractional Step 3}}:
\begin{equation}
\begin{cases}
    \begin{cases}
    \eta \frac{\partial \mathbf{p}}{\partial t} + \partial _{\mathbf{q}} I_S(\mathbf{p},\mathbf{H})\ni \boldsymbol{0}, \\
    \frac{\partial \mathbf{H}}{\partial t}  +\partial _{\mathbf{G}} I_S(\mathbf{p},\mathbf{H})\ni \boldsymbol{0},
    \end{cases}
    \text{ in } \Omega \times (t^n, t^{n+1}), \\
    (\mathbf{p}(t^{n}), \mathbf{H}(t^{n})) = (\mathbf{p}^{n+2/4}, \mathbf{H}^{n+2/4}),
\end{cases}
\label{FS3}
\end{equation}
and set
\begin{equation}
\label{FS3.1}
    (\mathbf{p}^{n+3/4}, \mathbf{H}^{n+3/4}) = (\mathbf{p}(t^{n+1}), \mathbf{H}(t^{n+1})). 
\end{equation}

\noindent\underline{\emph{Fractional Step 4}}:
\begin{equation}
    \begin{cases}
        \begin{cases}
            \eta \frac{\partial \mathbf{p}}{\partial t} + D _{\mathbf{q}} J_3(\mathbf{p}) + \partial_{\mathbf{q}}I_{\Sigma}(\mathbf{p})= \boldsymbol{0}, \\
            \frac{\partial \mathbf{H}}{\partial t}  = \boldsymbol{0},
        \end{cases}
        \text{ in } \Omega \times (t^n, t^{n+1}), \\
        (\mathbf{p}(t^{n}), \mathbf{H}(t^{n})) = (\mathbf{p}^{n+3/4}, \mathbf{H}^{n+3/4}),
    \end{cases}
    \label{FS4.0}
\end{equation}
and set
\begin{equation}
\label{FS4}
    (\mathbf{p}^{n+1}, \mathbf{H}^{n+1}) = (\mathbf{p}(t^{n+1}), \mathbf{H}(t^{n+1})).
\end{equation}
The Lie scheme in (\ref{FS1})-(\ref{FS4}) is currently semi-discrete, as the time discretization for subproblems (\ref{FS1}), (\ref{FS2}), (\ref{FS3}) and (\ref{FS4}) has not yet been specified. 
To fully discretize the method, we employ a Marchuk–Yanenko-type splitting scheme; see \cite{glowinski2017some} for further details. 
That is, for $n\ge 0$, the update from $(\mathbf{p}^{n}, \mathbf{H}^{n})$ to $(\mathbf{p}^{n+1}, \mathbf{H}^{n+1})$ is obtained through the following fully discretized fractional steps:
\begin{equation}
\label{distime0}
   ( \mathbf{p}^{0}, \mathbf{H}^{0}) = ( \mathbf{p}_{0}, \mathbf{H}_{0}).
\end{equation}
Then, for $n\geq 0$, $(\mathbf{p}^{n}, \mathbf{H}^{n}) \to $ $(\mathbf{p}^{n+1/4}, \mathbf{H}^{n+1/4}) \to $
$(\mathbf{p}^{n+2/4}, \mathbf{H}^{n+2/4})\to $
$(\mathbf{p}^{n+3/4}$, $\mathbf{H}^{n+3/4})\to$
$(\mathbf{p}^{n+1}, \mathbf{H}^{n+1})$ as follows:
\begin{align}
    &\begin{cases}
        \eta \frac{\mathbf{p}^{n+1/4} - \mathbf{p}^{n}}{\tau} + D_{\mathbf{q}} J_1(\mathbf{p}^{n+1/4}, \mathbf{H}^{n}) = \boldsymbol{0}, \\
        \frac{\mathbf{H}^{n+1/4}-\mathbf{H}^{n}}{\tau} + \partial_{\mathbf{G}} J_1(\mathbf{p}^{n+1/4}, \mathbf{H}^{n+1/4}) \ni \boldsymbol{0},
    \end{cases} \qquad\text{in} ~~\Omega\Rightarrow (\mathbf{p}^{n+1/4}, \mathbf{H}^{n+1/4}),
    \label{distime1} \\
        &\begin{cases}
        \eta  \frac{\mathbf{p}^{n+2/4} - \mathbf{p}^{n+1/4}}{\tau} + \partial _{\mathbf{q}} J_{2}(\mathbf{p}^{n+2/4})\ni \boldsymbol{0}, \\
        \frac{\mathbf{H}^{n+2/4}-\mathbf{H}^{n+1/4}}{\tau} = \boldsymbol{0},
    \end{cases}
    \qquad\qquad\text{in} ~~\Omega\Rightarrow (\mathbf{p}^{n+2/4}, \mathbf{H}^{n+2/4}),
    \label{distime2} \\
        &\begin{cases}
        \eta  \frac{\mathbf{p}^{n+3/4} - \mathbf{p}^{n+2/4}}{\tau} + \partial _{\mathbf{q}} I_S(\mathbf{p}^{n+3/4},\mathbf{H}^{n+3/4})\ni \boldsymbol{0}, \\
        \frac{\mathbf{H}^{n+3/4}-\mathbf{H}^{n+2/4}}{\tau}  +\partial _{\mathbf{G}} I_S(\mathbf{p}^{n+3/4},\mathbf{H}^{n+3/4})\ni \boldsymbol{0},
    \end{cases}
    ~~\text{in} ~~\Omega\Rightarrow (\mathbf{p}^{n+3/4}, \mathbf{H}^{n+3/4}),
    \label{distime3}\\
        &\begin{cases}
       \eta  \frac{\mathbf{p}^{n+1} - \mathbf{p}^{n+3/4}}{\tau} + D_{\mathbf{q}} J_3(\mathbf{p}^{n+1}) + \partial_{\mathbf{q}}I_{\Sigma}(\mathbf{p}^{n+1})\ni \boldsymbol{0}, \\
       \frac{\mathbf{H}^{n+1}-\mathbf{H}^{n+3/4}}{\tau}  = \boldsymbol{0}.
    \end{cases}
    \text{in} ~~\Omega\Rightarrow (\mathbf{p}^{n+1}, \mathbf{H}^{n+1}).
    \label{distime4}
\end{align}

\section{Solutions to the subproblems}
\label{sec: sub}
In this section, we discuss the solutions to various subproblems that arise when applying the scheme \eqref{distime0}–\eqref{distime4} to solve problem \eqref{model3}.

\subsection{On the solution of the subproblem (\ref{distime1})}
\label{subsec:1/4}
\subsubsection{Computing $\bf{p}^{n+1/4}$}

In (\ref{distime1}), $\mathbf{p}^{n+1/4}$ is the solution to the following minimization problem:
\begin{equation}
\label{zuiyouproblem1}
        \mathbf{p}^{n+1/4} =\underset{\mathbf{q} \in\left(\mathcal{L}^{2}(\Omega)\right)^{2}}{\arg \min } \left[\frac{\eta}{2} \int_{\Omega}\vert\mathbf{q}-\mathbf{p}^{n}\vert^{2} \mathrm{d} \mathbf{x} + \frac{\tau\alpha}{2}\int_{\Omega} \int_{0}^{2\pi} \frac{\vert\mathbf{t}^{\top} \mathbf{H}^{n} \mathbf{t}\vert}{  1 + (\mathbf{q} : \mathbf{t})^2}  \, \mathrm{d} \theta \, \mathrm{d} \mathbf{x}\right].
\end{equation}
By differentiating the functional in (\ref{zuiyouproblem1}), $\mathbf{p}^{n+1/4} = [p_{1}^{n+1/4}, p_{2}^{n+1/4} ]^{\top}$ satisfies
\begin{equation}
\label{equa:fix1}
    \eta(\mathbf{p}^{n+1/4}- \mathbf{p}^{n})-\tau \alpha \int_{0}^{2\pi} \frac{\vert\mathbf{t}^{\top} \mathbf{H}^{n} \mathbf{t}\vert \cdot (\mathbf{p}^{n+1/4} : \mathbf{t}) \cdot \mathbf{t}}{(1 + (\mathbf{p}^{n+1/4} : \mathbf{t})^2)^2} \, \mathrm{d} \theta=\mathbf{0},
\end{equation}
which can be solved using Newton’s method or a fixed-point iteration \cite{walker2011anderson}. Since the variable $\mathbf{p}^{n+1/4}$ appears in the denominator, applying Newton’s method involves differentiation, which leads to significant complexity. Instead, we adopt a fixed-point method that not only simplifies computation but also demonstrates fast convergence properties. The fixed-point algorithm is described in \cref{alg:Fix}, where \(\rho_{1} \in (0, 1]\) is a parameter that controls the updating rate of \(\mathbf{q}\). Finally, we set \(\mathbf{p}^{n+1/4} = \mathbf{q}^{k+1}\).

\begin{algorithm}[H]
\caption{Fixed-point solution for solving the problem (\ref{zuiyouproblem1})}
\label{alg:Fix}
\begin{algorithmic}
\STATE{\textbf{Input:} the variable $\mathbf{p}^{n}$, the parameters $\tau$, $\alpha$, $\eta$, $\rho_{1}$ and stopping threshold $tol$;}
\STATE{\textbf{Initialization:} $\mathbf{q}^{0} = \mathbf{p}^{n}$, $k=0$;}
\WHILE{(not converge)}
\STATE{1. Compute $\mathbf{F}$ by:\\
\indent \quad\quad~~~~~~~~$\mathbf{F}(\mathbf{q}^{k},\theta) =  \frac{\vert\mathbf{t}^{\top} \mathbf{H}^{n} \mathbf{t}\vert \cdot (\mathbf{q}^{k}: \mathbf{t}) \cdot \mathbf{t}}{(1 + (\mathbf{q}^{k}: \mathbf{t})^2)^2}; \label{fix1}$ }
\STATE{2. Compute $\Tilde{\mathbf{q}}$ from:\\
\indent \quad\quad~~~~~~~
$\widetilde{\mathbf{q}}^{k} = 
      \mathbf{q}^{0} + \frac{\tau\alpha}{\eta}\int_{0}^{2\pi} \mathbf{F}(\mathbf{q}^{k},\theta)\mathrm{d} \theta\label{fix2}$;}
\STATE{3. Update $\mathbf{q}^{k+1}$ from:\\
\indent \quad\quad~~~~~~~~$\mathbf{q}^{k+1} =(1-\rho_{1})\mathbf{q}^{k} + \rho_{1}\widetilde{\mathbf{q}}^{k}; \label{fix3}$
}
\STATE{4. Check the stopping criterion: \\
\indent \quad\quad~~~~~~~~$\|\mathbf{q}^{k+1} - \mathbf{q}^{k}\|_{\infty} \leq tol$; }
\ENDWHILE
\RETURN $\mathbf{p}^{n+1/4} = \mathbf{q}^{k+1}$.
\end{algorithmic}
\end{algorithm}
In \cref{alg:Fix}, the $\widetilde{\mathbf{q}}$-subproblem is solved by discretizing the integral through the composite trapezoidal rule, partitioning $[0, 2\pi]$ into $N$ equally spaced subintervals with the corresponding angular nodes $\theta_\ell$:
\begin{equation*}
\widetilde{\mathbf{q}}^{k} = \mathbf{q}^{0} + \frac{\tau\alpha}{\eta} \cdot \frac{2\pi}{N} \sum_{\ell=1}^{N} \mathbf{F}(\mathbf{q}^{k}, \theta_{\ell}),    
\end{equation*}
where the discrete angular nodes are given by $\theta_{\ell} = 2\pi(\ell-1)/N$ for $\ell = 1, \dots, N$, and we take $N = 8$ in all experiments.

\subsubsection{Computing $\mathbf{H}^{n+1/4}$} 
In (\ref{distime1}), $\mathbf{H}^{n+1/4}$ is the solution to
\begin{equation}
\label{sub_H1/4_01}
      \mathbf{H}^{n+1/4} = \underset{\mathbf{G} \in\left(\mathcal{L}^{2}(\Omega)\right)^{2\times 2}}{\arg \min }\left[ \frac{1}{2} \int_{\Omega} \vert \mathbf{G} - \mathbf{H}^n\vert^{2}_{F} \, \mathrm{d}\mathbf{x} +  \frac{\tau\alpha}{2}\int_{\Omega} \int_{0}^{2\pi} \frac{\vert\mathbf{t}^{\top} \mathbf{G} \mathbf{t}\vert}{ 1 + (\mathbf{p}^{n+1/4} : \mathbf{t})^2 }  \, \mathrm{d} \theta \, \mathrm{d} \mathbf{x} \right],
\end{equation}
where $\vert \cdot \vert_{F}$ is Frobenius norm defined by $\vert \mathbf{G} \vert_{F} = \sqrt{|G_{11}|^{2}+|G_{12}|^{2}+|G_{21}|^{2}+|G_{22}|^{2}}$. 
By denoting $\mathbf{B}=\mathbf{H}^{n}$, $\Delta_{1}(\theta) = \left(1 + (\mathbf{p}^{n+1/4} : \mathbf{t})^2\right)^{-1} > 0$, the minimization problem (\ref{sub_H1/4_01}) can be rewritten in a more compact form as
\begin{equation}
\label{sub_H1/4_02}
    \mathbf{M} = \underset{\mathbf{G} \in\left(\mathcal{L}^{2}(\Omega)\right)^{2\times 2}}{\arg \min }\left[ \frac{1}{2} \int_{\Omega} \vert \mathbf{G} - \mathbf{B}\vert^{2}_{F} \, \mathrm{d}\mathbf{x} +  \frac{\tau \alpha}{2}\int_{\Omega} \int_{0}^{2\pi}\Delta_{1}(\theta) \vert\mathbf{t}^{\top} \mathbf{G} \mathbf{t}\vert  \, \mathrm{d} \theta \, \mathrm{d} \mathbf{x} \right].
\end{equation}
The angular integral over $[0, 2\pi]$ is discretized into eight uniformly spaced directions, but the computation is reduced to four due to the integrand's $\pi$-periodicity, which allows pairing $\theta$ with $\theta + \pi$ as they yield identical values. The minimization is thus performed pointwise on $\Omega$ over the first four directions, yielding:
\begin{equation}
\label{sub_H1/4_03}
    \underset{\mathbf{G} \in \left(\mathcal{L}^{2}(\Omega)\right)^{2 \times 2}}{\min} ~~\frac{1}{2} \vert\mathbf{G} - \mathbf{B} \vert^{2}_{F} + \frac{\pi}{4} \tau \alpha \sum_{\ell=1}^{4} \Delta_{1}(\theta_{\ell}) \left| \mathbf{t}(\theta_{\ell})^{\top} \mathbf{G} \mathbf{t}(\theta_{\ell}) \right|,
\end{equation}
where $\theta_{\ell}$ denotes predefined discrete directional angles, and $\mathbf{t}(\theta_{\ell})$ is the unit vector corresponding to the directional angle $\theta_{\ell}$. For ease of handling, we vectorize the elements of matrices $\mathbf{G}$ and $\mathbf{B}$, thereby transforming this matrix optimization problem into a vector optimization problem. 
Let $\mathbf{w} = \mathrm{vec}(\mathbf{G}) =[G_{11}, G_{12}, G_{21}, G_{22}]^{\top} \in \mathbb{R}^{4}$ and $\mathbf{b} = \mathrm{vec}(\mathbf{B})=[b_{11}, b_{12}, b_{21}, b_{22}]^{\top} \in \mathbb{R}^{4}$. The nonsmooth term $\left| \mathbf{t}(\theta_{\ell})^{\top} \mathbf{G} \mathbf{t}(\theta_{\ell}) \right| = \left| \mathbf{a}_{\ell}^{\top} \mathbf{w} \right|$, where $\mathbf{a}_{\ell} = [\cos^{2} \theta_{\ell}, \cos \theta_{\ell} \sin \theta_{\ell}, \cos \theta_{\ell} \sin \theta_{\ell}, \sin^{2} \theta_{\ell}]^{\top} \in \mathbb{R}^{4}$. 
Thus, we can rewrite \eqref{sub_H1/4_03} as:
\begin{equation}
\label{sub_H1/4_04}
   \min_{\mathbf{w} \in \mathbb{R}^{4}} \frac{1}{2} \|\mathbf{w} - \mathbf{b}\|^{2} + \frac{\pi}{4} \tau \alpha \sum_{\ell=1}^{4} \Delta_{1}(\theta_{\ell}) \left| \mathbf{a}_{\ell}^{\top} \mathbf{w} \right|.
\end{equation}
We introduce an auxiliary variable \(\mathbf{u} = \mathbf{A} \mathbf{w} \in \mathbb{R}^{4}\), where \(\mathbf{A} = [\mathbf{a}_{1}^{\top}; \mathbf{a}_{2}^{\top}; \mathbf{a}_{3}^{\top}; \mathbf{a}_{4}^{\top}] \in \mathbb{R}^{4 \times 4}\), \(u_{\ell} = \mathbf{a}_{\ell}^{\top} \mathbf{w}\), and rewrite the minimization problem \eqref{sub_H1/4_04} into a constrained minimization as follows:
\begin{equation}
\label{sub_H1/4_05}
    \begin{aligned}
     \min_{\mathbf{w}, \mathbf{u} \in \mathbb{R}^{4}} & \quad \frac{1}{2} \|\mathbf{w} - \mathbf{b} \|^{2} + \frac{\pi}{4} \tau \alpha \sum_{\ell=1}^{4} \Delta_{1}(\theta_{\ell}) \left| u_{\ell} \right|, \\
    \text{s.t.} & \quad \mathbf{A} \mathbf{w} - \mathbf{u} = 0. 
    \end{aligned}
\end{equation}
The proximal augmented Lagrangian of the above constrained minimization problem can be defined as:
\begin{equation}
\label{sub_H1/4_06}
    \mathcal{L}(\mathbf{w}, \mathbf{u}; \mathbf{\Lambda}) = \frac{1}{2} \|\mathbf{w} - \mathbf{b}\|^{2} + C \sum_{\ell=1}^{4} \Delta_{1}(\theta_{\ell}) \left| u_{\ell} \right| + 
    \mathbf{\Lambda}^{\top} (\mathbf{A} \mathbf{w} - \mathbf{u}) + \frac{\rho_{2}}{2} \| \mathbf{A} \mathbf{w} - \mathbf{u}\|^{2},
\end{equation}
where $\mathbf{\Lambda} \in \mathbb{R}^{4}$ represents the Lagrangian multiplier, $\rho_{2}$ is a positive parameter, and $C = \frac{\pi}{4} \tau \alpha$. Given some $(\mathbf{w}^{n,k}, \mathbf{u}^{n,k})$, we employ the ADMM-based algorithm to iteratively solve the variables $\mathbf{w}^{n,k+1}$ and $\mathbf{u}^{n,k+1}$ as follows
\begin{equation}
    \begin{cases}
        \mathbf{w}^{n, k+1} = \underset{\mathbf{w}}{\arg \min} \, \mathcal{L}(\mathbf{w}, \mathbf{u}^{n,k}; \mathbf{\Lambda}^{n, k}), \\
        \mathbf{u}^{n, k+1} = \underset{\mathbf{u}}{\arg \min} \, \mathcal{L}(\mathbf{w}^{n,k+1}, \mathbf{u}; \mathbf{\Lambda}^{n, k}),
    \end{cases}
\end{equation}
followed by an update of the Lagrangian multiplier $\mathbf{\Lambda}^{n, k+1}$. 
Here, the superscripts $n$ and $k$ denote the indices for the outer and inner loops, respectively.
The $\mathbf{w}$-subproblem is a standard quadratic optimization problem that admits a closed-form solution. The $\mathbf{u}$-subproblem is separable and can be efficiently solved using the shrinkage operator \cite{beck2009fast}, defined as:
\begin{equation*}
\label{eq:shrinkage}
    \mathrm{shrinkage}(a, b) = \max(\vert a \vert - b, 0) \cdot \frac{a}{\vert a \vert}.
\end{equation*}
We provide the detailed algorithm as \cref{alg:ADMM}. Finally, after solving the vectorized problem, the resulting variable $\mathbf{w}^{n,k+1}$ is reshaped back to the matrix form $\mathbf{H}^{n+1/4} = \mathrm{mat}(\mathbf{w}^{n,k+1})$.


\begin{algorithm}
\caption{ADMM-based algorithm for solving problem (\ref{sub_H1/4_05})}
\label{alg:ADMM}
\begin{algorithmic}
\STATE{\textbf{Input:}  the variable $\mathbf{b} = \mathrm{vec}(\mathbf{H}^{n}$), parameters $\alpha$, $\tau$, $\rho_{2}$,  maximum iteration number $I_{max}$ and stopping threshold $tol$;}
\STATE{\textbf{Initialization:} $\mathbf{w}^{n, 0}=\mathbf{b}$, $\mathbf{u}^{n, 0}=\mathbf{A}\mathbf{w}^{n, 0}$, $\mathbf{\Lambda}^{n, 0}=\mathbf{\Lambda}^{n-1}$, $k=0$;}
\WHILE{(not converge and $k\leq I_{max}$)}
\STATE{1. Compute $\mathbf{w}^{n, k+1}$ from:\\
\indent \quad\quad~~~~~~ $    \mathbf{w}^{n, k+1} = \left( \mathbf{I} + \rho \mathbf{A}^{\top} \mathbf{A}\right)^{-1} \left( \mathbf{b} - \mathbf{A}^{\top} \mathbf{\Lambda}^{n,k} + \rho_{2} \mathbf{A}^{\top} \mathbf{u}^{n, k} \right);$ }
\STATE{2. Compute $\mathbf{u}^{n, k+1}$ from:\\
\indent \qquad\qquad~~
$u_{\ell}^{n, k+1} =\mathrm{shrinkage}\left( (\mathbf{A}\mathbf{w}^{n, k+1})_{\ell} + \frac{(\mathbf{\Lambda}^{n,k})_{\ell}}{\rho_{2}}, \frac{C \Delta_{1}(\theta_{\ell})}{\rho_{2}} \right), ~\mbox{for}~ \ell =1,\ldots,8;$}
\STATE{3. Update $\mathbf{\Lambda}^{n, k+1}$ from:\\
\begin{center}
    $\mathbf{\Lambda}^{n, k+1} = \mathbf{\Lambda}^{n, k}+\rho_{2}(\mathbf{A}\mathbf{w}^{n, k+1}-\mathbf{u}^{n, k+1});$
\end{center}
}
\STATE{4. Check convergence via primal and dual residuals:\\
\begin{center}
   $\|\mathbf{A}\mathbf{w}^{n,k+1}-\mathbf{u}^{n,k+1}\|_{2}\leq tol ~~\mbox{and}~~
 \|\rho_{2}\mathbf{A}^{\top}(\mathbf{u}^{n,k+1}-\mathbf{u}^{n, k})\|_{2}\leq tol$;
\end{center}
}
\ENDWHILE
\RETURN $\mathbf{H}^{n+1/4}$ = $\mathrm{mat}(\mathbf{w}^{n,k+1}$) and $\boldsymbol{\Lambda}^{n,k+1}$.
\end{algorithmic}
\end{algorithm}

\subsection{On the solution of the subproblem (\ref{distime2})}
\label{subsec:2/4}
We can directly obtain the solution $\mathbf{H}^{n+2/4}$ from (\ref{distime2}) as
\begin{equation}
\label{equ:Hn2/4}
    \mathbf{H}^{n+2/4} = \mathbf{H}^{n+1/4}.
\end{equation}
Then, $\mathbf{p}^{n+2/4}$ is obtained by minimizing the following problem:
\begin{equation}
    \min_{\mathbf{q}\in( \mathcal{L}^{2}(\Omega))^{2}}\left[ \frac{\eta}{2}\int_{\Omega}\vert \mathbf{q} - \mathbf{p}^{n+1/4} \vert^{2} \mathrm{d} \mathbf{x} + \tau \beta \int_{\Omega} \vert \mathbf{q}\vert \mathrm{d} \mathbf{x}\right].
\end{equation}
This convex problem has the closed-form solution:
\begin{equation}
\label{equ:pn2/4}
    \mathbf{p}^{n+2/4} = \max \left\{ 0, 1 - \frac{\tau \beta / \eta}{\vert\mathbf{p}^{n+1/4}\vert} \right\} \mathbf{p}^{n+1/4}.
\end{equation}
\subsection{On the solution of the subproblem (\ref{distime3})}
Similarly, system (\ref{distime3}) for $(\mathbf{p}^{n+3/4}, \mathbf{H}^{n+3/4})$ is the Euler-Lagrange equation of the following minimization problem
\begin{equation}
\label{2ziwenti11}
\left\{
\begin{aligned}
\mathbf{H}^{n+3/4}&=\nabla \mathbf{p}^{n+3 / 4}, \\
\mathbf{p}^{n+3/4}&=\underset{\mathbf{q} \in\left(\mathcal{H}^{1}(\Omega)\right)^{2}}{\arg \min }\left[\frac{1}{2} \int_{\Omega}\left(\eta\vert\mathbf{q}-\mathbf{p}^{n+2/4}\vert^{2}+\vert\nabla \mathbf{q}-\mathbf{H}^{n+2/4}\vert^{2}_{F}\right) \mathrm{d} \mathbf{x}\right]. \\
\end{aligned}\right.  
\end{equation}
It follows that $\mathbf{p}^{n+3/4}$ is the unique solution of the following well-posed linear variational problem
\begin{equation}
\label{pH3/4:well}
\left\{
\begin{aligned}
&\textbf{p}^{n+3/4} \in (\mathcal{H}^1(\Omega))^2, \\
&\int_{\Omega} \left( \eta\textbf{p}^{n+3/4} \cdot \textbf{q} + \nabla \textbf{p}^{n+3/4} : \nabla \textbf{q} \right) \mathrm{d}\mathbf{x} = \int_{\Omega} \left( \eta \textbf{p}^{n+2/4} \cdot \textbf{q} + \textbf{H}^{n+2/4} \cdot \nabla \textbf{q} \right) \mathrm{d}\mathbf{x} , \\
&\forall \textbf{q} \in (\mathcal{H}^1(\Omega))^2.
\end{aligned}\right.  
\end{equation}
The solution to (\ref{pH3/4:well}) is the weak solution of the following linear elliptic problem:
\begin{equation}
\label{pH3/4}
 \begin{cases} 
-\nabla^2 p_k^{n+3/4} + \eta p_k^{n+3/4} = \eta p_k^{n+2/4} - \nabla \cdot \mathbf{H}_k^{n+2/4} & \text{in } \Omega, \\
(\nabla p_{k}^{n+3/4} - \mathbf{H}^{n+2/4})\cdot \mathbf{n} = 0 & \text{on } \partial \Omega,
\end{cases}   \quad \mbox{for} ~~ k=1,2,
\end{equation}
where $\mathbf{H}_k^{n+2/4}$ denotes the $k$-th row of $\mathbf{H}^{n+2/4}$.

\subsection{On the solution of the subproblem (\ref{distime4})}
We clearly have
\begin{equation}
\label{Hn+1}
    \mathbf{H}^{n+1}= \mathbf{H}^{n+3/4}.
\end{equation}
And $\mathbf{p}^{n+1}$ is the solution to 
\begin{equation}
\label{3ziwenti11}
    \begin{cases}
        \begin{aligned}
            \mathbf{p}^{n+1}&=\nabla u^{n+1}, \\
            u^{n+1}&=\underset{v \in\mathcal{H}^{1}(\Omega)}{\arg \min }\left[ \frac{1}{2} \int_{\Omega}\eta\vert\nabla v-\mathbf{p}^{n+3 / 4}\vert^{2} \mathrm{d} \mathbf{x}+ \frac{\gamma\tau}{2}\int_{\Omega}\vert f - v\vert ^{2}\mathrm{d} \mathbf{x}\right],
    \end{aligned}
    \end{cases}
\end{equation}
where $u^{n+1}$ is the unique solution to the linear variational problem
\begin{equation}
\label{diedai31}
    \begin{cases}
        \begin{aligned}
        &u^{n+1} \in \mathcal{H}^{1}(\Omega),\\
        & \int_\Omega \eta\nabla u^{n+1}  \cdot \nabla v \, \mathrm{d}\mathbf{x} + \gamma \tau \int_\Omega u^{n+1} v \, \mathrm{d}\mathbf{x} = \eta \int_\Omega \mathbf{p}^{n+3/4} \cdot \nabla v \, \mathrm{d}\mathbf{x} + \gamma \tau \int_\Omega f v \, \mathrm{d}\mathbf{x},\\
        & \forall v \in   \mathcal{H}^{1}(\Omega).
        \end{aligned}
    \end{cases}
\end{equation}
Additionally, $u^{n+1}$ is the weak solution to the following Neumann problem:
\begin{equation}
\label{H}
 \left\{\begin{array}{ll}
-\eta \nabla^{2}u^{n+1} + \gamma \tau u^{n+1} = \gamma \tau f- \eta \nabla \cdot \mathbf{p}^{n+3/4} & \text { in } \Omega, \\
(\nabla u^{n+1} - \mathbf{p}^{n+3/4})\cdot \mathbf{n} = 0 & \text { on } \partial \Omega.
\end{array}\right.   
\end{equation}
Finally, we summarize the operator-splitting algorithm for solving problem \eqref{model3} in Algorithm \ref{alg:osm}. 
\begin{algorithm}[H]
\caption{An operator-splitting method for solving the problem (\ref{model3})}
\label{alg:osm}
\begin{algorithmic}
\STATE{\textbf{Input:} Noisy image $f$, model parameters: $\alpha, \beta$, $\gamma$, algorithm parameters: $\eta,\tau$ and stopping threshold $\epsilon$;}
\STATE{\textbf{Initialization:} $u^{0}=f$, $(\mathbf{p}^{0},\mathbf{H}^{0})=(\mathbf{p}_{0},\mathbf{H}_{0})$, and set $n=0$;}
\WHILE{not converge}
\STATE{1. Solve (\ref{distime1}) using the methods in Section \ref{subsec:1/4} to obtain $(\mathbf{p}^{n+1/4}, \mathbf{H}^{n+1/4})$;}
\STATE{2. Solve (\ref{distime2}) using  (\ref{equ:Hn2/4}) and (\ref{equ:pn2/4}) to obtain $(\mathbf{p}^{n+2/4}, \mathbf{H}^{n+2/4})$;}
\STATE{3. Solve (\ref{distime3}) using (\ref{2ziwenti11}) and (\ref{pH3/4}) to obtain $(\mathbf{p}^{n+3/4}, \mathbf{H}^{n+3/4})$;}
\STATE{4. Solve (\ref{distime4}) using (\ref{Hn+1})-(\ref{H}) to obtain $(u^{n+1}, \mathbf{p}^{n+1}, \mathbf{H}^{n+1})$;}
\STATE{5. Check the stopping criterion:\\
\begin{center}
$\|u^{n+1} - u^{n}\|_{2}/ \|u\|_{2} \leq \epsilon$; 
\end{center}}
\STATE{6. Set $n = n + 1$;}
\ENDWHILE
\RETURN $u^{n+1}$.
\end{algorithmic}
\end{algorithm}

\subsection{Initial conditions}
\label{subsec: Initial condition}
The numerical scheme defined by (\ref{distime1})-(\ref{distime4}) requires initial conditions $(\mathbf{p}_{0}, \mathbf{H}_{0})$. Two common approaches are available for selection.
First, a direct initialization utilizes image derivatives:
\begin{equation}
\label{inital1}
    \mathbf{p}_{0} = \nabla f ~~\mbox{and}~~ \mathbf{H}_{0}= \nabla \mathbf{p}_{0}.
\end{equation}
Alternatively, for improved stability, a smoothed initialization can be employed, involving a small parameter $\varepsilon > 0$. This approach first determines $u_{0}$ as the weak solution of the following elliptic problem: 
\begin{equation}
\label{inital2}
\left\{\begin{aligned}
&u_0 \in \mathcal{H}^1(\Omega), \\
&\int_{\Omega} u_0 v \, \mathrm{d}\mathbf{x} + \varepsilon \int_{\Omega} \nabla u_0 \cdot \nabla v \, \mathrm{d}\mathbf{x} = \int_{\Omega} f v \, \mathrm{d}\mathbf{x}, \quad \forall v \in \mathcal{H}^1(\Omega).
\end{aligned}\right.  
\end{equation}
This variational formulation corresponds to the following Neumann problem:
\begin{align}
\label{inital3}
\begin{cases}
u_0 - \varepsilon \nabla^2 u_0 = f & \text{in } \Omega, \\
\nabla u_0 \cdot \mathbf{n}( = \partial u_0 / \partial \mathbf{n} )= 0 & \text{on } \partial \Omega.
\end{cases}
\end{align}
Subsequently, $\mathbf{p}_0$ and $\mathbf{H}_0$ are defined as:
\begin{equation}
\label{inital4}
    \mathbf{p}_0 = \nabla u_0~~\mbox{and}~~ \ \mathbf{H}_0 = \nabla \mathbf{p}_0.
\end{equation}

\subsection{On periodic boundary conditions}
\label{subsec:perbound}
While the operator-splitting method and its associated subproblem solvers, as presented in earlier sections, have been developed under the assumption of Neumann boundary conditions, many image processing applications benefit from the use of periodic boundary conditions. This subsection addresses this by detailing the necessary adaptations to our proposed algorithms and their solvers to incorporate periodic boundary conditions. These adaptations are vital for efficient numerical solutions, particularly when using Fourier-based methods.

Consider a domain $\Omega = [0, L_1] \times [0, L_2]$. The first modification required is to replace the function space $\mathcal{H}^1(\Omega)$ with $\mathcal{H}_P^1(\Omega)$, defined as follows:
\begin{equation*}
\mathcal{H}_P^1(\Omega) = \left\{ v \in \mathcal{H}^1(\Omega) : v(0, :) = v(L_1, :), v( :, 0) = v(:, L_2) \right\}.
\end{equation*}
All previously defined sets and functions implicitly dependent on the function space naturally inherit this periodicity. 
Then, problem (\ref{model3}) is replaced by
\begin{multline}
      \min_{\substack{ \mathbf{q} \in (\mathcal{H}_{P}^1(\Omega))^2,\ \\ \mathbf{G} \in (\mathcal{L}^2(\Omega))^{2 \times 2}}} J(\mathbf{q},\mathbf{G}) = \left [ \frac{\alpha}{2}\int_{\Omega} \int_{0}^{2\pi} \frac{\vert\mathbf{t}^{\top} \mathbf{G} \mathbf{t}\vert}{ 1 + (\mathbf{q} : \mathbf{t})^2}  \, \mathrm{d} \theta \, \mathrm{d} \mathbf{x} +\beta \int_{\Omega} \vert \mathbf{q}\vert \mathrm{d}\mathbf{x} \right. \\ \left. +  \frac{\gamma}{2}\int_{\Omega} \vert f-v_{\mathbf{q}} \vert ^{2} \mathrm{d} \mathbf{x} + I_{\Sigma}(\mathbf{q}) + I_{S}(\mathbf{q}, \mathbf{G}) \right],
\end{multline}
with the boundary condition (\ref{boundcondition}) replaced by
\begin{equation}
\label{bound2}
\begin{cases}
\nabla^2 v_{\mathbf{q}} = \nabla \cdot \mathbf{q} ~~ \text{in } \Omega, \\
v_{\mathbf{q}}~\text{verifies periodic boundary conditions},\\
(\nabla v_{\mathbf{q}} - \mathbf{q}) \cdot \mathbf{e}_{j} = 0~\text{is periodic in the $Ox_{j}$-direction, $\forall j=1,2,$} \\
 \int_{\Omega}v_{\mathbf{q}}~ \mathrm{d}\mathbf{x} = \int_{\Omega}f~\mathrm{d}\mathbf{x},
\end{cases}
\end{equation}
where $\mathbf{e}_j$ is the unit vector along the $Ox_{j}$-axis. Then, the updates $\mathbf{p}^{n+3/4}$ and $\mathbf{H}^{n+3/4}$ are obtained by solving the following system, replacing (\ref{2ziwenti11}) and (\ref{pH3/4}):
\begin{equation}
\left\{\begin{array}{l}
\mathbf{H}^{n+3/4}=\nabla \mathbf{p}^{n+3/4}, \\
\mathbf{p}^{n+3/4}=\underset{\mathbf{q} \in\left(\mathcal{H}_{P}^{1}(\Omega)\right)^{2}}{\arg \min }\left[\frac{1}{2} \int_{\Omega}\left(\eta\vert\mathbf{q}-\mathbf{p}^{n+2/4}\vert^{2}+\vert\nabla \mathbf{q}-\mathbf{H}^{n+2/4}\vert^{2}_{F}\right) \mathrm{d} \mathbf{x}\right],
\end{array}\right.
\end{equation}
and 
\begin{equation}
\label{cond-inital0}
 \begin{cases} 
-\nabla^2 p_k^{n+3/4} + \eta p_k^{n+3/4} = \eta p_k^{n+2/4} - \nabla \cdot \mathbf{H}_k^{n+2/4}, ~~\text{in } \Omega, \\
p_{k}^{n+3/4}(0, x_2) = p_{k}^{n+3/4}(L_1, x_2), \ 0 < x_2 < L_2, \\
p_{k}^{n+3/4}(x_1, 0) = p_{k}^{n+3/4}(x_1, L_2), \ 0 < x_1 < L_1, \\
\left( \frac{\partial p_k^{n+3/4}}{\partial x_1} - H_{k1}^{n+2/4} \right)(0, x_2) = \left( \frac{\partial p_k^{n+3/4}}{\partial x_1} - H_{k1}^{n+2/4} \right)(L_1, x_2), \ 0 < x_2 < L_2, \\
\left( \frac{\partial p_k^{n+3/4}}{\partial x_2} - H_{k2}^{n+2/4} \right)(x_1, 0) = \left( \frac{\partial p_k^{n+3/4}}{\partial x_2} - H_{k2}^{n+2/4} \right)(x_1, L_2), \ 0 < x_1 < L_1,\\
\end{cases}   
\end{equation}
for $k =$ 1, 2, respectively.

For $\mathbf{p}^{n+1}$, we replace (\ref{3ziwenti11}) and (\ref{H}) by
\begin{equation}
\begin{cases}
\begin{aligned}
\mathbf{p}^{n+1}&=\nabla u^{n+1}, \\
u^{n+1}&=\underset{u \in\mathcal{H}_{P}^{1}(\Omega)}{\arg \min }\left[\frac{1}{2} \int_{\Omega}\eta\vert\nabla v-\mathbf{p}^{n+3/4}\vert ^{2} \mathrm{d} \mathbf{x}+ \frac{\gamma\tau}{2}\int_{\Omega}\vert f - v\vert ^{2}\mathrm{d} \mathbf{x}\right],\\
\end{aligned}
\end{cases}
\end{equation}
and 
\begin{align}
\label{cond-inita1}
\begin{cases}
-\eta \nabla^2 u^{n+1} + \gamma\tau u^{n+1} = \gamma \tau f - \eta\nabla \cdot \mathbf{p}^{n+2/3}, ~~~\quad \qquad \qquad \text{ in } \Omega, \\
u^{n+1}(0, x_2) = u^{n+1}(L_1, x_2), \qquad \qquad \qquad \qquad \qquad \qquad \quad 0 < x_2 < L_2, \\
u^{n+1}(x_1, 0) = u^{n+1}(x_1, L_2), \qquad \qquad \qquad \qquad \qquad \qquad \quad 0 < x_1 < L_1, \\
\left( \frac{\partial u^{n+1}}{\partial x_1} - p_1^{n+3/4} \right)(0, x_2) = \left( \frac{\partial u^{n+1}}{\partial x_1} - p_1^{n+3/4} \right)(L_1, x_2), ~\quad 0 < x_2 < L_2, \\
\left( \frac{\partial u^{n+1}}{\partial x_2} - p_2^{n+3/4} \right)(x_1, 0) = \left( \frac{\partial u^{n+1}}{\partial x_2} - p_2^{n+3/4} \right)(x_1, L_2), ~\quad 0 < x_1 < L_1,
\end{cases}
\end{align}
respectively. On the other hand, we replace the initial conditions (\ref{inital2}) and (\ref{inital3}) by
\begin{align}
\label{cond-inital2}
\begin{cases}
u_0 \in \mathcal{H}_{P}^1(\Omega), \\
\int_{\Omega} u_0 v \, \mathrm{d} \mathbf{x} + \varepsilon \int_{\Omega} \nabla u_0 \cdot \nabla v \, \mathrm{d} \mathbf{x} = \int_{\Omega} f v \, \mathrm{d} \mathbf{x}, \\
\forall v \in \mathcal{H}_{P}^1(\Omega),
\end{cases}
\end{align}
and
\begin{align}
\label{cond-inital3}
\begin{cases}
u_0 - \varepsilon \nabla^2 u_0 = f \text{ in } \Omega, \\
u_0(0, x_2) = u_0(L_1, x_2), \quad~~~ 0 < x_2 < L_2, \\
u_0(x_1, 0) = u_0(x_1, L_2), \quad~~~ 0 < x_1 < L_1, \\
\frac{\partial u_0}{\partial x_1}(0, x_2) = \frac{\partial u_0}{\partial x_1}(L_1, x_2), \quad 0 < x_2 < L_2, \\
\frac{\partial u_0}{\partial x_2}(x_1, 0) = \frac{\partial u_0}{\partial x_2}(x_1, L_2), \quad 0 < x_1 < L_1,
\end{cases}
\end{align}
respectively. 
Equations (\ref{cond-inital0}), (\ref{cond-inita1}), and (\ref{cond-inital3}) represent linear elliptic PDEs with periodic boundary conditions. These can be efficiently solved using the Fast Fourier Transform (FFT) method after finite difference discretization. 
Throughout the remainder of this work, we consistently apply periodic boundary conditions.

\section{Spacial discretization}
\label{sec:dis}
This section details the finite difference discretization of equations (\ref{distime0})-(\ref{distime4}) on a uniform \(M \times N\) grid over \(\Omega = (0, L_1) \times (0, L_2)\), with spatial step size \(h = L_1/M = L_2/N\). The discrete approximation \(u_{i,j}\) corresponds to the nodal value \(u(ih, jh)\),  and all field variables satisfy periodic boundary conditions.

We begin by defining the forward ($+$) and backward ($-$) finite difference operators for $1 \leq i \leq M$, $1 \leq j \leq N$:
\begin{equation*}
\partial_{1}^{-} v(i, j)=\left\{\begin{array}{ll}
(v(i, j)-v(i-1, j)) / h, & 1<i \leq M, \\
\left(v(1, j)-v\left(M, j\right)\right) / h, & i=1,
\end{array}\right.
\end{equation*}
\begin{equation*}
\partial_{2}^{-} v(i, j)=\left\{\begin{array}{ll}
(v(i, j)-v(i, j-1)) / h, & 1<j \leq N, \\
\left(v(i, 1)-v\left(i, N\right)\right) / h, & j=1,
\end{array}\right.
\end{equation*}
\begin{equation*}
\partial_{1}^{+} v(i, j)=\left\{\begin{array}{ll}
(v(i+1, j)-v(i, j)) / h, & 1 \leq i < M, \\
\left(v(1, j)-v\left(M, j\right)\right) / h, & i=M,
\end{array}\right.
\end{equation*}
\begin{equation*}
\partial_{2}^{+} v(i, j)=\left\{\begin{array}{ll}
(v(i, j+1)-v(i, j)) / h, & 1 \leq j < N, \\
\left(v(i, 1)-v\left(i, N\right)\right) / h, & j=N.
\end{array}\right.
\end{equation*} 
With the above notation, we define the discrete forward ($+$) and backward ($-$) gradient operators for a scalar function $v$ as:
\begin{equation*}
\nabla^{\pm} v(i,j) = (\partial_1^{\pm} v(i,j), \partial_2^{\pm} v(i,j)).
\end{equation*}
and for a vector function $\mathbf{q} = (q_1, q_2)$:
\begin{equation*}
\operatorname{div}^\pm\mathbf{q}(i,j)=\partial_1^\pm q_1(i,j)+\partial_2^\pm q_2(i,j),\quad\nabla^\pm\mathbf{q}(i,j)=
\begin{pmatrix}
\partial_1^\pm q_1(i,j) & \partial_1^\pm q_1(i,j) \\
\partial_2^\pm q_2(i,j) & \partial_2^\pm q_2(i,j)
\end{pmatrix}.
\end{equation*}
The shifting and identity operators are defined by:
\begin{equation}
\mathcal{S}_1^\pm v(i,j) = v(i \pm 1, j), ~~ \mathcal{S}_2^\pm v(i,j) = v(i,j \pm 1), ~~ \mathcal{I} v(i,j) = v(i,j).
\end{equation}
Denoting the discrete Fourier transform and its inverse by $\mathcal{F}$ and $\mathcal{F}^{-1}$ respectively, we obtain
\begin{align}
\mathcal{F}(\mathcal{S}_1^\pm v)(i,j) &= (\cos z_i \pm \sqrt{-1} \sin z_i) \mathcal{F}(v)(i,j), \\
\mathcal{F}(\mathcal{S}_2^\pm v)(i,j) &= (\cos z_j \pm \sqrt{-1} \sin z_j) \mathcal{F}(v)(i,j),
\end{align}
where
\begin{equation}
\label{zizj}
z_i = \frac{2\pi}{M} (i - 1), \quad z_j = \frac{2\pi}{N} (j - 1).
\end{equation}

\subsection{Discretization scheme and subproblem solutions}
\label{subsec: dis}
For the discrete analogue of $\mathbf{p}^{n+1/4}$, we first compute
\[\vert \mathbf{t}^{\top} (\theta_{\ell}) \mathbf{H}^{n}(i,j)\mathbf{t}(\theta_{\ell})\vert=\vert v_{xx}(i,j)\cos^{2}\theta_{\ell} + 2v_{xy}(i,j)\cos\theta_{\ell} \sin\theta_{\ell} + v_{yy}(i,j)\sin^{2}\theta_{\ell}\vert,\]
and
\[\mathbf{q}^{k}(i,j):\mathbf{t}(\theta_{\ell}) = q_{1}^{k}(i,j)\cos\theta_{\ell} +q_{2}^{k}(i,j)\sin\theta_{\ell}, \quad \theta_{\ell}\in\{0, \pi/4, \dots, 7\pi/4\}.\]
Then $\mathbf{q}^{k+1}$ is updated according \cref{alg:Fix}.
After $\mathbf{q}^{k+1}$ converges,
we set $\mathbf{p}^{n+1/4}=\mathbf{q}^{k
+1}$.

For the discrete analogue of $\mathbf{H}^{n+1/4}$, we compute
\begin{equation*}
    \Delta_{1}(\theta_{\ell}) = \left(1 + \left(p_{1}^{n+1/4}(i,j)\cos\theta_{\ell} + p_{2}^{n+1/4}(i,j)\sin\theta_{\ell}\right)^{2}\right)^{-1}.
\end{equation*}
Then $\mathbf{w}^{n,k+1}$ is updated according to \cref{alg:ADMM}. After convergence,
we reshape $\mathbf{w}^{n,k+1}$ to matrix form and set $\mathbf{H}^{n+1/4}=\mathrm{mat}(\mathbf{w}^{n,k+1})$.

\subsection{Computing the discrete analogue of 
$\mathbf{p}^{n+2/4}$ and $\mathbf{H}^{n+2/4}$}
\label{subsec:dis2/4}

Following equation (\ref{equ:pn2/4}), we have
\begin{equation}
    \mathbf{p}^{n+2/4}(i,j) = \max \left\{ 0, 1 - \frac{\tau \beta / \eta}{\sqrt{\left( p_1^{n+1/4}(i,j) \right)^2 + \left( p_2^{n+1/4}(i,j) \right)^2}} \right\} \mathbf{p}^{n+1/4}(i,j)
\end{equation}
and then set $\mathbf{H}^{n+2/4}(i,j)=\mathbf{H}^{n+1/4}(i,j)$.

\subsection{Computing the discrete analogue of 
$\mathbf{p}^{n+3/4}$ and $\mathbf{H}^{n+3/4}$}
\label{subsec:dis3/4}
We first compute $\mathbf{p}^{n+3/4}$ according to (\ref{cond-inital0}).
The discretization of problem (\ref{cond-inital0}) yields:
\begin{equation}
\label{dis-pH-2/3}
    -\mathrm{div}^{+}\nabla^{-}p_k^{n+3/4} + \eta p_k^{n+3/4} = \eta p_k^{n+2/4} - \mathrm{div}^{+}\mathbf{H}_k^{n+2/4} \quad \text{ for $k = 1, 2$}.
\end{equation}
We use FFT to achieve computational efficiency. Initially, we reformulate the discrete equation (\ref{dis-pH-2/3}) as:
\begin{align*}
\left[ \eta h^2 \mathcal{I} - (\mathcal{S}_1^+ - \mathcal{I}) (\mathcal{I} - \mathcal{S}_1^-) - (\mathcal{S}_2^+ - \mathcal{I}) (\mathcal{I} - \mathcal{S}_2^-) \right] p_k^{n+3/4} = g_k,
\end{align*}
where we define $g_k = \eta h^2 p_{k}^{n+2/4} - h^2 \text{div}^{-} \mathbf{H}_k^{n+2/4}$ for $k=1,2$. Applying the Fourier transform to both sides results in
\begin{align*}
a\mathcal{F}(p_{k}^{n+3/4}) = \mathcal{F}(g_{k}).
\end{align*}
The coefficient $a(i,j)$ is determined by 
\begin{align*}
a(i,j) &= \eta h^2 - (\cos z_i + \sqrt{-1} \sin z_i - 1) (1 - \cos z_i + \sqrt{-1} \sin z_i) \\
&\qquad - (\cos z_j + \sqrt{-1} \sin z_j - 1) (1 - \cos z_j + \sqrt{-1} \sin z_j) \\
&= \eta h^2 + 4 - 2 \cos z_i - 2 \cos z_j,
\end{align*}
where $z_i, z_j$ defined in (\ref{zizj}). Then the update $p_k^{n+3/4}$ is given by
\begin{align*}
p_{k}^{n+3/4} = \operatorname{Re} \left[ \mathcal{F}^{-1} \left( \frac{\mathcal{F}(g_k)}{a} \right) \right],
\end{align*}
where $\operatorname{Re}(\cdot)$ denotes the real part.
Finally, the quantity $\mathbf{H}^{n+3/4}$ is computed by 
\begin{align*}
\mathbf{H}^{n+3/4} = \nabla^{-} \mathbf{p}^{n+3/4}.
\end{align*}

\subsection{Computing the discrete analogue of $\mathbf{p}^{n+1}$ and $\mathbf{H}^{n+1}$}
\label{subsec:dis4/4}
We begin by computing $u^{n+1}$ from the discrete system corresponding to (\ref{cond-inita1}), given by
\begin{equation}
\label{dis-pH-n1}
    -\eta \text{div}^{-} \nabla^{+} u^{n+1} + \gamma\tau u^{n+1} = \gamma \tau f - \eta \text{div}^{-} \mathbf{p}^{n+3/4},
\end{equation}
which can be equivalently expressed in the form as
\begin{equation*}
    \left[ \gamma\tau h^2 \mathcal{I} - \eta(\mathcal{I} - \mathcal{S}_1^-)(\mathcal{S}_1^+ - \mathcal{I}) - \eta(\mathcal{I} - \mathcal{S}_2^-)(\mathcal{S}_2^+ - \mathcal{I}) \right] u^{n+1} = g
\end{equation*}
with $g = \gamma\tau h^2 f - \eta h^2 \text{div}^{-} \mathbf{p}^{n+3/4}$.
Taking the Fourier transform on both sides yields
\begin{align*}
b\mathcal{F}(u^{n+1}) = \mathcal{F}(g),
\end{align*}
with $b = \gamma\tau h^2 + 4\eta - 2\eta \cos z_i - 2\eta \cos z_j$, where $z_{i},z_{j}$ are defined in (\ref{zizj}).

The solution $u^{n+1}$ is then obtained by
\begin{align*}
u^{n+1} = \operatorname{Re} \left[ \mathcal{F}^{-1} \left( \frac{\mathcal{F}(g)}{b} \right) \right],
\end{align*}
where $\mathrm{Re}(\cdot)$ denotes the real part. Finally, we set
\begin{equation*}
    \mathbf{p}^{n+1} = \nabla^{+} u^{n+1}, \mathbf{H}^{n+1} =\mathbf{H}^{n+3/4}.
\end{equation*}

\subsection{On the discrete analogue of $(\mathbf{p}_0, \mathbf{H}_0)$}
\label{subsec:disint}
For the initialization in (\ref{inital1}), we simply set
\begin{equation}
\mathbf{p}_{0} = \nabla^{+} f, ~ \mathbf{H}_{0}= \nabla^{-} \mathbf{p}_{0}.
\end{equation}
For the alternative initialization in (\ref{inital2}), we first compute $u_0$ by solving the discrete analogue of (\ref{cond-inital2}),  in the same manner as $u^{n+1}$:
\begin{equation*}
    u_0 = \operatorname{Re} \left[ \mathcal{F}^{- 1} \left( \frac{\mathcal{F}(f)}{c} \right) \right],
\end{equation*}
where $c = h^2 + 4\varepsilon - 2\varepsilon \cos z_i - 2\varepsilon \cos z_j.$ We then define
\begin{equation*}
    \mathbf{p}_{0} = \nabla^{+} u_{0}, \quad \mbox{and} \quad \mathbf{H}_{0} = \nabla^{-} \mathbf{p}_{0}.
\end{equation*}
All subproblems in our numerical scheme can be solved either in closed form or efficiently using fast numerical methods. 
Specifically, the shrinkage updates for $\mathbf{p}^{n+2/4}$ and the gradient-based update of $\mathbf{H}^{n+2/4}$ admit closed-form solutions.
The linear PDE subproblems, including those governing $\mathbf{p}^{n+3/4}$, $\mathbf{H}^{n+3/4}$, $\mathbf{p}^{n+1}$, and $\mathbf{H}^{n+1}$, are efficiently solved using the FFT in the frequency domain. For the nonconvex $\mathbf{H}^{n+1/4}$ subproblem, although there is no closed-form solution, we employ a variable splitting technique with auxiliary variables and approximate the solution through a single iteration of the ADMM-based algorithm.

\begin{figure}[t]
	\centering
	\begin{tabular}{c@{\hspace{2pt}}c@{\hspace{2pt}}c@{\hspace{2pt}}c}\\
        \includegraphics[width=0.18\textwidth]{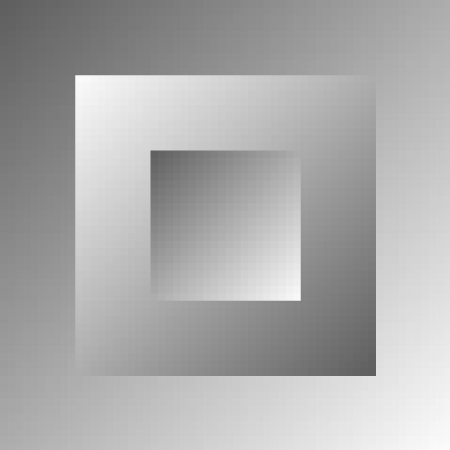}&
        \includegraphics[width=0.18\textwidth]{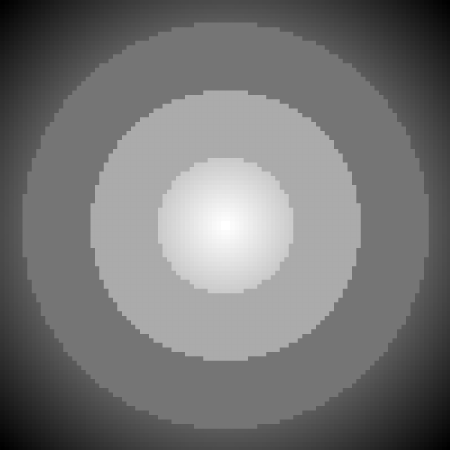}&
	\includegraphics[width=0.18\textwidth] 
       {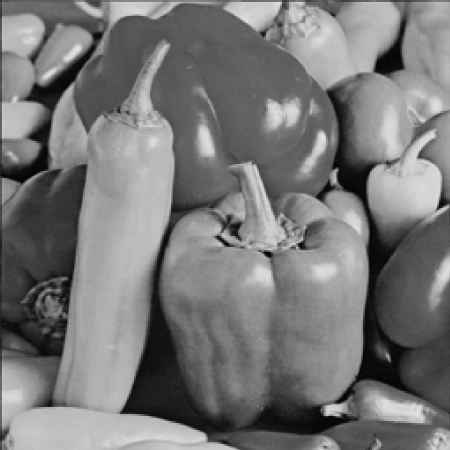}&
       \includegraphics[width=0.18\textwidth] 
       {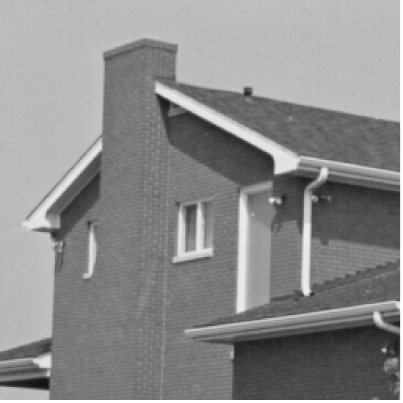}\\
        (a) & (b)  & (c) & (d)\\ 
        \includegraphics[width=0.18\textwidth]{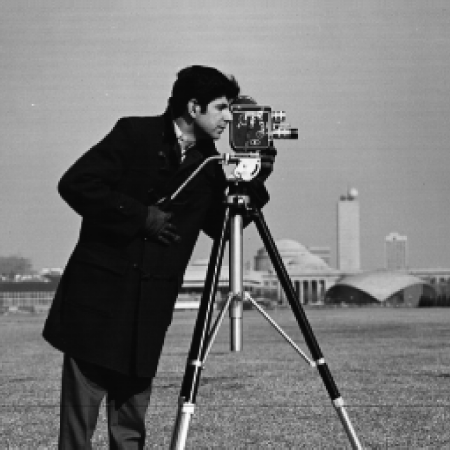}&
        \includegraphics[width=0.18\textwidth]{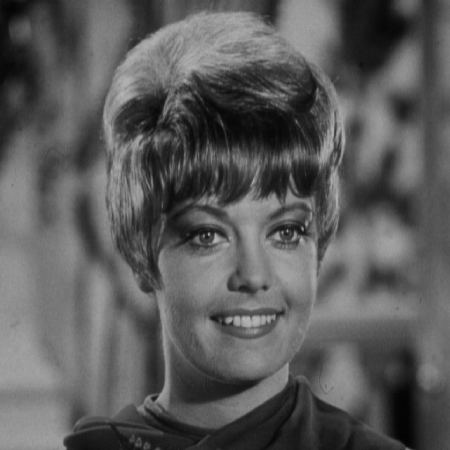}&
        \includegraphics[width=0.18\textwidth]{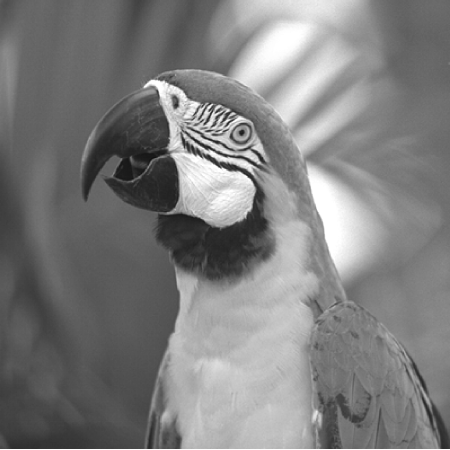}&
        \includegraphics[width=0.18\textwidth]{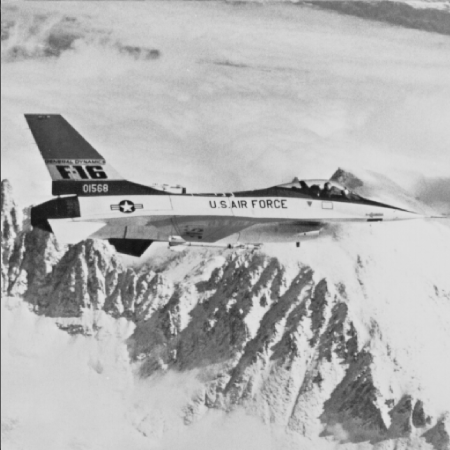}\\
        (e) & (f)  & (g) & (h)\\	
        \end{tabular}
   \caption{Testing images. Synthetic image: (a) Square ($60 \times 60$); (b) Rings  ($100\times 100$). Natural image: (c) Peppers ($256 \times 256$); 
   (d) House ($256 \times 256$);
   (e) Cameraman ($256 \times 256$); (f) Zelda ($512 \times 512$); (g) Parrot ($512 \times 512$); (h) Plane ($512 \times 512$).}
   \label{GTimages}
\end{figure}

\section{Numerical experiments}
\label{sec:experiments}

We demonstrate the effectiveness of the proposed method through several experiments on surface smoothing and image denoising. All experiments are implemented in MATLAB R2021a on a laptop equipped with an AMD Ryzen 7 6800H processor (3.20 GHz) and 16 GB RAM. 

The test images are presented in \Cref{GTimages}, with panels (a) and (b) showing synthetic smoothed images and panels (c)–(h) displaying natural images. All grayscale images are normalized to the range $[0, 1]$, and a mesh size $h = 1$ is used throughout. Our method involves several model and algorithmic parameters. The model parameters include the regularization weights $\alpha$, $\beta$ and $\gamma$, all of which are noise-level sensitive and specified in each experiment. 
The algorithmic parameters associated with \Cref{alg:osm} include the evolution parameter $\eta$, which is fixed to $\eta = 1$ in all experiments, the time step $\tau$ (set in the numerical experiments) in the operator-splitting scheme and the stopping tolerance $\epsilon= 10^{-5}$. 
In the first subproblem, $\mathbf{p}^{n+1/4}$ is solved via \Cref{alg:Fix} with $\rho_1 = 0.8$ and a tolerance of $10^{-5}$, while $\mathbf{H}^{n+1/4}$ is updated using \Cref{alg:ADMM} with $\rho_2 = 0.5$, a tolerance of $10^{-5}$, and $I_{\max} = 1$. 
The other subproblems have closed-form solutions and do not require additional parameters. 

\begin{figure}[t]
	\centering
	\begin{tabular}{c@{\hspace{2pt}}c@{\hspace{2pt}}c}
	(a)  & (b)  & (c) \\  
        \includegraphics[width=0.3\textwidth]{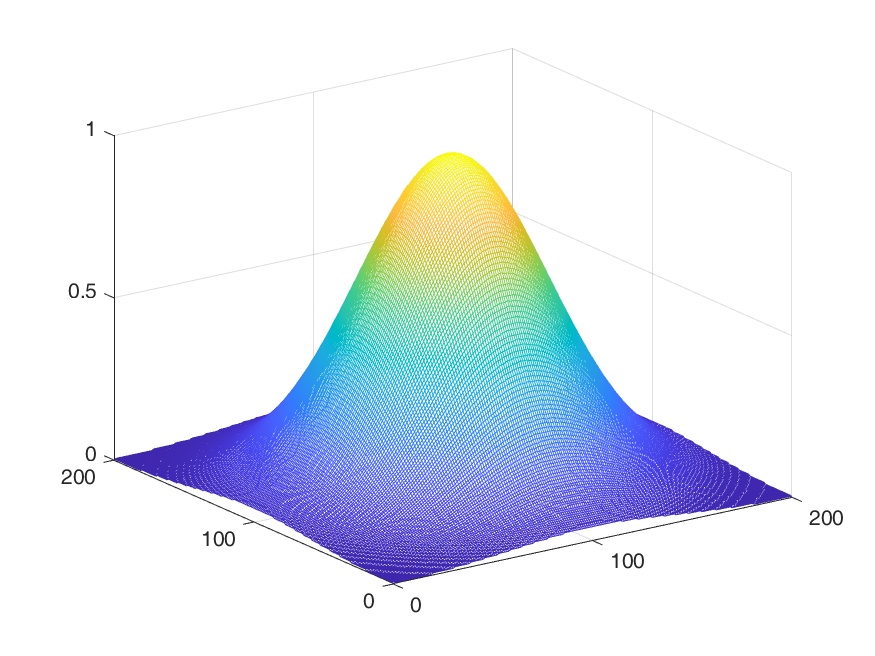}&
        \includegraphics[width=0.3\textwidth]{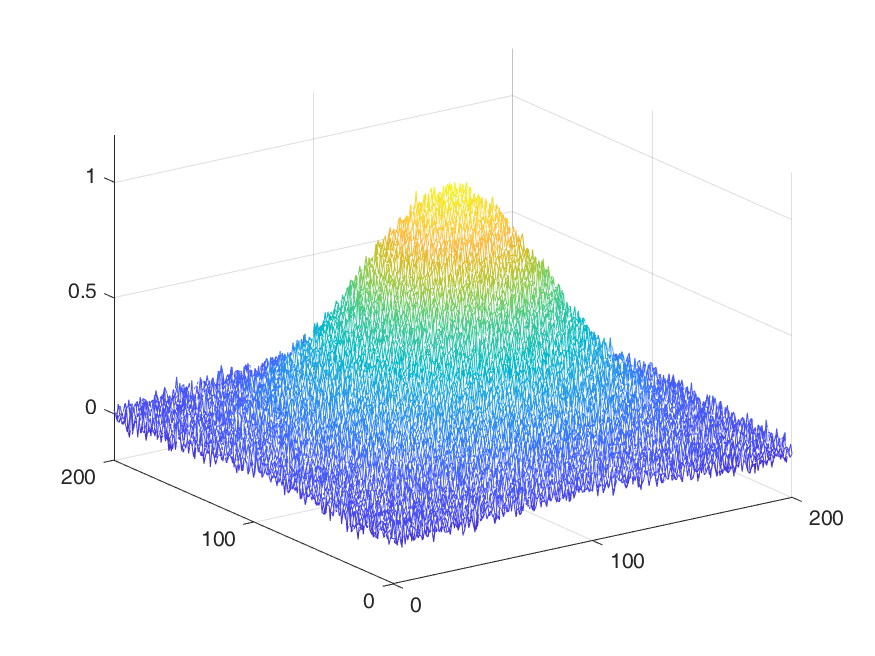}&
	\includegraphics[width=0.3\textwidth]{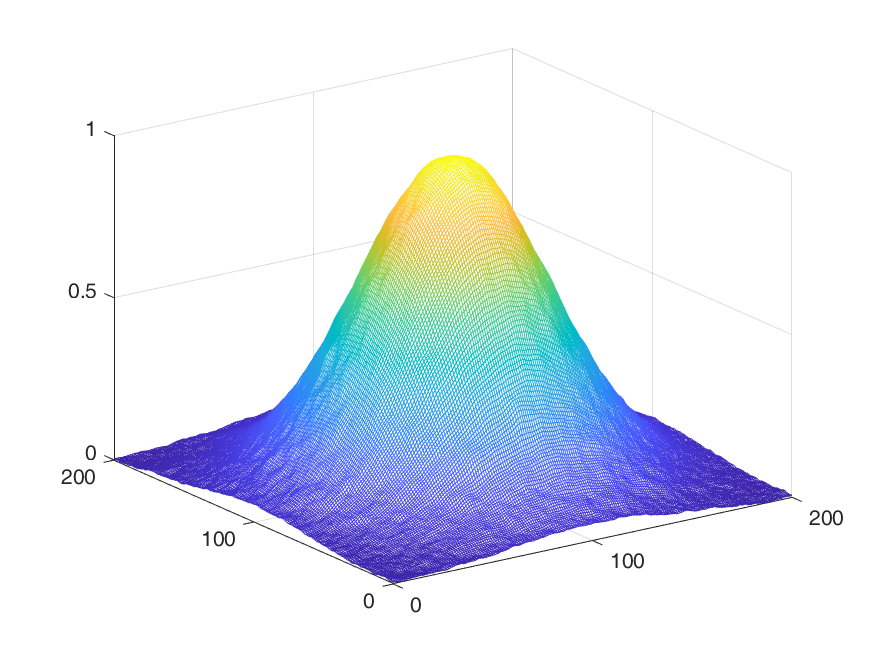}\\
        \includegraphics[width=0.3\textwidth]{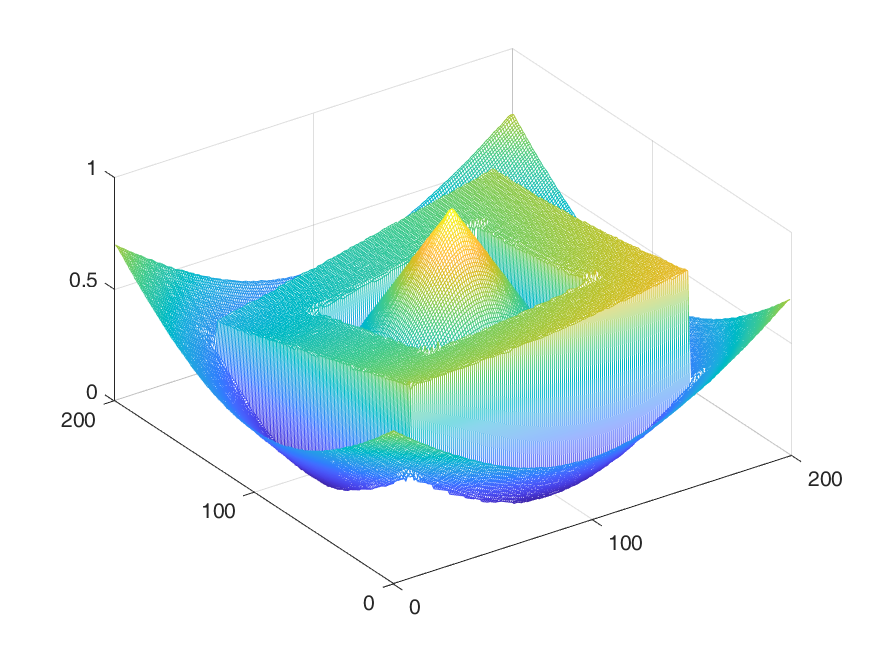}&
        \includegraphics[width=0.3\textwidth]{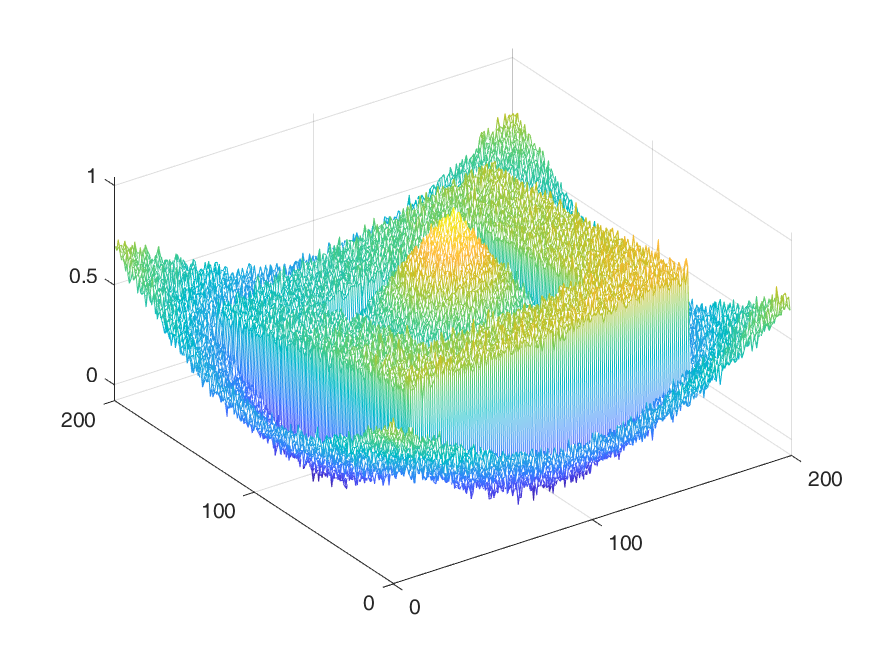}&
        \includegraphics[width=0.3\textwidth]{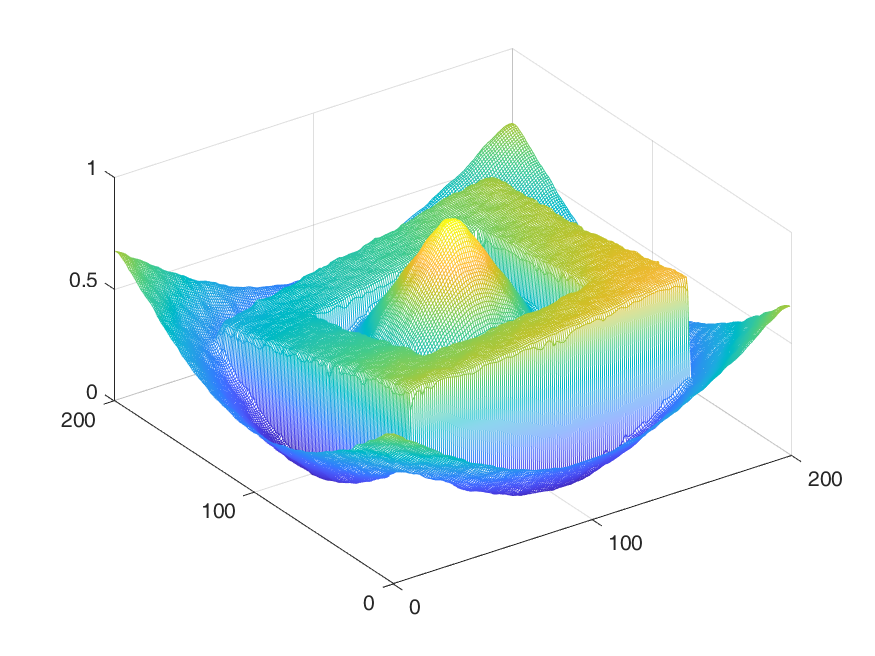}\\
        \end{tabular}
   \caption{The image surfaces of the clean images and reconstructed images by our TNC model. (a) Clean surfaces. (b) Noisy surfaces with $\sigma = 10^{-3}$. (c)
Smoothed surfaces by the proposed model with $\alpha = 0.1$, $\beta = 0.4$, $\gamma=8$, $\tau=0.01$.}
   \label{surface_smoothing1}
\end{figure}

\begin{figure}[t]
	\centering
	\begin{tabular}{c@{\hspace{2pt}}c@{\hspace{2pt}}c@{\hspace{2pt}}c}
    (a)  & (b)  & (c) & (d)\\ 
\includegraphics[width=0.235\textwidth]{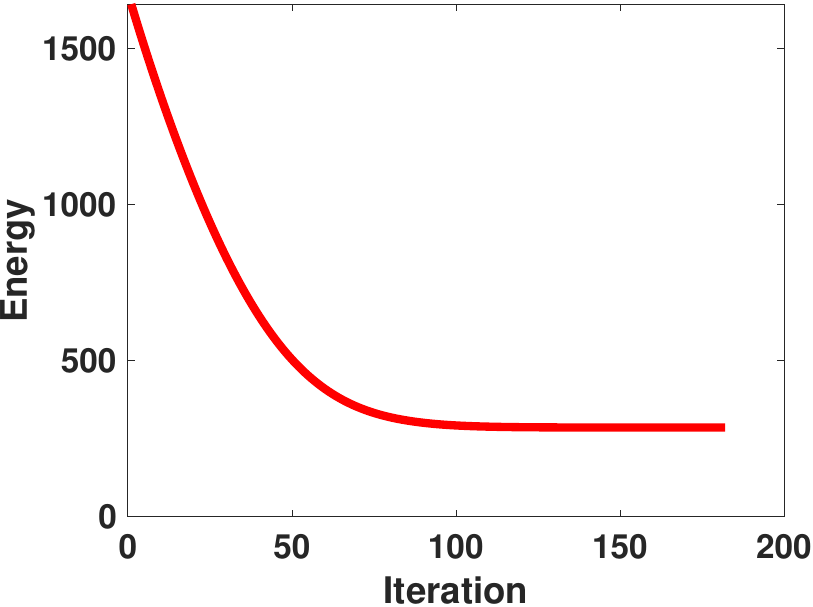}&
\includegraphics[width=0.235\textwidth]{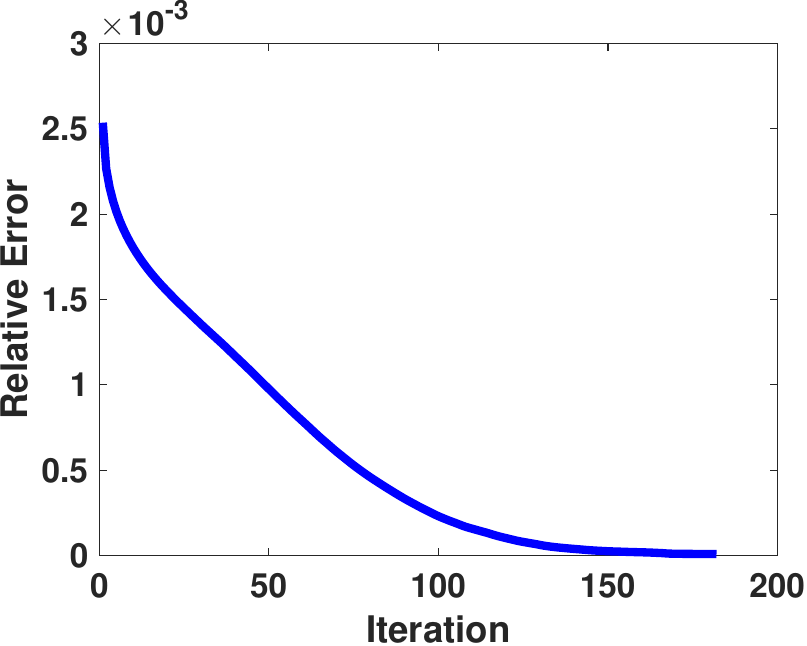}&
\includegraphics[width=0.235\textwidth]{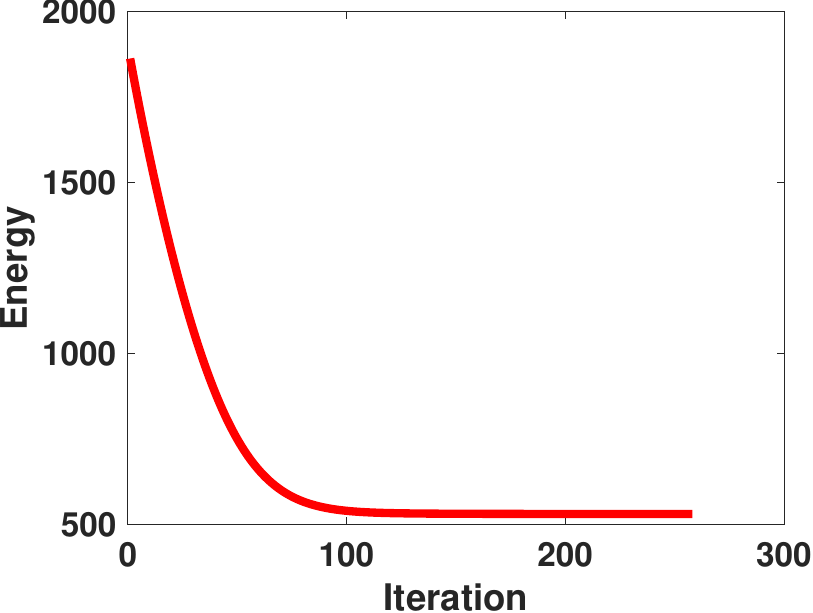}&
\includegraphics[width=0.235\textwidth]{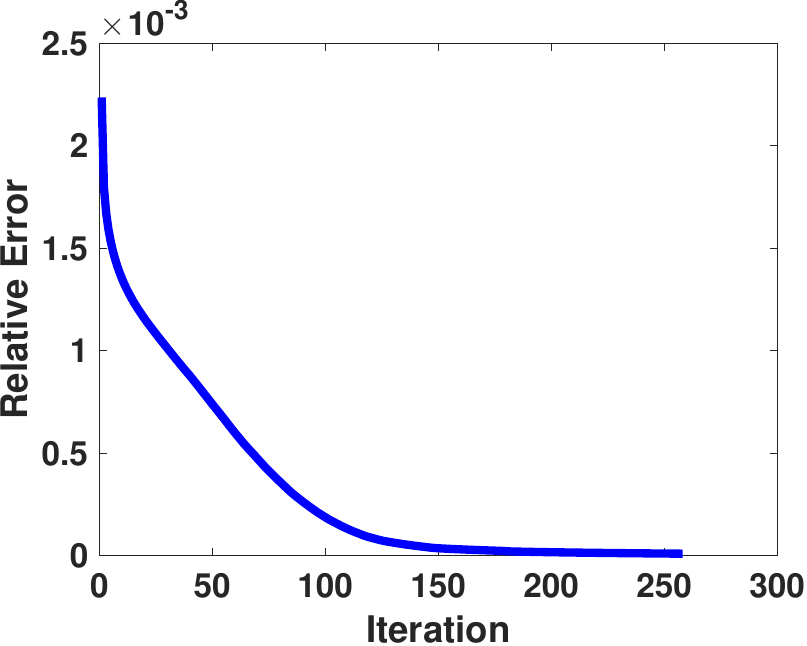}\\
   \end{tabular}
   \caption{Surface smoothing results of Figure \ref{surface_smoothing1}: (a) Energy history for the top row; (b) Relative error history for the top row; (c) Energy history for the bottom row; (d) Relative error history for the bottom row.}
   \label{surface_smoothing2}
\end{figure}

\subsection{Numerical convergence analysis}
\label{subsec:surface}
We demonstrate the performance of the proposed TNC model (\ref{model1}) on surface smoothing problems. We examine the clean surfaces depicted in Figure \ref{surface_smoothing1}(a). The noisy surfaces, shown in Figure \ref{surface_smoothing1}(b), are generated by adding Gaussian noise with \(\sigma = 10^{-3}\). In our algorithm, we set \(\alpha = 0.1\), \(\beta = 0.4\), \(\gamma = 8\), and \(\tau = 0.01\). The resulting smoothed surfaces are presented in Figure \ref{surface_smoothing1}(c). The differences between the smoothed surfaces \(u\) and the clean surfaces \(f^{\ast}\) are illustrated in Figure \ref{surface_smoothing2}(a). The smoothed surfaces closely resemble the clean surfaces, with only minor discrepancies. To demonstrate the efficiency of the proposed method, we provide the histories of the energy and relative error with respect to the number of iterations in Figures \ref{surface_smoothing2}(b) and (c), respectively. In both examples, the energy reaches its minimum after approximately 100 iterations, and sublinear convergence is observed for the relative error.

\begin{figure}[]
	\centering
    \begin{minipage}{\textwidth}
    \centering
        \begin{tabular}{c@{\hspace{2pt}}c@{\hspace{2pt}}c@{\hspace{2pt}}c@{\hspace{2pt}}c}
        (a) $\mathrm{Noisy}$ & (b) $\mathrm{EE}$ & (c) $\mathrm{Zoom}$ & (d) $\mathrm{TNC}$ & (e) $\mathrm{Zoom}$\\  
        \includegraphics[width=0.18\textwidth]{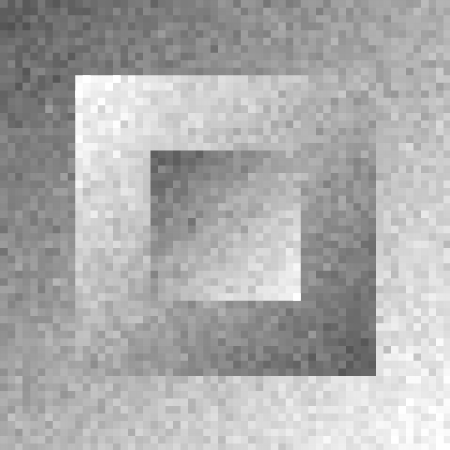}&
        \includegraphics[width=0.18\textwidth]{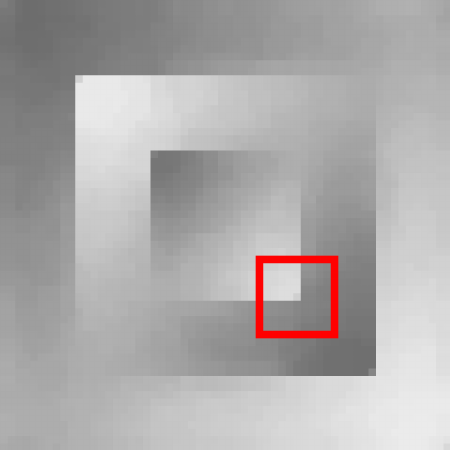}&
        \includegraphics[width=0.18\textwidth]{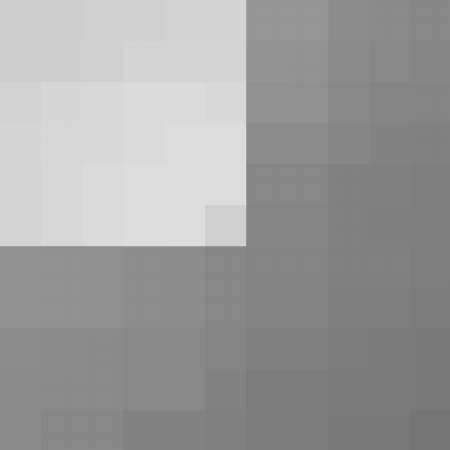}&
        \includegraphics[width=0.18\textwidth]{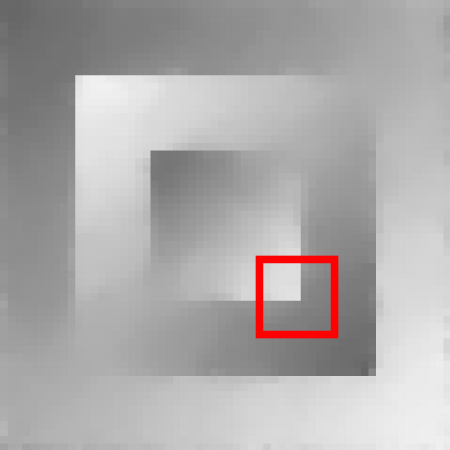}&
        \includegraphics[width=0.18\textwidth]{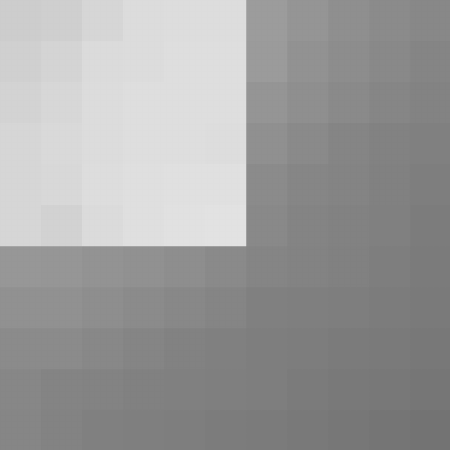}\\
        \includegraphics[width=0.18\textwidth]{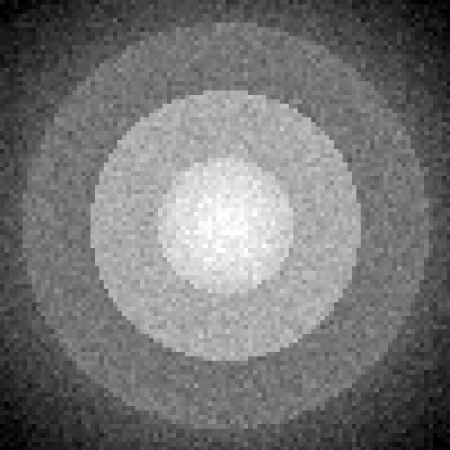}&
        \includegraphics[width=0.18\textwidth]{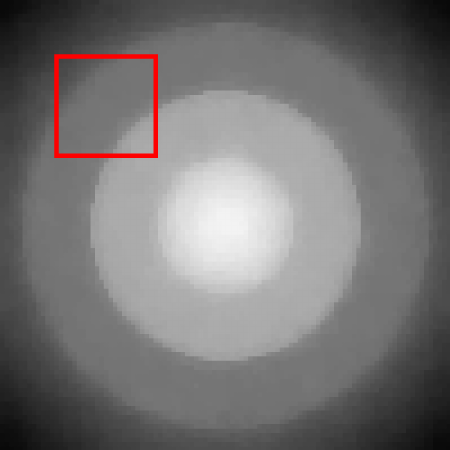}&
        \includegraphics[width=0.18\textwidth]{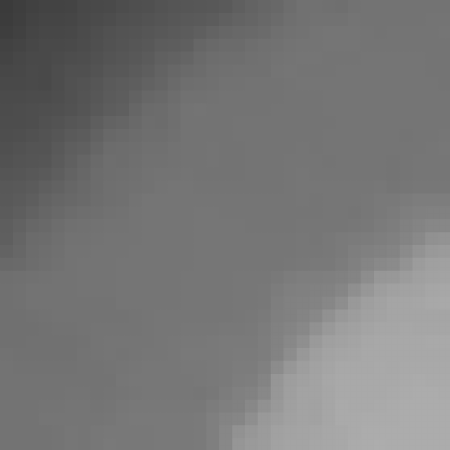}&
        \includegraphics[width=0.18\textwidth]{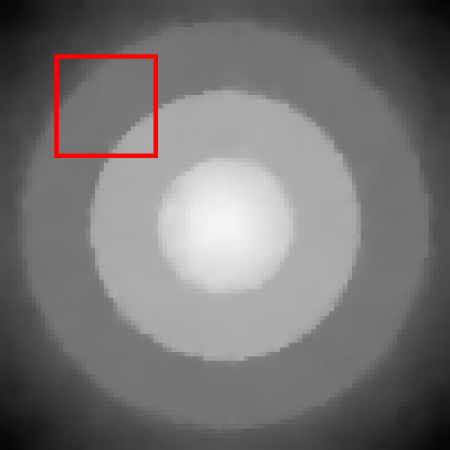}&
        \includegraphics[width=0.18\textwidth]{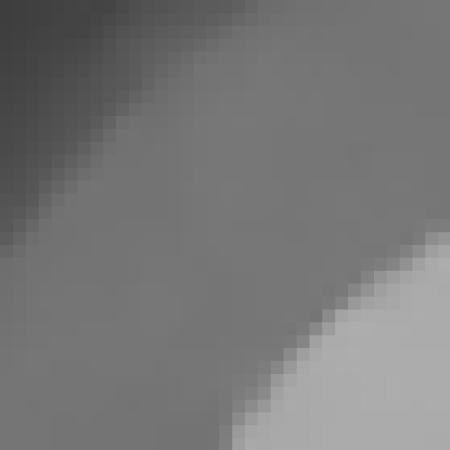}\\
        \end{tabular}
        \caption{The denoising results on ``Square'' and ``Rings'' (top to bottom) using the Euler’s elastica and our TNC model.}
        \label{synthetic_smooth1}
        \end{minipage}

        \bigskip 
        
        \begin{minipage}{\textwidth}
        \begin{tabular}{c@{\hspace{2pt}}c@{\hspace{2pt}}c}
        (a) $\mathrm{Clean}$ & (b) $\mathrm{EE}$ & (c) $\mathrm{TNC}$\\  
        \includegraphics[width=0.30\textwidth]{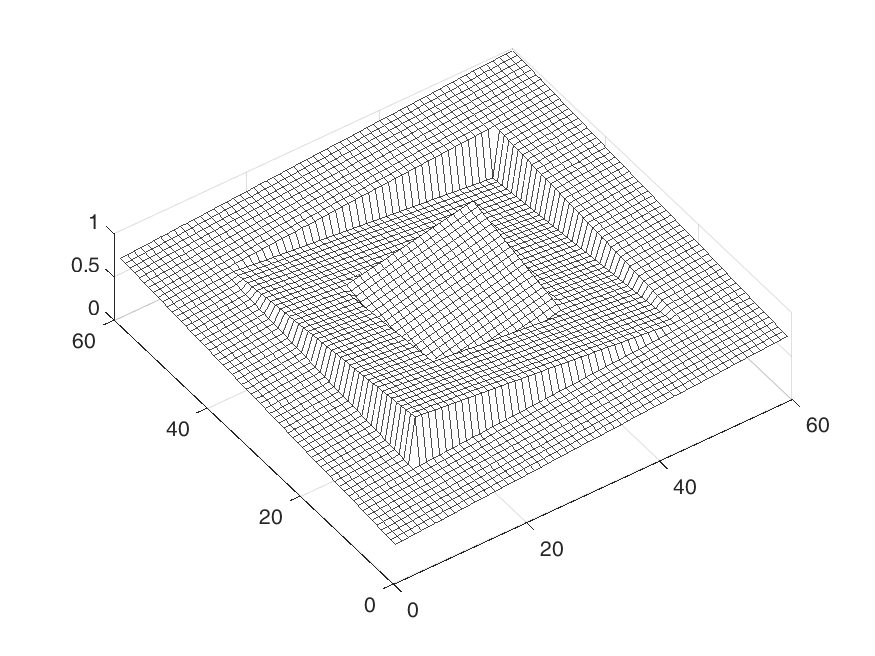}&
        \includegraphics[width=0.30\textwidth]     {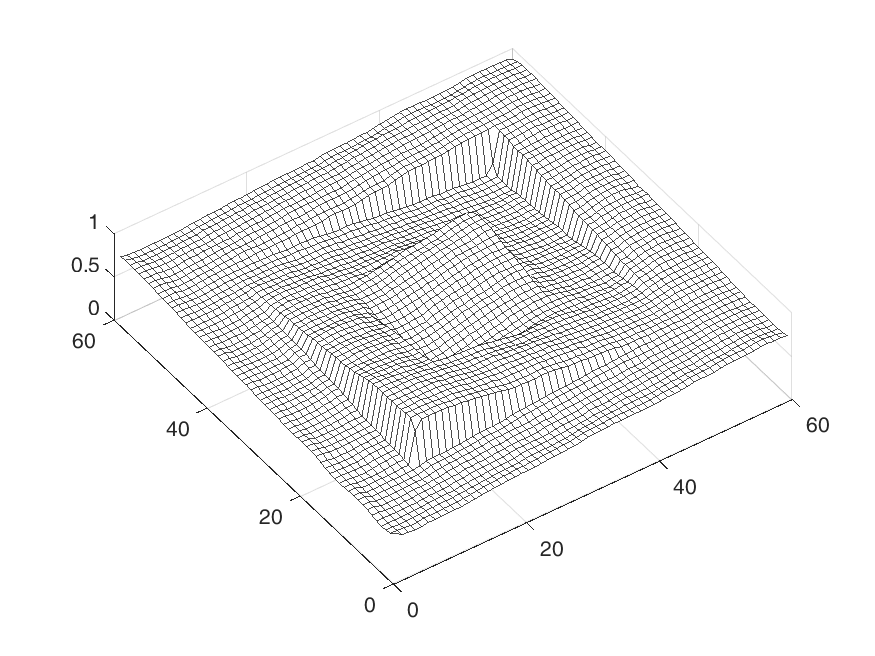}&
        \includegraphics[width=0.30\textwidth]{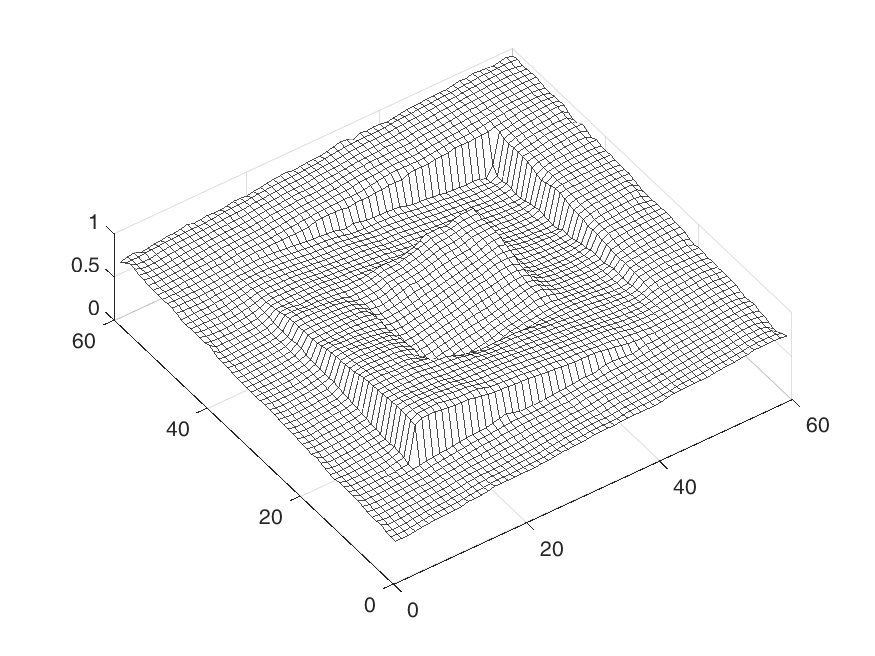}
        \\
        \includegraphics[width=0.30\textwidth]{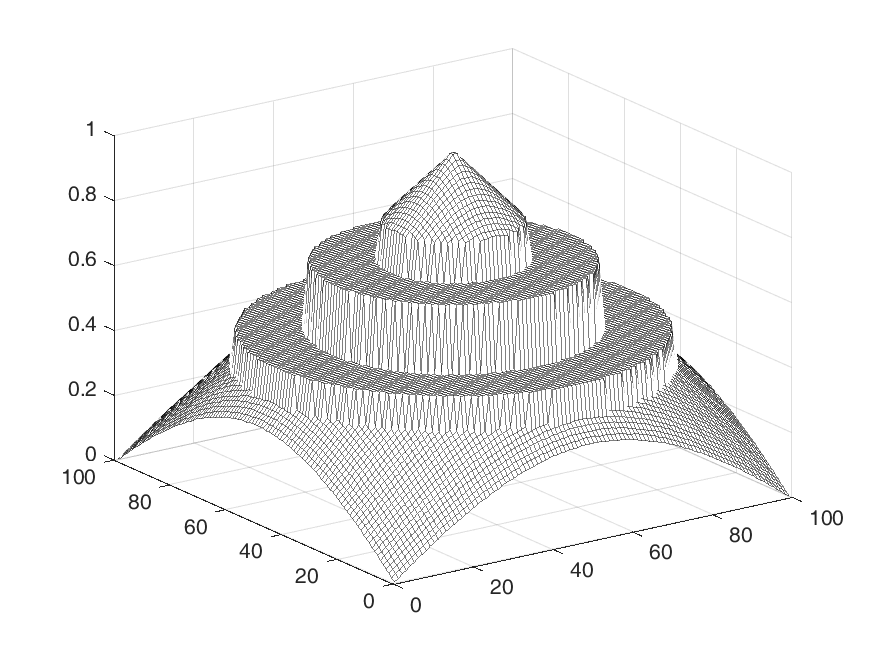}&
        \includegraphics[width=0.30\textwidth]{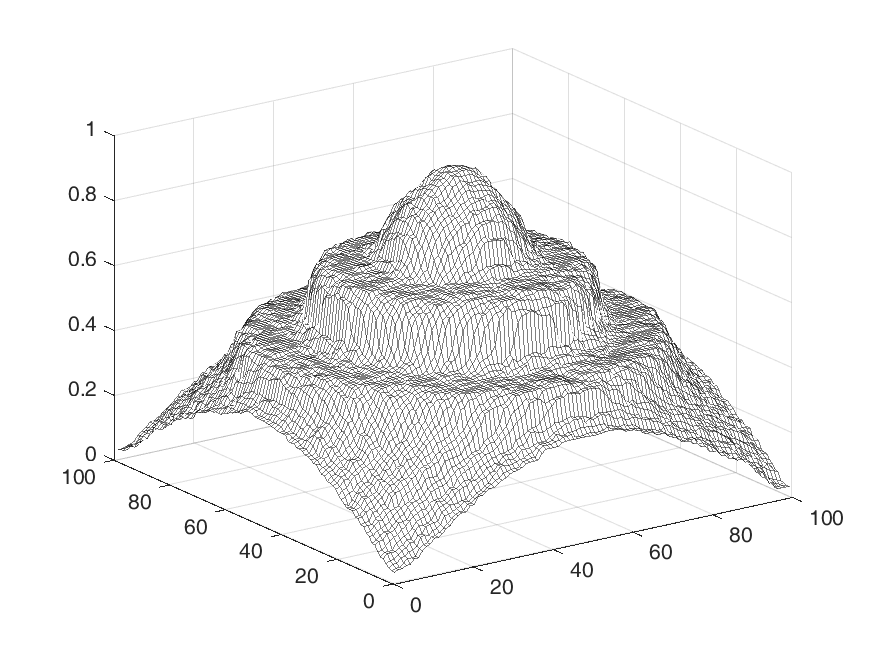}&
        \includegraphics[width=0.30\textwidth]{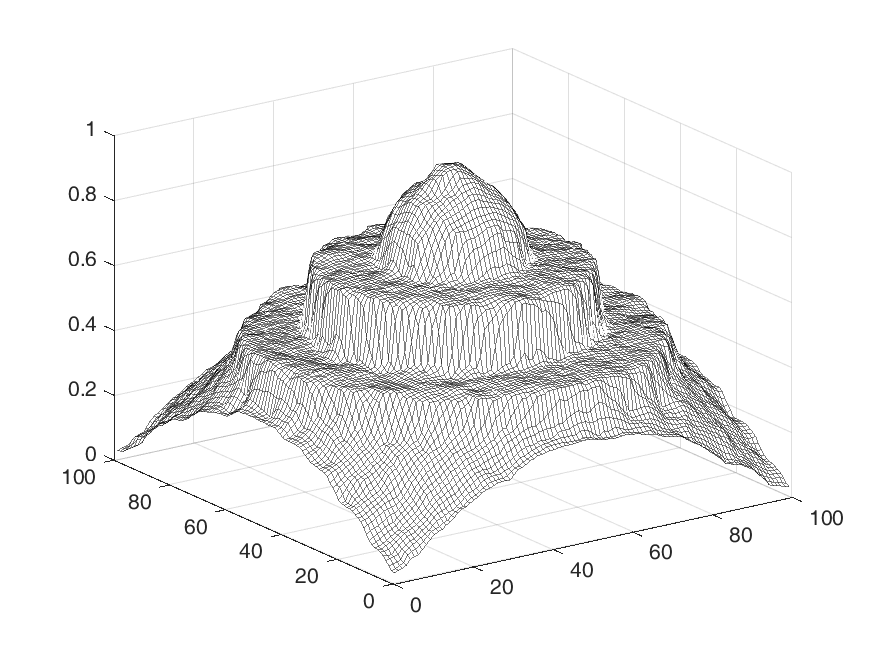}
        \end{tabular}
        \caption{The image surfaces of the clean images and reconstructed images by the Euler’s elastica and our TNC model.}
        \label{synthetic_smooth2}
        \end{minipage}

        \bigskip
        
        \begin{minipage}{\textwidth}
        \centering
        \begin{tabular}{c|c|c|c|c}
        \toprule
        \multicolumn{1}{c|}{Figure \ref{synthetic_smooth2}} & \multicolumn{2}{c|}{Square} & \multicolumn{2}{c}{Rings} \\
        \hline       
        Methods & EE & TNC & EE & TNC\\
        \hline
        $\|u-f^*\|_1$ & 69.32 & 49.86 &144.66 &95.32\\
        \hline
        $\|u-f^*\|_{\infty}$ & 0.1700 & 0.1343 & 0.1221 & 0.1187\\
        \bottomrule
        \end{tabular}
        \captionof{table}{Comparison of Euler’s elastica and our TNC model for surface smoothing on synthetic images in \Cref{synthetic_smooth2} in terms of $\ell_1$ and $\ell_\infty$ errors.}
        \label{tab.syn_diff}
        \end{minipage}
        
\end{figure}

\subsection{Synthetic images denoising}
In this section, we evaluate the proposed model against the Euler’s elastica (EE) model \cite{tai2011fast} through comparisons on smoothed synthetic images and analyses of the corresponding reconstructed surfaces. Two synthetic images, ``Square'' and ``Rings'', were corrupted with zero-mean Gaussian noise with a standard deviation of 10/255 for this evaluation. To ensure a fair comparison, the parameters for the comparative methods were selected as recommended in the respective papers: 
\begin{itemize}
    \item Euler’s elastica \cite{tai2011fast}:
    $a=1$, $b=10$, $r_{1}=1$, $r_{2}=2 \times 10^{2}$, and $r_{4}=5 \times 10^{2}$ ($\epsilon=1.3 \times 10^{-3}$ for the ``Square'' image, and $\epsilon=10^{-2}$ for the ``Rings'' image).
    \item Our model: $\alpha=0.1$, $\beta=0.4$, $\gamma=10$.
\end{itemize}
The noisy images are displayed in \cref{synthetic_smooth1}(a). The results obtained using the Euler's elastica model and our proposed model are shown in columns (b) and (d) of \cref{synthetic_smooth1}, respectively. A visual inspection of these results clearly demonstrates that our model more effectively preserves critical image structures, such as edges and sharp corners, compared to the Euler's elastica model. This observation is further supported by the surface visualizations in \cref{synthetic_smooth2}, which compare the ground truth surfaces with those reconstructed by both models. 

To quantitatively evaluate the reconstruction accuracy, we report the \(\ell_1\) and \(\ell_\infty\) errors between the restored surface \(u\) and the reference \(f^*\) in \cref{tab.syn_diff}. The proposed model consistently achieves lower errors than the Euler's elastica model. These surface plots align with the quantitative results, demonstrating the effectiveness of our curvature regularization in preserving edge fidelity and corner sharpness, outperforming the curvature formulation used in Euler’s elastica.

\begin{figure}[]
	\centering
	\begin{tabular}{c@{\hspace{2pt}}c@{\hspace{2pt}}c@{\hspace{2pt}}c@{\hspace{2pt}}c@{\hspace{2pt}}c}
	(a) $\mathrm{TV}$ & (b) $\mathrm{EE}$ & (c) $\mathrm{MC}$ & (d) $\mathrm{TAC}$ & (e) $\mathrm{GCTV}$ & (f) $\mathrm{TNC}$\\  
        \includegraphics[width=0.15\textwidth]{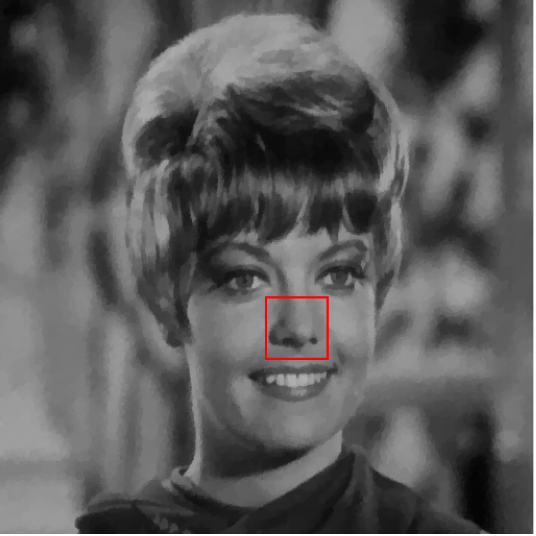}&
        \includegraphics[width=0.15\textwidth]{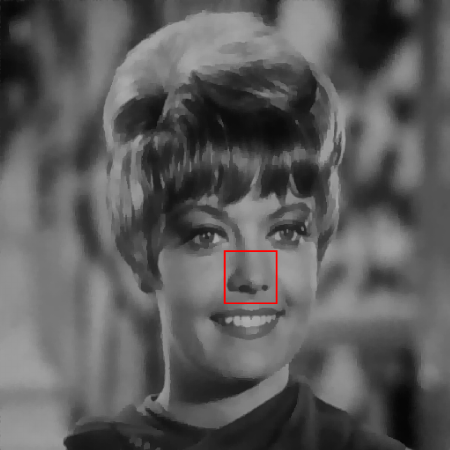}&
	\includegraphics[width=0.15\textwidth] 
        {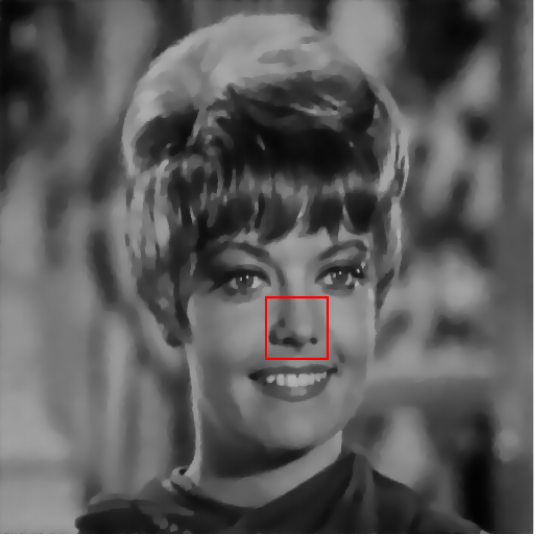}&
        \includegraphics[width=0.15\textwidth]{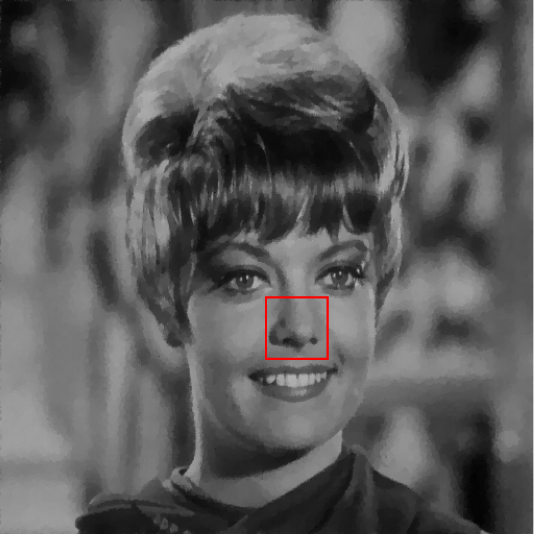}&
        \includegraphics[width=0.15\textwidth]{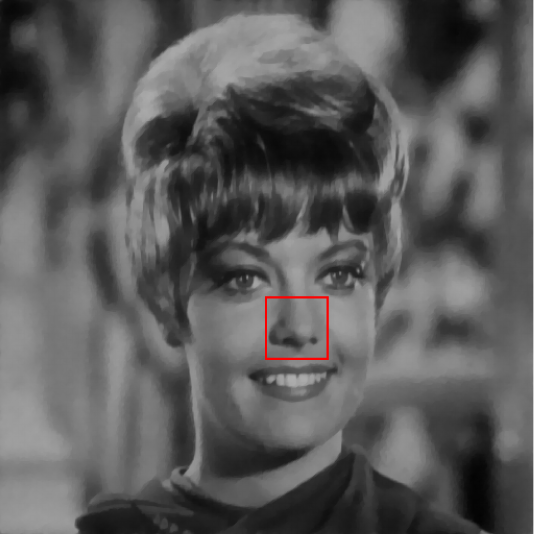}&
        \includegraphics[width=0.15\textwidth]
        {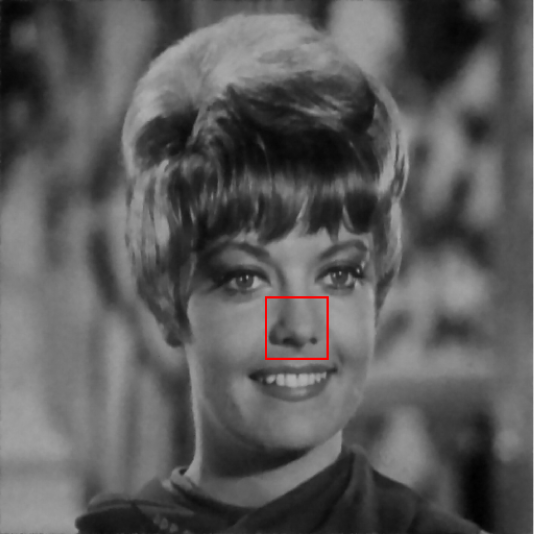}\\
        \includegraphics[width=0.15\textwidth]{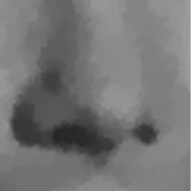}&
        \includegraphics[width=0.15\textwidth]{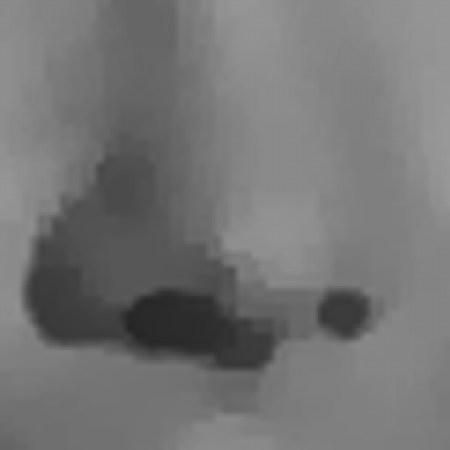}&
	\includegraphics[width=0.15\textwidth] 
        {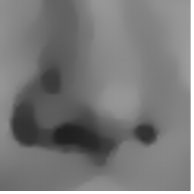}&
        \includegraphics[width=0.15\textwidth]{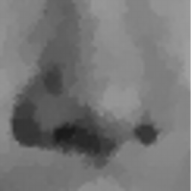}&
        \includegraphics[width=0.15\textwidth]{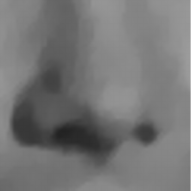}&
        \includegraphics[width=0.15\textwidth]  {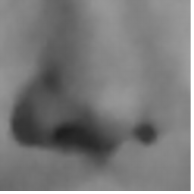}
        \end{tabular}
   \caption{The denoising results of ``Zelda'' ($\sigma=10/255$). (a) TV: $\lambda=1.2$. (b) EE: $a=1$, $b=5$. (c) MC: $r_{1}=10, r_{2}=10, r_{3}=10^{4}, r_{4}=10^{4}$. (d) TAC: $a=1$, $b=5$, (e) GCTV: $\alpha=0.1$, $\beta=0.5$. (f) TNC:  $\alpha=0.1$, $\beta=0.4$, $\gamma=12$.}
   \label{fig:Zelda}
\end{figure}

\begin{figure}[]
	\centering
	\begin{tabular}{c@{\hspace{2pt}}c@{\hspace{2pt}}c@{\hspace{2pt}}c@{\hspace{2pt}}c@{\hspace{2pt}}c}
	(a) $\mathrm{TV}$ & (b) $\mathrm{EE}$ & (c) $\mathrm{MC}$ & (d) $\mathrm{TAC}$ & (e) $\mathrm{GCTV}$ & (f) $\mathrm{TNC}$\\  
        \includegraphics[width=0.15\textwidth]{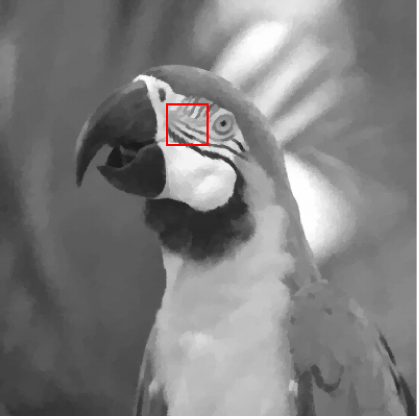}&
        \includegraphics[width=0.15\textwidth]{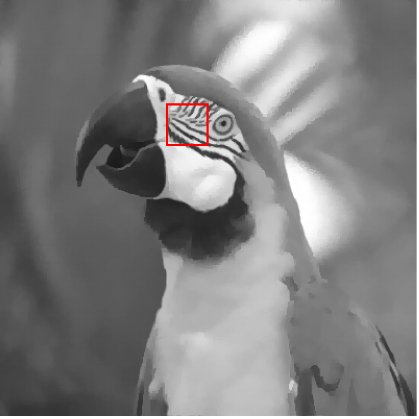}&
	\includegraphics[width=0.15\textwidth] 
        {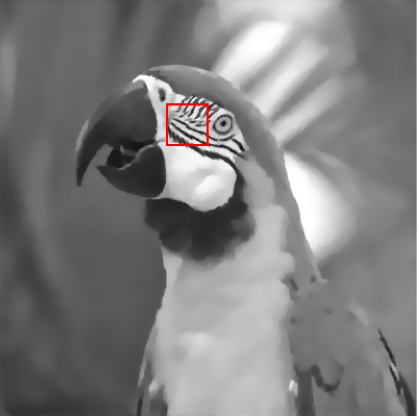}&
        \includegraphics[width=0.15\textwidth]{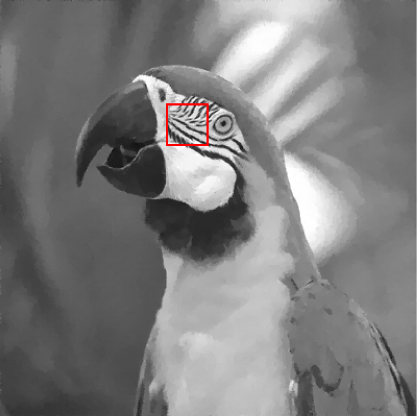}&
        \includegraphics[width=0.15\textwidth]{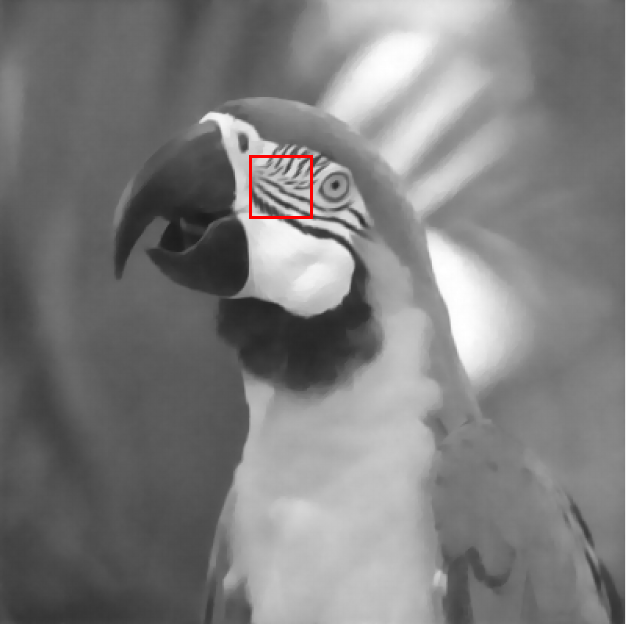}&
        \includegraphics[width=0.15\textwidth]
        {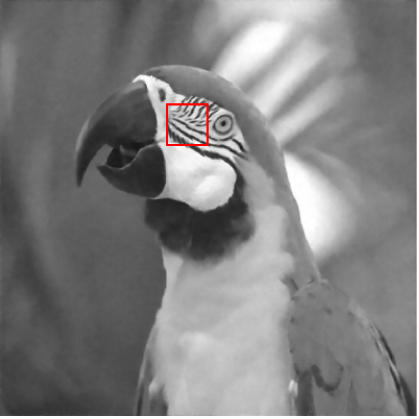}\\
        \includegraphics[width=0.15\textwidth]{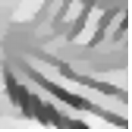}&
        \includegraphics[width=0.15\textwidth]{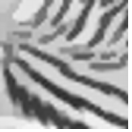}&
	\includegraphics[width=0.15\textwidth] 
        {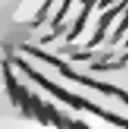}&
        \includegraphics[width=0.15\textwidth]{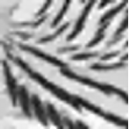}&
        \includegraphics[width=0.15\textwidth]{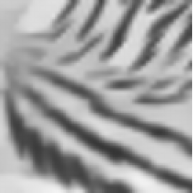}&
        \includegraphics[width=0.15\textwidth]  {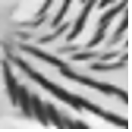}
        \end{tabular}
   \caption{The denoising results of ``Parrot'' ($\sigma=10/255$), where all methods employ the same parameters as those specified in Figure \ref{fig:Zelda}.}
   \label{fig:Parrot}
\end{figure}

\begin{figure}[]
	\centering
	\begin{tabular}{c@{\hspace{2pt}}c@{\hspace{2pt}}c@{\hspace{2pt}}c@{\hspace{2pt}}c@{\hspace{2pt}}c}
	(a) $\mathrm{TV}$ & (b) $\mathrm{EE}$ & (c) $\mathrm{MC}$ & (d) $\mathrm{TAC}$ & (e) $\mathrm{GCTV}$ & (f) $\mathrm{TNC}$\\  
        \includegraphics[width=0.15\textwidth]{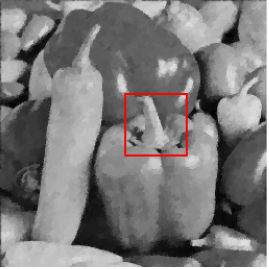}&
        \includegraphics[width=0.15\textwidth]{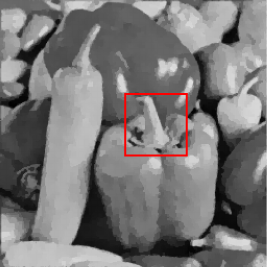}&
	\includegraphics[width=0.15\textwidth] 
        {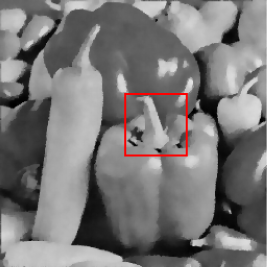}&
        \includegraphics[width=0.15\textwidth]{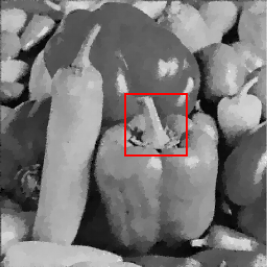}&
        \includegraphics[width=0.15\textwidth]{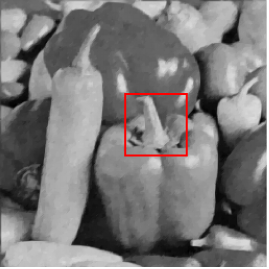}&
        \includegraphics[width=0.15\textwidth]
        {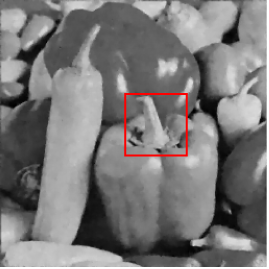}\\
        \includegraphics[width=0.15\textwidth]{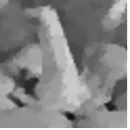}&
        \includegraphics[width=0.15\textwidth]{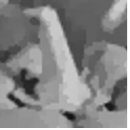}&
	\includegraphics[width=0.15\textwidth] 
        {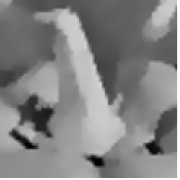}&
        \includegraphics[width=0.15\textwidth]{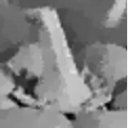}&
        \includegraphics[width=0.15\textwidth]{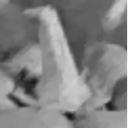}&
        \includegraphics[width=0.15\textwidth]{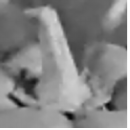}
        \end{tabular}
   \caption{The denoising results of ``Peppers'' ($\sigma=20/255$), where the parameters are selected as (a) TV: $\lambda=1.3$; (b) EE: $a=1$, $b=10$; (c) MC: $r_{1}=40, r_{2}=40, r_{3}=10^{5}, r_{4}=1.5 \cdot 10^{5}$; (d) TAC: $a=1$, $b=5$; (e) GCTV: $\alpha=0.1$, $\beta=0.5$; (f) TNC:  $\alpha=0.1$, $\beta=0.4$, $\gamma=10$.
   }
   \label{fig:Peppers}
\end{figure}
\begin{figure}[t]
	\centering
	\begin{tabular}{c@{\hspace{2pt}}c@{\hspace{2pt}}c@{\hspace{2pt}}c@{\hspace{2pt}}c@{\hspace{2pt}}c}
	(a) $\mathrm{TV}$ & (b) $\mathrm{EE}$ & (c) $\mathrm{MC}$ & (d) $\mathrm{TAC}$ & (e) $\mathrm{GCTV}$ & (f) $\mathrm{TNC}$\\  
        \includegraphics[width=0.15\textwidth]{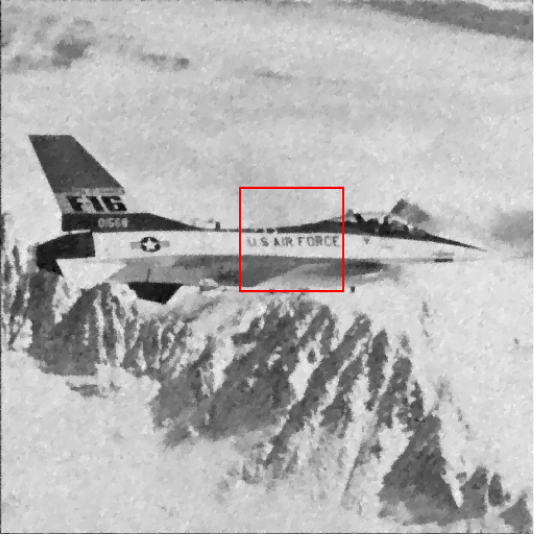}&
        \includegraphics[width=0.15\textwidth]{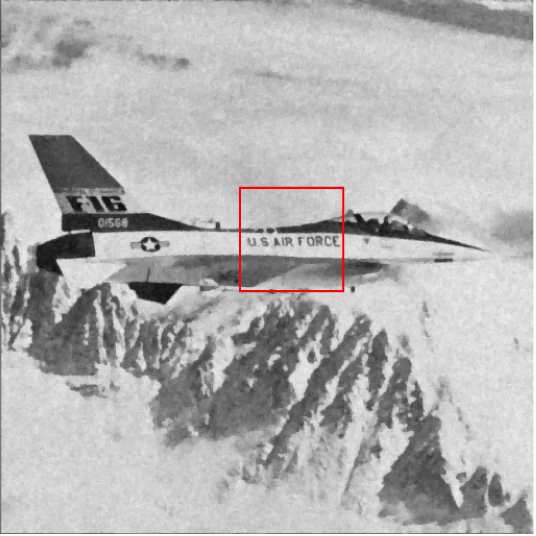}&
	\includegraphics[width=0.15\textwidth] 
        {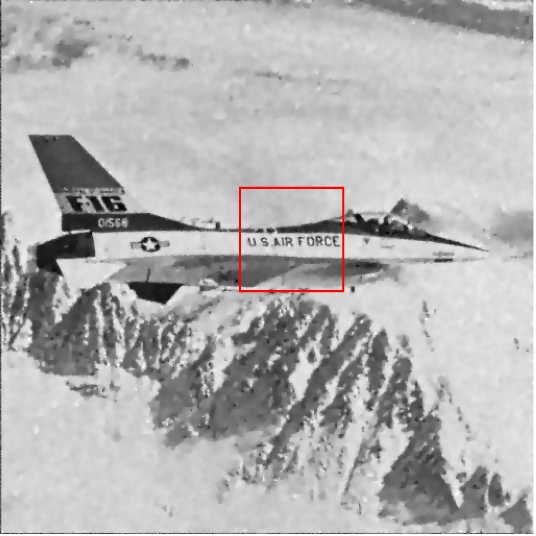}&
        \includegraphics[width=0.15\textwidth]{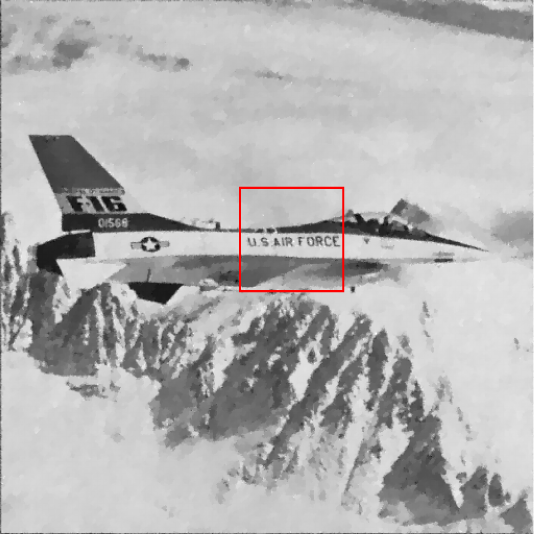}&
        \includegraphics[width=0.15\textwidth]{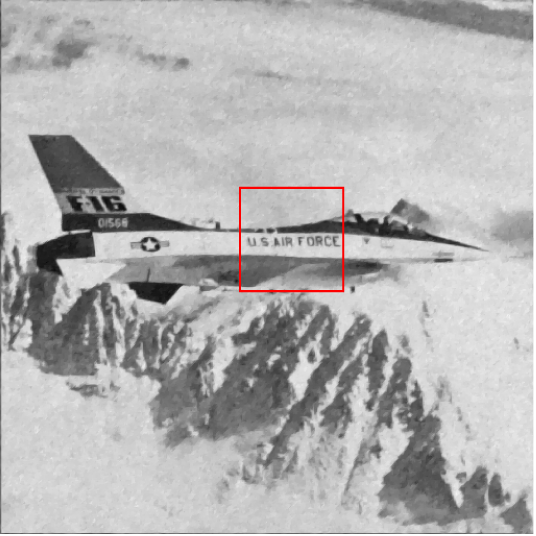}&
        \includegraphics[width=0.15\textwidth]
        {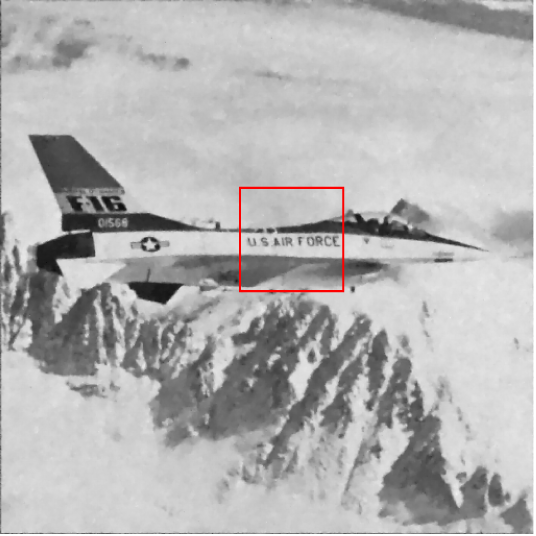}\\
        \includegraphics[width=0.15\textwidth]{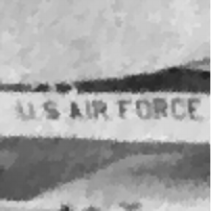}&
        \includegraphics[width=0.15\textwidth]{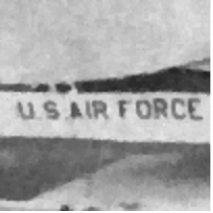}&
	\includegraphics[width=0.15\textwidth] 
        {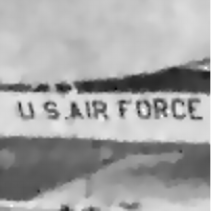}&
        \includegraphics[width=0.15\textwidth]{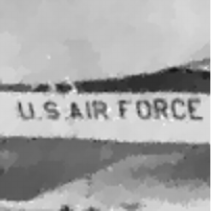}&
        \includegraphics[width=0.15\textwidth]{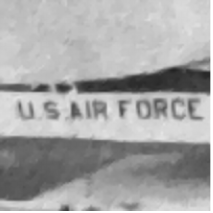}&
        \includegraphics[width=0.15\textwidth]  {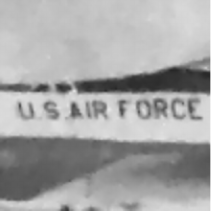}
        \end{tabular}
   \caption{The denoising results of ``Plane'' ($\sigma=20/255$), where all methods employ the same parameters as those specified in Fig. \ref{fig:Peppers}.}
   \label{fig:Plane}
\end{figure}

\subsection{Natural images denoising}
We then evaluate the proposed model on natural image denoising, comparing it against several state-of-the-art methods, including TV \cite{rudin1992nonlinear, chambolle2011first}, EE \cite{tai2011fast}, MC \cite{zhu2012image}, TAC (Total Absolute Curvature) \cite{zhong2020minimizing}, and the GCTV \cite{liu2022operator} model. Four test images are used: ``Zelda'' and ``Parrot'' are corrupted with zero-mean Gaussian noise $\sigma = 10/255$, while ``Peppers'' and ``Plane'' are corrupted with $\sigma = 20/255$. The parameters for the comparison models are selected according to the settings recommended in their original papers. Our model parameters are set as follows: $\alpha=0.1$, $\beta=0.4$ for all images; $\gamma=12$, $\tau=0.02$ for ``Zelda'' and ``Parrot''; $\gamma=10$, $\tau=0.01$ for ``Peppers'' and ``Plane''.

Quantitative comparisons in terms of PSNR and SSIM are reported in Tables \ref{tab:deniose_10/255} and \ref{tab:deniose_20/255}. The corresponding visual results are presented in Figures \ref{fig:Parrot} and \ref{fig:Peppers}, including both full-image reconstructions and local magnified views. Tables \ref{tab:deniose_10/255} and  \ref{tab:deniose_20/255} illustrate that our proposed model achieves slightly higher PSNR and SSIM than other models. While all methods effectively remove noise and recover major structures and features, the comparative models exhibit stair-casing artifacts and speckle noise in homogeneous regions, e.g., the alar rim in \cref{fig:Zelda}, or the sky and aircraft in \cref{fig:Plane}. In contrast, our reconstructions achieve better smoothness in these areas while preserving sharper edges, as shown in the magnified views.

\begin{table}[t]
    \centering
    \begin{minipage}{\textwidth}
    \centering
    \begin{tabular}{l|c|c|c|c|c|c|c}
    \toprule
    Image & Index & TV & EE & MC & TAC & GCTV & TNC\\ \hline Zelda&PSNR&$33.16$&$33.57$&$33.85$&	$34.19$	&$34.45$& $\mathbf{34.76}$\\&SSIM&$0.8692$&$0.8706$&$0.8729$&$0.8726$&$0.8870$&$\mathbf{0.8936}$\\ \hline
   Parrot&PSNR&$32.80$&$33.35$&$33.98$&$34.20$	&$34.32$& $\mathbf{34.64}$\\&SSIM&$0.8968$&$0.9047$&$0.9056$&$0.9096$&$0.9098$&$\mathbf{0.9176}$\\ 
    \bottomrule
    \end{tabular}
    \caption{PSNR/SSIM comparison on Gaussian noise removal ($\sigma=10/255$).}\label{tab:deniose_10/255}
    \end{minipage}
      
    \begin{minipage}{\textwidth}
    \centering
    \begin{tabular}{l|c|c|c|c|c|c|c}
    \toprule
    Image & Index & TV & EE & MC & TAC & GCTV & TNC\\ \hline Peppers&PSNR&$28.98$&$29.37$&$29.67$&$29.83$&$30.01$&$\mathbf{30.38}$\\ &SSIM&$0.8413$&$0.8525$&$0.8671$&$0.8644$&$0.8698$&$\mathbf{0.8829}$\\ \hline
    Plane&PSNR&$29.46$&$30.10$&$30.32$&$30.45$&$30.53$&$\mathbf{30.92}$\\ 
    &SSIM&$0.8485$&$0.8564$&$0.8718$&$0.8732$&$0.8673$&$\mathbf{0.8793}$\\ 
    \bottomrule
    \end{tabular}
    \caption{PSNR/SSIM comparison on Gaussian noise removal ($\sigma=20/255$).}\label{tab:deniose_20/255}
    \end{minipage}
\end{table}

\begin{figure}[t]
	\centering
	\begin{tabular}{c@{\hspace{2pt}}c@{\hspace{2pt}}c}
	(a) $p^{n+1/4}$  & (b) $H^{n+1/4}$ & (c) Energy\\  
        \includegraphics[width=0.3\textwidth]{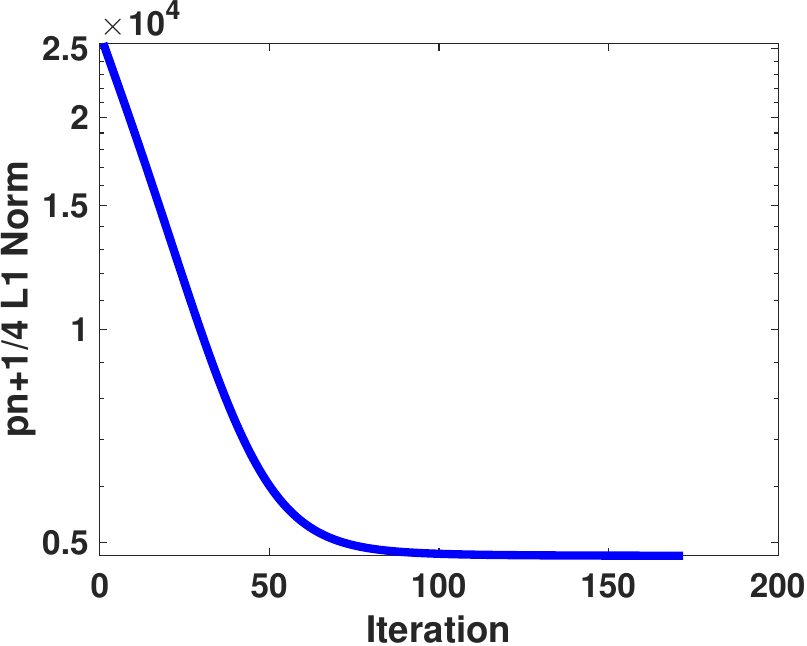}&
        \includegraphics[width=0.3\textwidth]{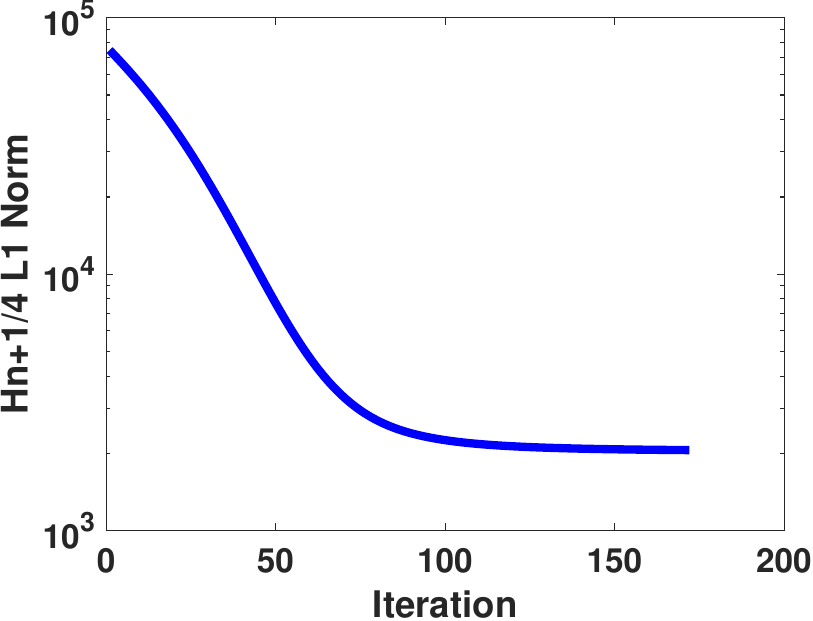}&
	\includegraphics[width=0.3\textwidth]{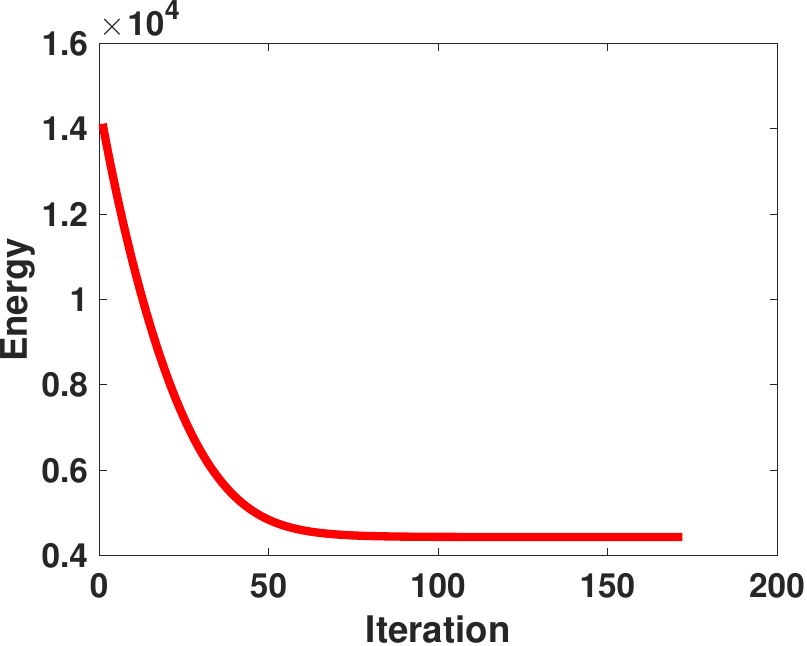}\\
    	(d) $p^{n+1}$  & (e) $H^{n+1}$ & (f) Relative error\\  
        \includegraphics[width=0.3\textwidth]{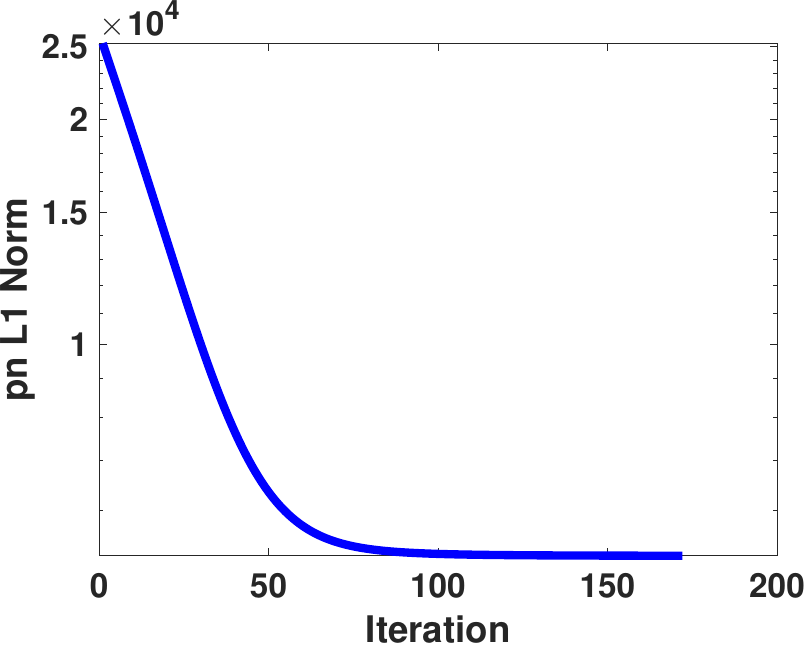}&
        \includegraphics[width=0.3\textwidth]{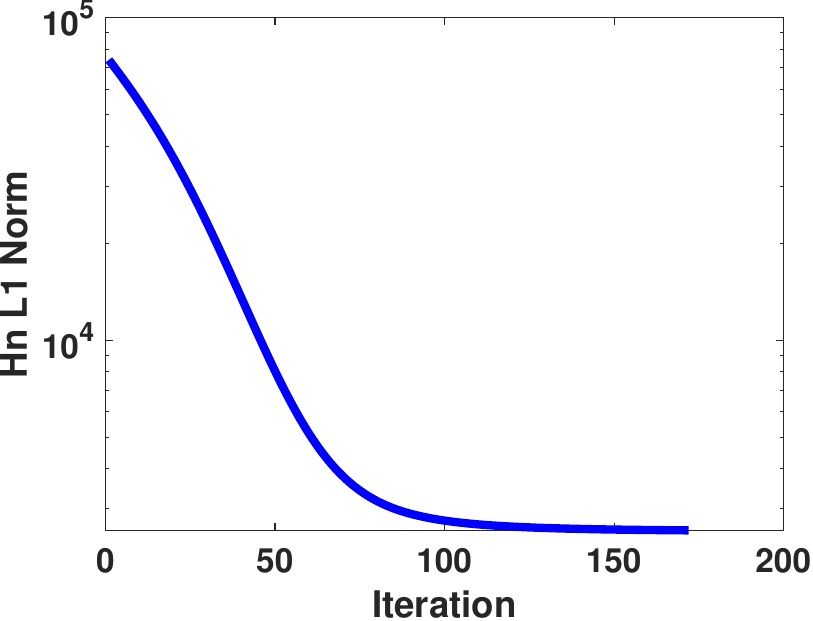}&
        \includegraphics[width=0.3\textwidth]{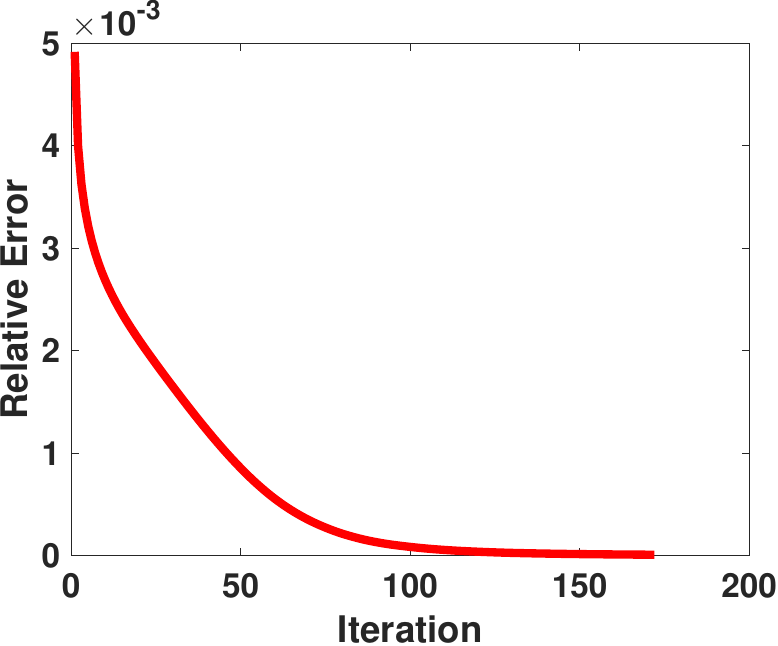}\\
        \end{tabular}
   \caption{Subproblem energy evolution for ``Zelda'' image (in Figure \ref{fig:Zelda}(f)): 
$\mathbf{p}^{n+1/4}$-problem (\ref{zuiyouproblem1}), 
$\mathbf{H}^{n+1/4}$-problem (\ref{sub_H1/4_01}), 
$\mathbf{p}^{n+1}$/$\mathbf{H}^{n+1}$-problem (\ref{3ziwenti11}), 
total energy and relative error.
   }
   \label{changes:Zelda}
\end{figure}

In \cref{changes:Zelda}, we monitor the energy evolution of the subproblems and the original problem \eqref{org_problem}. The results show that the energies associated with the \(\mathbf{p}^{n+1/4}\)-subproblem \eqref{zuiyouproblem1}, the \(\mathbf{H}^{n+1/4}\)-subproblem \eqref{sub_H1/4_01}, the joint \((\mathbf{p}^{n+1}, \mathbf{H}^{n+1})\)-subproblem \eqref{3ziwenti11}, as well as the total energy and relative error, all decrease as \(n\) increases. Furthermore, subproblems \eqref{zuiyouproblem1} and \eqref{sub_H1/4_01}, solved by Algorithm \ref{alg:Fix} and Algorithm \ref{alg:ADMM} respectively, maintain robust convergence characteristics despite employing approximate solutions. The empirical convergence supports the practical stability of the proposed optimization framework.

\begin{figure}[t]
	\centering
	\begin{tabular}{c@{\hspace{2pt}}c@{\hspace{2pt}}c@{\hspace{2pt}}c}\\
        (a) Peppers & (b) Plane & (c) Zelda  & (d) Parrot\\
         \includegraphics[width=0.235\textwidth]{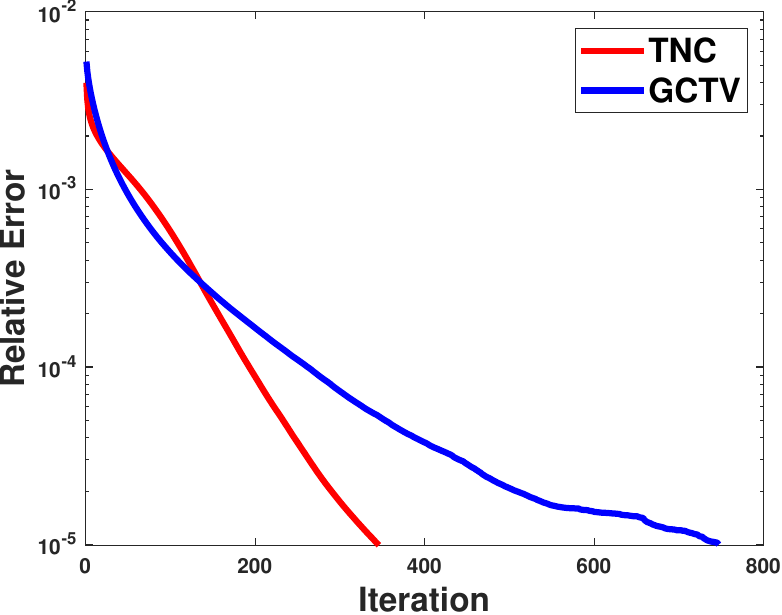}&
        \includegraphics[width=0.235\textwidth]{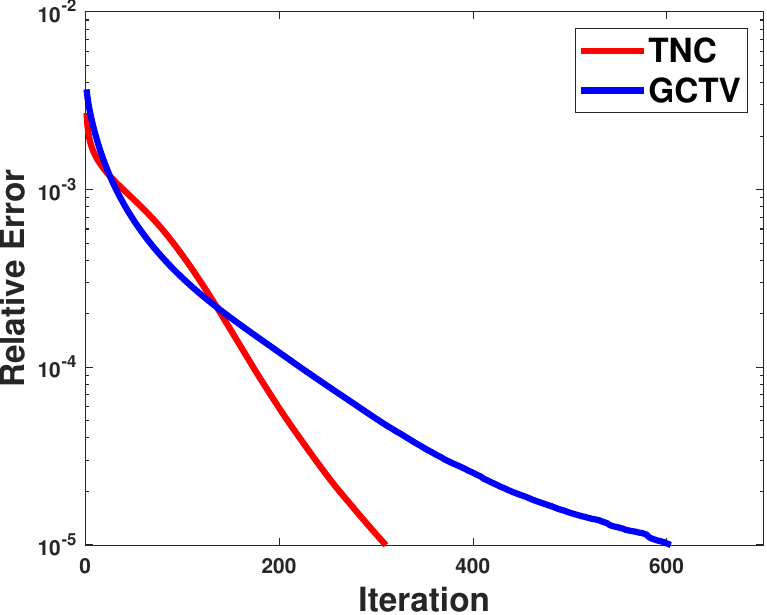}& 
        \includegraphics[width=0.235\textwidth]{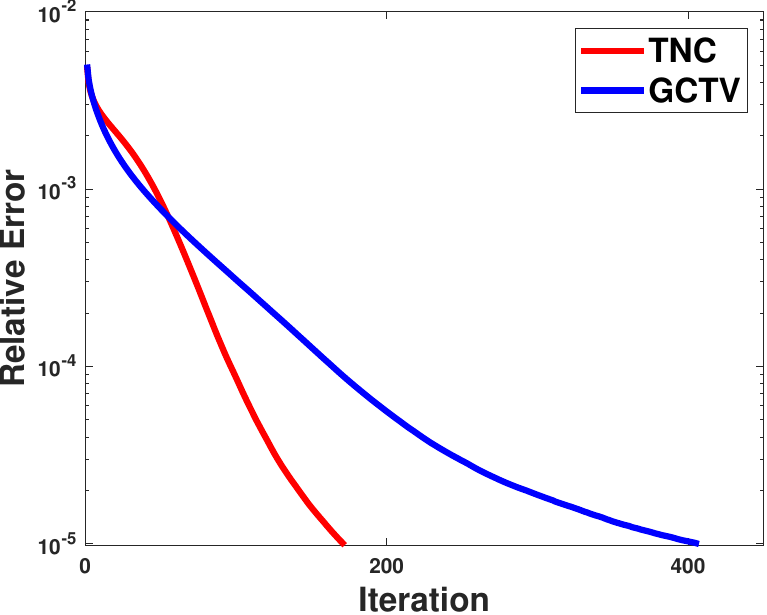}&
        \includegraphics[width=0.235\textwidth]{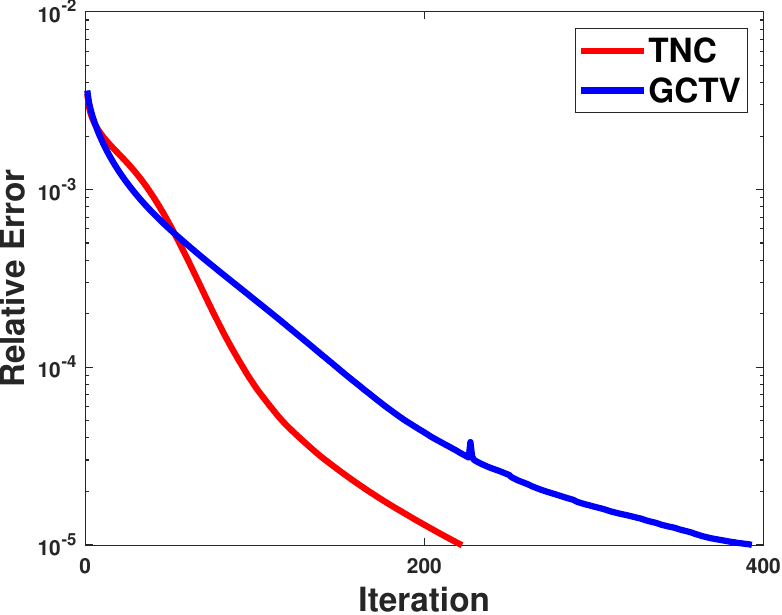}\\
        \end{tabular}
   \caption{Comparison of the relative error versus iteration count for the proposed method and the GCTV model, showing that the proposed method converges significantly faster. The subfigure correspond to the four test images, whose reconstruction results are shown in columns (f) and (g) of Figures~\ref{fig:Parrot}–\ref{fig:Peppers}. The stopping criterion is based on a fixed $tol=1e-5$.}
   \label{TNC_VS_GCTV}
\end{figure}

We further compare the convergence behavior of the proposed method with the GCTV model \cite{liu2022operator}, which integrates Gaussian curvature and total variation regularization within a unified reconstruction framework through an operator-splitting strategy. The evolution of relative errors across iterations is documented in \cref{TNC_VS_GCTV}, with corresponding denoising results presented in columns (f) and (g) of Figures \ref{fig:Parrot}-\ref{fig:Peppers}. Our TNC model achieves the target tolerance in fewer iterations, demonstrating substantially faster convergence compared to GCTV. Quantitative performance comparisons in \Cref{tab:performance} include total runtime, iteration counts, and computational time per iteration. Despite slightly higher computational cost per iteration due to the complexity of multi-directional normal curvature computation, our method achieves the target accuracy with reduced iteration numbers and decreased total computation time.

\begin{table}[t]
\begin{center}
    \begin{tabular}{l|c|c|c|c}
        \toprule
        Image & Methods & Iterations  & Time     & Time/per iteration \\ \hline
        \multirow{2}{*}{Peppers}&{TNC}&{347}&{19.15}&{0.055}\\ \cline{2-5}
        & GCTV &  747 & 31.37  &  0.042 \\  \hline
        \multirow{2}{*}{Plane}&{TNC}&{310}&{120.32}&{0.388}\\\cline{2-5}
        & GCTV &  604 & 172.36  &  0.285  \\ \hline
        \multirow{2}{*}{Zelda}&{TNC}&{172}&{66.16}&{0.385}\\ \cline{2-5}
        & GCTV & 407 & 105.96  &  0.260 \\ \hline
        \multirow{2}{*}{Parrot}&{TNC}&{222}&{58.30}&{0.263}\\ \cline{2-5}
        & GCTV &  393 & 71.33  &  0.182   \\ \bottomrule
    \end{tabular}
\end{center}
    \caption{Efficiency comparison with the GCTV model in terms of numbers of iterations, total time and averaged computational time per iteration, where time is recorded in seconds.}
    \label{tab:performance}
    
\end{table}

\subsection{Effects of parameters}\label{sec_effect_alpha_beta}
Our TNC model (\ref{model1}) incorporates three parameters: $\alpha$, \(\beta\), and $\gamma$, which govern the curvature regularization term, the TV regularization term, and the data fidelity term, respectively. To investigate the influence of each parameter, we employ additive Gaussian noise with a standard deviation of $\sigma = 20/255$ and adhere to a strategy of varying only one parameter at a time while keeping the others fixed during parameter analysis.

\begin{figure}[t]
	\centering
	\begin{tabular}{c@{\hspace{2pt}}c@{\hspace{2pt}}c@{\hspace{2pt}}c}
	(a) & (b) & (c) & (d) \\  
        \includegraphics[width=0.22\textwidth]{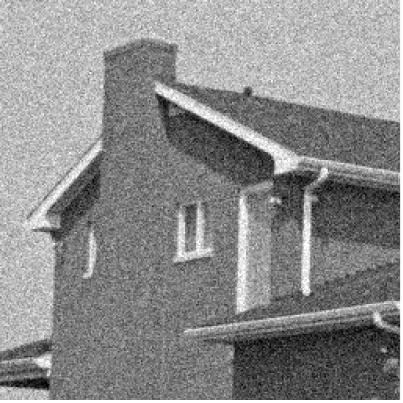}&
        \includegraphics[width=0.22\textwidth]{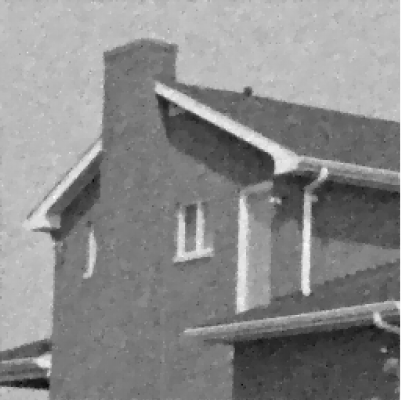}&
        \includegraphics[width=0.22\textwidth]{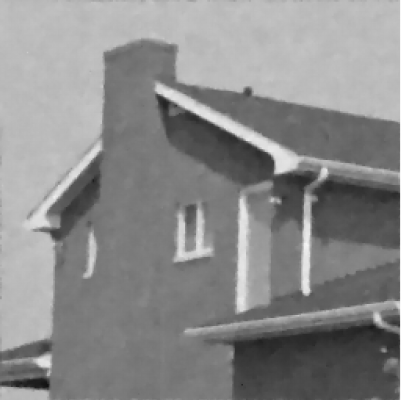}&
        \includegraphics[width=0.22\textwidth]
        {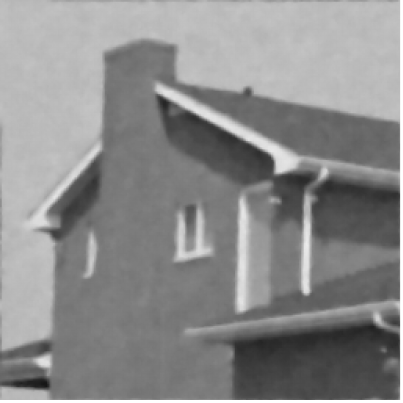}
        \end{tabular}
   \caption{The denoising results with different $\alpha$ by fixing $\beta=0.5$ and $\gamma=10$: (a) Noisy (Gaussian noise and $\sigma = 20/255$), (b) $\alpha=0$, (c) $\alpha=0.1$, (d) $\alpha=0.4$.}
   \label{Effect of alpha}
\end{figure}

\begin{figure}[!htp]
	\centering
	\begin{tabular}{c@{\hspace{2pt}}c@{\hspace{2pt}}c@{\hspace{2pt}}c}
	(a) & (b) & (c) & (d) \\  
        \includegraphics[width=0.22\textwidth]{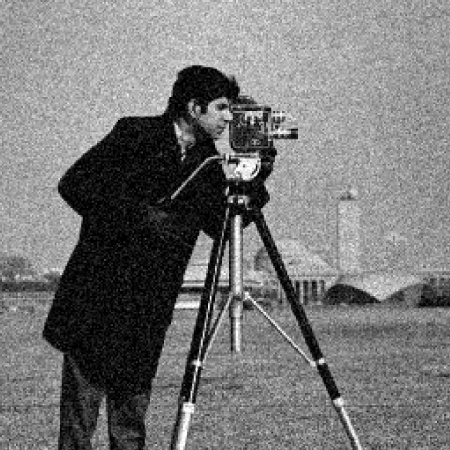}&
        \includegraphics[width=0.22\textwidth]{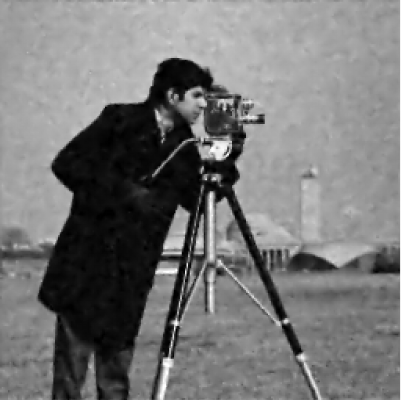}&
        \includegraphics[width=0.22\textwidth]{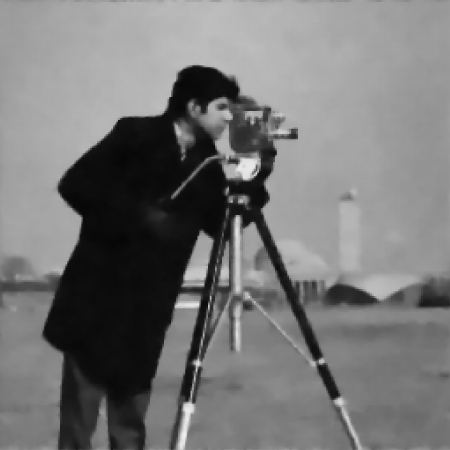}&
        \includegraphics[width=0.22\textwidth]
        {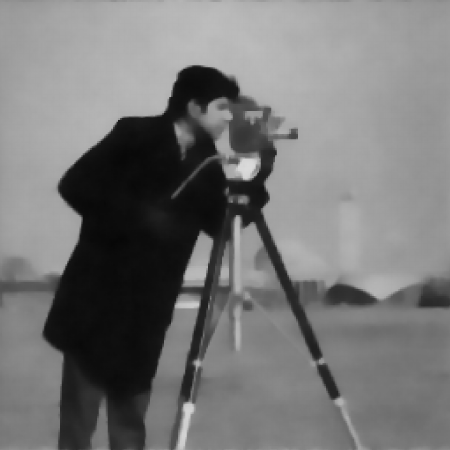}
        \end{tabular}
   \caption{The denoising results with different $\beta$ by fixing $\alpha=0.1$ and $\gamma=5$: (a) Noisy (Gaussian noise and $\sigma = 20/255$), 
(b) $\beta=0$, (c) $\beta=0.3$, (d) $\beta=0.8$.}
   \label{Effect of beta}
\end{figure}

We first evaluate the impact of parameter $\alpha$ selected from the set $\{0, 0.1, 0.4\}$ on the ``House'' image with fixed $\beta$ and $\gamma$. As shown in \cref{Effect of alpha}, compared to using TV regularization alone (i.e., \(\alpha=0\)), introducing curvature regularization and increasing the weight \(\alpha\) enhances image smoothing. Further analysis of \(\beta \in \{0, 0.3, 0.8\}\) via \cref{Effect of beta} reveals that when \(\beta\) is too small, insufficient TV regularization results in residual noise; whereas an excessively large \(\beta\) causes the TV term to dominate the regularization process, leading to over-smoothed images and loss of details. Notably, when \(\beta=0\), i.e., only curvature regularization is applied, denoising performance degrades, validating the necessity of combining curvature and TV terms (further analysis is provided in Appendix \ref{appendix:B}). For the data fidelity parameter \(\gamma\), with fixed \(\alpha\) and \(\beta\), its influence follows a clear trend: an overly small \(\gamma\) causes excessive smoothing of structural features, while an overly large \(\gamma\) weakens the regularization effect, resulting in residual noise. Thus, the three parameters need to be chosen according to image characteristics and noise levels to achieve an optimal trade-off between denoising and feature preservation.

\begin{figure}[htp]
	\centering
	\begin{tabular}{c@{\hspace{2pt}}c@{\hspace{2pt}}c@{\hspace{2pt}}c}
	(a) & (b) & (c) & (d) \\ 
        \includegraphics[width=0.235\textwidth]{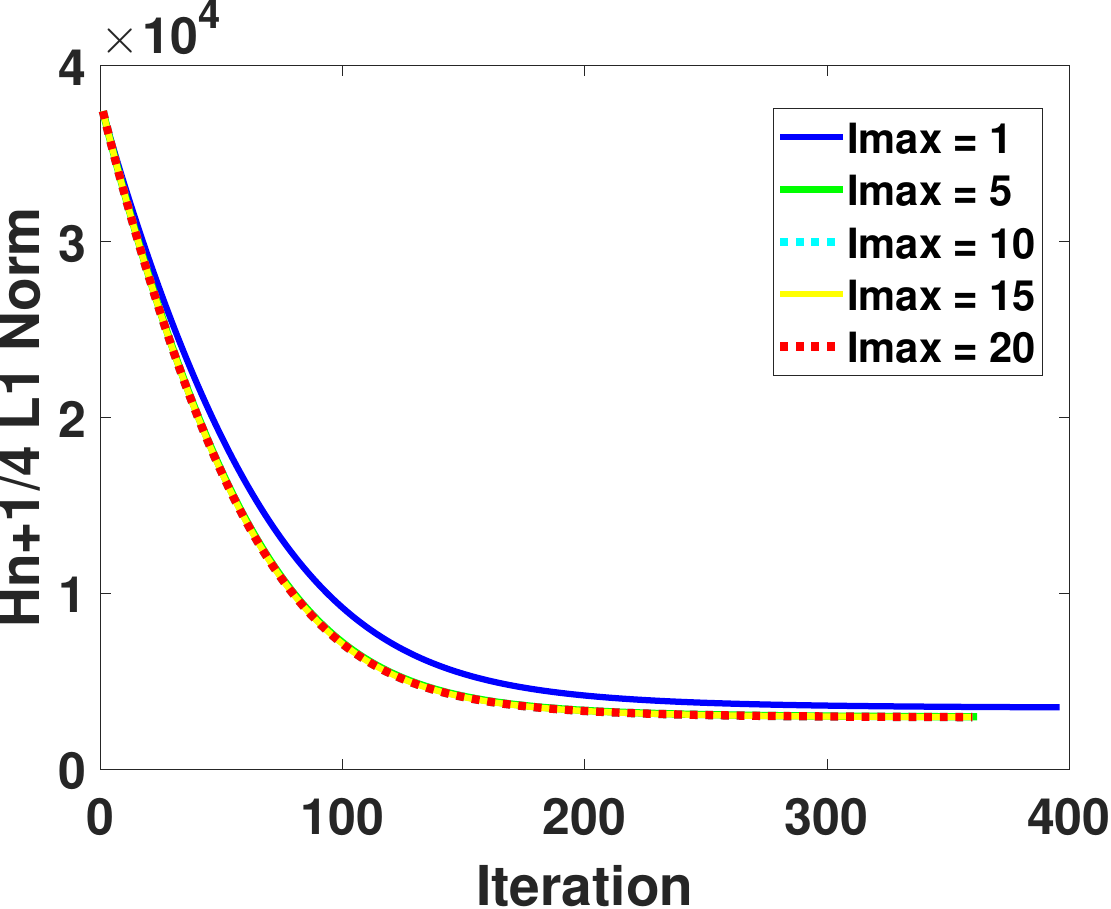}&
        \includegraphics[width=0.235\textwidth]{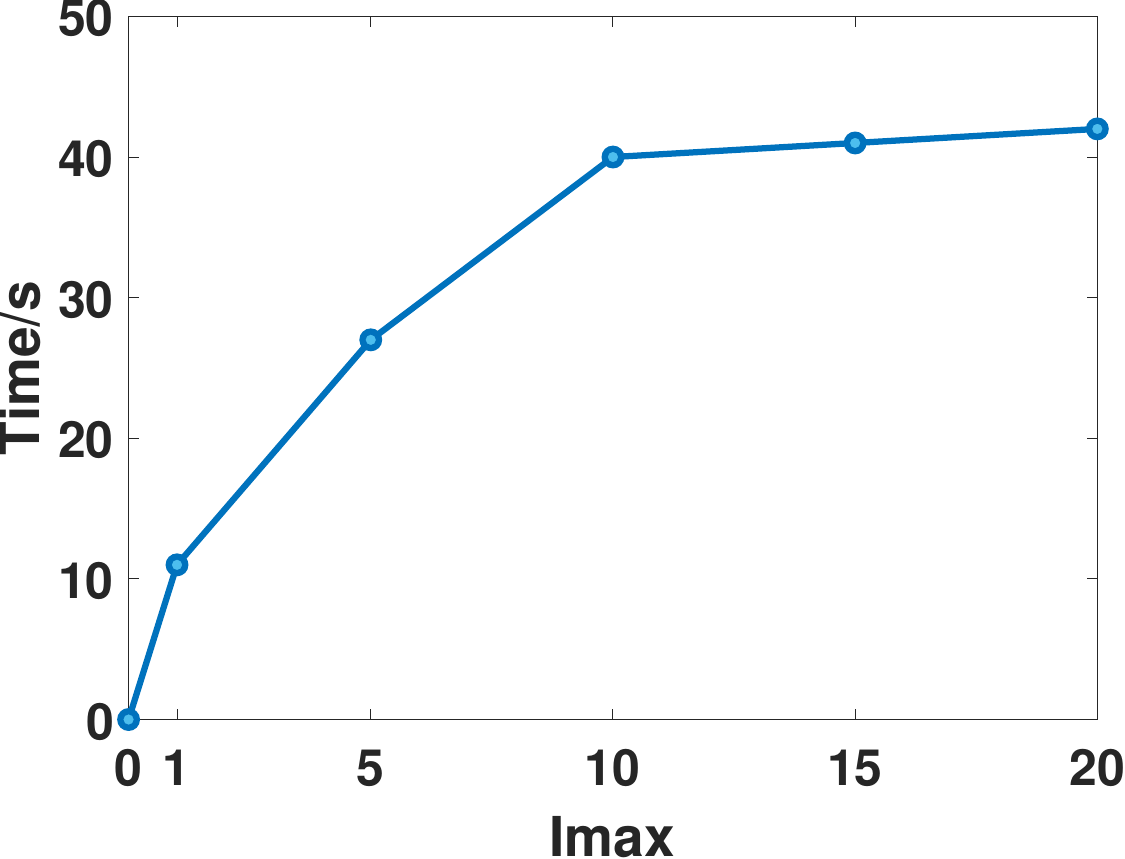}&
        \includegraphics[width=0.235\textwidth]{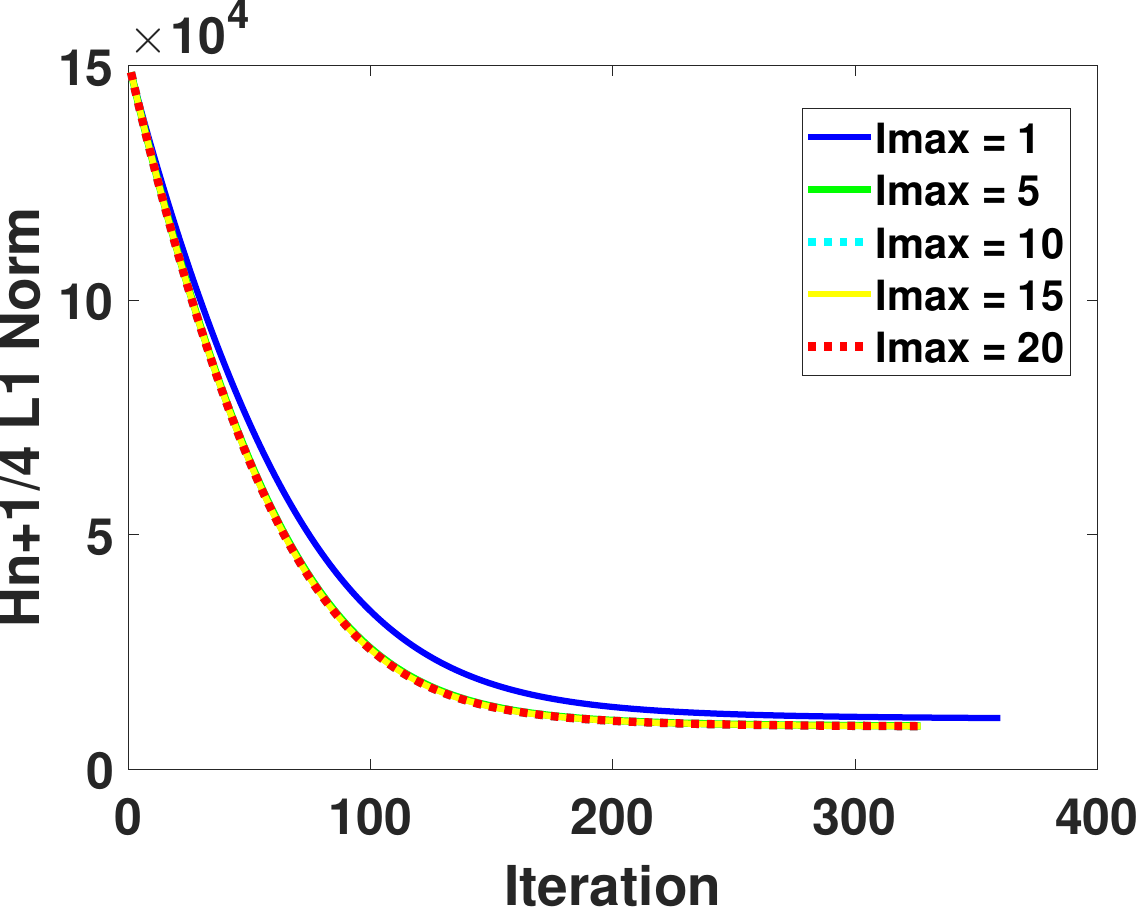}&
        \includegraphics[width=0.235\textwidth]{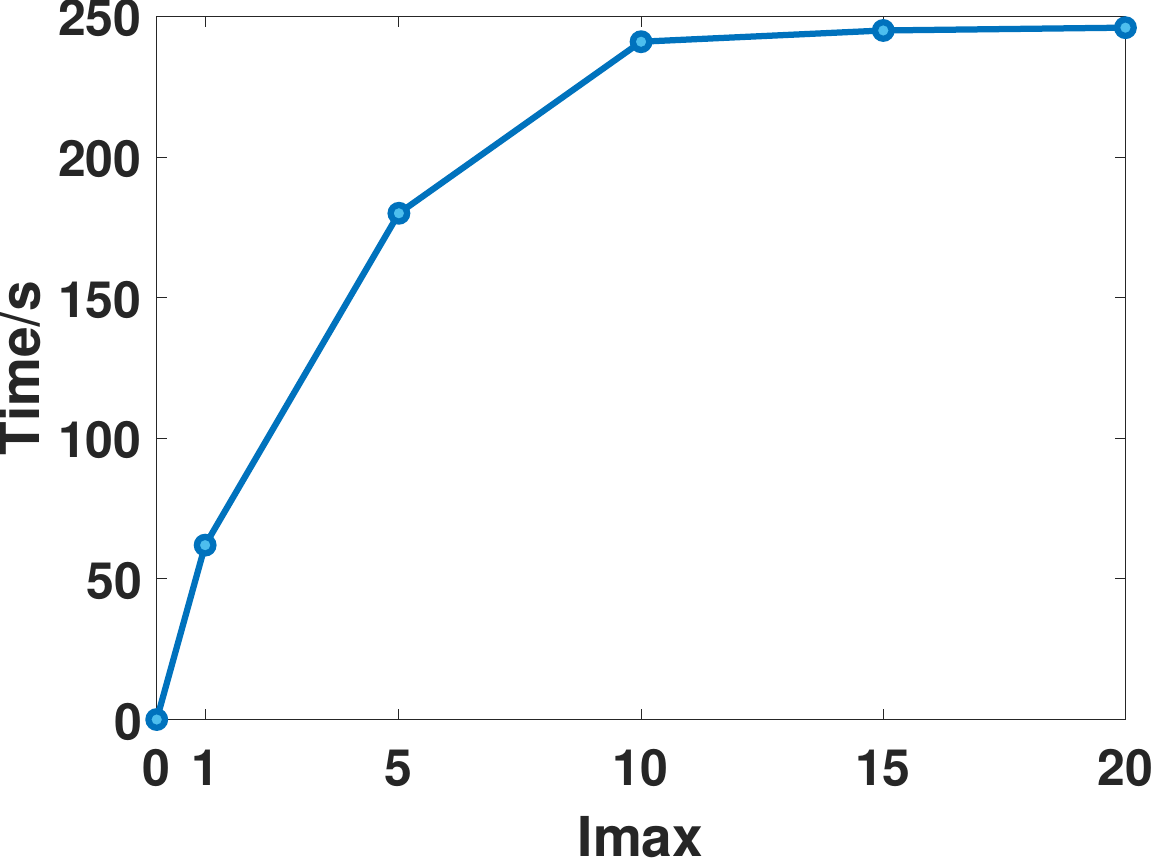}
        \end{tabular}
   \caption{Convergence and computational efficiency of \cref{alg:ADMM} under different values of $I_{\max}$. (a), (c) the $\ell_1$ norm of $\mathbf{H}^{n+1/4}$ for the Peppers and Plane images; (b), (d) corresponding cumulative computation time.}
   \label{Effect of Imax}
\end{figure}

Finally, we investigate the effect of the maximum number of inner iterations $I_{\max}$ used in \cref{alg:ADMM} for solving the $\mathbf{H}^{n+1/4}$ subproblem.  
The test images are ``Peppers'' and ``Plane'', with all other parameters held fixed.  
We track the $\ell_{1}$ norm of the computed $\mathbf{H}^{n+1/4}$ to evaluate the convergence behavior of the ADMM.
\Cref{Effect of Imax} illustrates the $\ell_1$ norm as a function of iteration count for $I_{\max} \in \{1, 5, 10, 15, 20\}$. The results confirm that all tested values of $I_{\max}$ lead to a monotonic reduction in the $\ell_1$ norm and eventual convergence. Notably, the convergence trajectories for $I_{\max} = 1$ in subfigures (a) and (c) closely match those obtained with larger $I_{\max}$ values. Furthermore, subfigures (b) and (d) report the total computational time required for this subproblem across different $I_{\max}$ settings, demonstrating that $I_{\max} = 1$ yields the highest computational efficiency. Since the outer iterations in \Cref{alg:osm} progressively update the Lagrange multipliers, performing just one ADMM iteration per outer cycle suffices to ensure convergence while minimizing computational overhead.

\section{Concluding remarks}
\label{sec:conclusions}

We proposed a new total curvature regularization based on multi-directional normal curvatures, which effectively preserves sharp features. The associated high-order problem is efficiently solved via operator splitting method, where each subproblem is either analytically tractable or numerically efficient. The method requires minimal parameter tuning and is robust across different settings. The efficiency and performance of the proposed method are demonstrated through systematic numerical experiments on surface smoothing and image denoising. 

Our future work will focus on establishing a rigorous variational analysis theory within the bounded variation function space to theoretically validate the edge-preserving properties of curvature regularization. Second, we will develop more efficient numerical solvers and extend the total normal curvature regularization model to 3D geometric data processing and image restoration tasks. Furthermore, we also plan to investigate effective methodologies for integrating curvature regularization with deep learning architectures, such as \cite{li2025curvpnp,wang2026taylor}, to enhance the geometric awareness and generalization capabilities of deep neural networks.

\appendix
\section{{Definition of normal curvature (\ref{eq:normal_curvature})}}
\label{appendix:A}
Let $\mathbf{r} = \mathbf{r}(x,y)$ be a regular parametric surface $\mathcal{S}$ and $(x,y)$ be the coordinates on $\Omega$. The partial derivatives of $\mathbf{r}$ with respect to $x$ and $y$ yield the tangent vectors
\begin{equation*}
    \mathbf{r}_{x} = (1,0,v_{x}) \quad \text{and} \quad \mathbf{r}_{y}=(0,1,v_{y}),
\end{equation*}
where $v_x = \frac{\partial v}{\partial x}$ and $v_y = \frac{\partial v}{\partial y}$ are the first-order partial derivatives of $v$. For any angle $\theta \in [0, 2\pi)$, we can define a direction vector in the tangent plane as 
\[\mathbf{t} = \cos\theta \cdot \mathbf{r}_x + \sin\theta \cdot \mathbf{r}_y = \left( \cos\theta, \sin\theta, v_x \cos\theta + v_y \sin\theta \right).\]
The unit normal vector to the surface at point $(x,y,v(x,y))$ is given by $$\mathbf{n} = \frac{\mathbf{N}}{|\mathbf{N}|} = \frac{(-v_x, -v_y, 1)}{\sqrt{v_x^2 + v_y^2 + 1}},$$ where $\mathbf{N} = (-v_x, -v_y, 1)$ is the non-normalized normal vector.

We employ the first and second fundamental forms to describe the local geometry of the surface. The first fundamental form characterizes how distances on the surface relate to distances in the parameter domain and is defined by coefficients 
\[E = 1 + v_x^2, \quad F = v_x v_y, \quad G = 1 + v_y^2.\]
For a direction vector $\mathbf{t}$, the first fundamental form can be expressed as 
\[\mathbf{I}(\mathbf{t}, \mathbf{t}) = E \cos^2\theta + 2F \cos\theta \sin\theta + G \sin^2\theta = 1 + \left( v_x \cos\theta + v_y \sin\theta \right)^2.\]
The second fundamental form describes how the surface bends in different directions and is characterized by coefficients $$L = \frac{v_{xx}}{\sqrt{1 + v_x^2 + v_y^2}}, \quad M = \frac{v_{xy}}{\sqrt{1 + v_x^2 + v_y^2}}, \quad N = \frac{v_{yy}}{\sqrt{1 + v_x^2 + v_y^2}},$$ where $v_{xx} = \frac{\partial^2 v}{\partial x^2}$, $v_{xy} = \frac{\partial^2 v}{\partial x \partial y}$, and $v_{yy} = \frac{\partial^2 v}{\partial y^2}$ are the second-order partial derivatives of $v$.

For a direction vector $\mathbf{t}$, the second fundamental form can be written as 
\begin{equation}
\begin{aligned}
      \mathbf{II}(\mathbf{t}, \mathbf{t}) &= L \cos^2\theta + 2M \cos\theta \sin\theta + N \sin^2\theta \\
      &= \frac{v_{xx} \cos^2\theta + 2v_{xy} \cos\theta \sin\theta + v_{yy} \sin^2\theta}{\sqrt{1 + v_x^2 + v_y^2}}.  
\end{aligned}
\end{equation}
The normal curvature in direction $\theta$ is defined as the ratio of the second fundamental form to the first fundamental form: 
\begin{equation}
    \kappa_n(\theta) = \frac{\mathbf{II}(\mathbf{t}, \mathbf{t})}{\mathbf{I}(\mathbf{t}, \mathbf{t})} = \frac{v_{xx} \cos^2\theta + 2v_{xy} \cos\theta \sin\theta + v_{yy} \sin^2\theta}{\sqrt{1 + v_x^2 + v_y^2} \cdot \left[1 + (v_x \cos\theta + v_y \sin\theta)^2 \right]},
\end{equation}
which can be elegantly expressed using the Hessian matrix $\mathbf{H}$ of $v$ as 
\begin{equation*}
  \kappa_n(\theta) = \frac{\mathbf{t}^T \mathbf{H} \mathbf{t}}{\sqrt{1 + \vert \nabla v \vert^{2}} \cdot \left( 1 + (\nabla v : \mathbf{t})^2 \right)}, 
\end{equation*}
where $\mathbf{t} = (\cos\theta, \sin\theta)^{\top}$ is now treated as a unit vector in $\mathbb{R}^2$, $\nabla v = (v_x, v_y)$ is the gradient of $v$, 
$\nabla v :\mathbf{t}=v_{1}t_{1}+v_{2}t_{2}$, and
\begin{equation*}
   \mathbf{t}^T \mathbf{H} \mathbf{t}= v_{xx} \cos^2 \theta + 2v_{xy} \cos \theta \sin \theta + v_{yy} \sin^2 \theta ,
\end{equation*}
represents the second directional derivative of $v$ in the direction $\mathbf{t}$. Therefore, our TNC regularization term is:
\begin{equation}
\label{a:defTNC}
    \int_{\Omega}\int_{0}^{2\pi}\vert\kappa_{n}(\theta)\vert \mathrm{d}\theta \mathrm{d}\mathrm{s} =\int_{\Omega} \int_{0}^{2\pi} \frac{\vert \mathbf{t}^{\top} \mathbf{H} \mathbf{t} \vert }{\sqrt{1 + \vert \nabla v \vert^{2}} \cdot \left( 1 + (\nabla v : \mathbf{t})^2 \right)} \mathrm{d}\theta \mathrm{d}\mathrm{s}.
\end{equation}

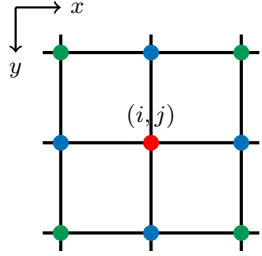
\begin{figure}[]
    \centering
    \begin{tikzpicture}[scale=1.2]
    \draw[very thick] (-0.2,0) -- (2.2,0);
    \draw[very thick] (-0.2,1) -- (2.2,1);
    \draw[very thick] (-0.2,2) -- (2.2,2);
    
    \draw[very thick] (0,-0.2) -- (0,2.2);
    \draw[very thick] (1,-0.2) -- (1,2.2);
    \draw[very thick] (2,-0.2) -- (2,2.2);
    
    \foreach \x/\y in {0/0, 2/0, 0/2, 2/2} {
        \fill[ForestGreen] (\x,\y) circle (2.5pt);
    }
    \foreach \x/\y in {1/0, 0/1, 2/1, 1/2} {
        \fill[RoyalBlue] (\x,\y) circle (2.5pt);
    }
    \fill[red] (1,1) circle (2.5pt);
    
    
    \node at (1,1.3) {\small $(i,j)$};
    
    
    \coordinate (arrowstart) at (-0.5, 2.5);
    \draw[->, thick] (arrowstart) -- ++(0.5, 0) node[right] {\small $x$};
    \draw[->, thick] (arrowstart) -- ++(0, -0.5) node[below] {\small $y$};
    \end{tikzpicture}
    \caption{The points used to compute the Hessian matrix at the central point $(i,j)$, where the red point is the center itself, the blue axial points compute the second-order partial derivatives, and the green diagonal points compute the mixed partial derivatives.}
    \label{fig:Hessian_fig}
\end{figure}

\section{Anisotropy of discrete second-order derivatives}
\label{appendix:B}
In image processing, second-order derivatives are crucial for characterizing local curvature and textural information within an image.  
For a scalar-valued image function $v(x,y)$, the continuous Hessian matrix $\mathbf{H}$ comprises $v_{xx}$, $v_{yy}$, and $v_{xy}$ (with $v_{xy} = v_{yx}$). In discrete settings, these quantities are typically approximated by finite differences.
\cref{fig:Hessian_fig} illustrates the commonly used stencil to compute these approximations at the center pixel $(i,j)$ with the explicit formulas provided in \eqref{eq:hessian_formulas}, where $h$ denotes the grid spacing, typically set as $1$ for pixel-based computations.
\begin{equation}
    \begin{aligned}
        v_{x x}(i, j) &\approx \frac{v(i+1, j)-2 v(i, j) + v(i-1, j)}{h^{2}}; \\
    v_{y y}(i, j) &\approx \frac{v(i, j+1)-2 v(i, j) + v(i, j-1)}{h^{2}};\\
    v_{xy}=v_{yx} &\approx \frac{v(i+1, j+1)- v(i+1, j-1)-v(i-1, j+1) + v(i-1, j-1)}{4h^{2}}.
    \end{aligned}
    \label{eq:hessian_formulas}
\end{equation}
In the continuous domain, a straight edge, such as a vertical or diagonal line, has zero Gaussian curvature, and the determinant of the Hessian, given by $\det(\mathbf{H}) = v_{xx}v_{yy} - v_{xy}^2$, is identically zero. Additionally, the second directional derivative along the tangent of the edge vanishes. Ideally, these geometric properties should be preserved during discretization. However, it is not always the case, particularly for edges that are not aligned with the coordinate axes.

\begin{figure}[t]
    \centering
    \begin{subfigure}[b]{0.2\textwidth}
    \centering
    \begin{tikzpicture}[scale=0.3]
    \draw[step=1cm,gray,line width=0.2mm] (0,0) grid (7,7);
    \fill[black] (0,7) rectangle (3,0);
    \draw[step=1cm,gray,line width=0.2mm] (0,0) grid (7,7);
    \end{tikzpicture}
    \caption*{(I)}
    \end{subfigure}
    \hfill
    \begin{subfigure}[b]{0.2\textwidth}
    \centering
    \begin{tikzpicture}[scale=0.3]
    \draw[step=1cm,gray,line width=0.2mm] (0,0) grid (7,7);
    \fill[black] (0,0) -- (0,7) -- (7,7) -- cycle;
    \draw[step=1cm,gray,line width=0.2mm] (0,0) grid (7,7);
    \end{tikzpicture}
    \caption*{(II)}
    \end{subfigure}
    \hfill
    \begin{subfigure}[b]{0.2\textwidth}
    \centering
    \begin{tikzpicture}[scale=0.3]
    \draw[step=1cm,gray,line width=0.2mm] (0,0) grid (7,7);
    \foreach \x in {0,...,6} {
        \foreach \y in {0,...,6} {
        \pgfmathparse{int(\y > \x)}
        \ifnum\pgfmathresult=1
        \fill[black] (\x,\y) rectangle (\x+1,\y+1);
        \fi
        }
    }
    \draw[step=1cm,gray,line width=0.2mm] (0,0) grid (7,7);
    \end{tikzpicture}
    \caption*{(III)}
    \end{subfigure}
    \hfill
    \begin{subfigure}[b]{0.2\textwidth}
    \centering
    \begin{tikzpicture}[scale=0.3]
    \draw[step=1cm,gray,line width=0.2mm] (0,0) grid (7,7);
    \foreach \x in {0,...,6} {
        \foreach \y in {0,...,6} {
        \pgfmathparse{int(\y == \x)} %
        \ifnum\pgfmathresult=1
        \fill[black!30] (\x,\y) rectangle (\x+1,\y+1);
        \else
            \pgfmathparse{int(\y > \x)}
            \ifnum\pgfmathresult=1
                \fill[black] (\x,\y) rectangle (\x+1,\y+1);
            \fi
        \fi
        }
    }
    \draw[step=1cm,gray,line width=0.2mm] (0,0) grid (7,7);
    \end{tikzpicture}
    \caption*{(IV)}
    \end{subfigure}
    \caption{Illustrative edge patterns and their discretizations. (I) A continuous vertical line pattern (Its discrete representation is identical to its continuous form). (II) A continuous diagonal line pattern. (III) A binary discrete representation of pattern (II). (IV) A multi-level discrete representation of pattern (II), where pixels along the diagonal transition are assigned a value of $1/2$ (black and white correspond to 0 and 1, respectively). }
    \label{fig:pattern_line}
\end{figure}
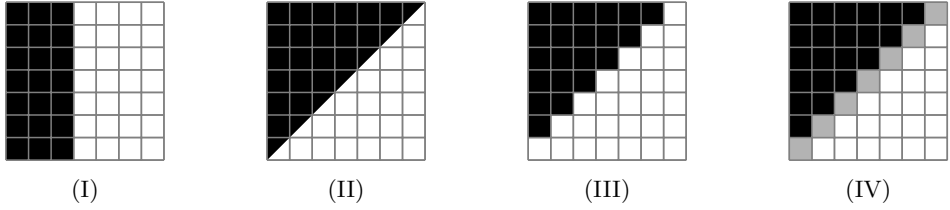

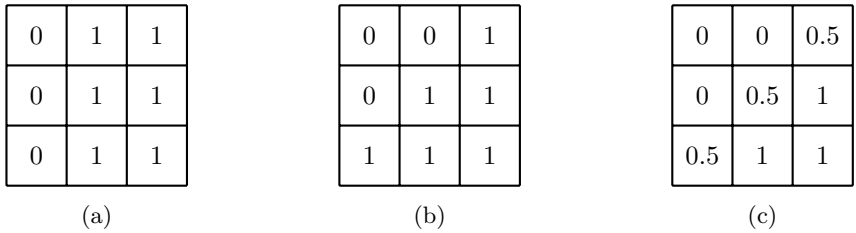
\begin{figure}[htbp]
    \centering
    \begin{subfigure}[b]{0.32\textwidth}
        \centering
        \begin{tikzpicture}[scale=0.8]
            \draw[black,thick] (0,0) grid (3,3);
            \def\pixelvalues{{
                {0, 1, 1}, 
                {0, 1, 1}, 
                {0, 1, 1}  
            }}
            \foreach \row in {0,1,2} {
                \foreach \col in {0,1,2} {
                    \pgfmathsetmacro{\xpos}{\col + 0.5}
                    \pgfmathsetmacro{\ypos}{2 - \row + 0.5} 
                    \pgfmathsetmacro{\value}{\pixelvalues[\row][\col]}
                    \node at (\xpos, \ypos) {\pgfmathprintnumber{\value}};
                }
            }
        \end{tikzpicture}
        \caption*{(a)}
    \end{subfigure}
    \hfill
    \begin{subfigure}[b]{0.32\textwidth}
        \centering
        \begin{tikzpicture}[scale=0.8]
            \draw[black,thick] (0,0) grid (3,3);
            \def\pixelvalues{{
                {0, 0, 1}, 
                {0, 1, 1}, 
                {1, 1, 1}  
            }}
            \foreach \row in {0,1,2} {
                \foreach \col in {0,1,2} {
                    \pgfmathsetmacro{\xpos}{\col + 0.5}
                    \pgfmathsetmacro{\ypos}{2 - \row + 0.5} 
                    \pgfmathsetmacro{\value}{\pixelvalues[\row][\col]}
                    \node at (\xpos, \ypos) {\pgfmathprintnumber{\value}};
                }
            }
        \end{tikzpicture}
        \caption*{(b)}
        \label{fig:pattern_binary_staircase}
    \end{subfigure}
    \hfill
    \begin{subfigure}[b]{0.32\textwidth}
        \centering
        \begin{tikzpicture}[scale=0.8]
            \draw[black,thick] (0,0) grid (3,3);
            \def\pixelvalues{{
                {0, 0, 0.5}, 
                {0, 0.5, 1}, 
                {0.5, 1, 1}  
            }}
            \foreach \row in {0,1,2} {
                \foreach \col in {0,1,2} {
                    \pgfmathsetmacro{\xpos}{\col + 0.5}
                    \pgfmathsetmacro{\ypos}{2 - \row + 0.5} 
                    \pgfmathsetmacro{\value}{\pixelvalues[\row][\col]}
                    \node at (\xpos, \ypos) {\pgfmathprintnumber{\value}};
                }
            }
        \end{tikzpicture}
        \caption*{(c)}
        \label{fig:pattern_multi_level}
    \end{subfigure}
    \caption{Typical $3 \times 3$ pixel neighborhoods representing the discrete patterns (I), (III), and (IV) from \cref{fig:pattern_line}, respectively. }
    \label{fig:all_patterns_comparison}
\end{figure}

We illustrate this discrepancy by comparing discrete representations of a diagonal edge, which is originally defined continuously in \cref{fig:pattern_line} (II). We consider two discretizations: a binary staircase approximation, as shown in \cref{fig:pattern_line} (III), and a multi-level version with intermediate values along the diagonal transition, as depicted in \cref{fig:pattern_line} (IV). The associated \(3 \times 3\) local neighborhoods for these discretizations are presented in \cref{fig:all_patterns_comparison}. The discrete Hessian components, calculated using \eqref{eq:hessian_formulas}, are summarized in \cref{tab:hessian_values}.

Interestingly, the binary discretization (Pattern III) results in a nonzero mixed derivative and a negative Hessian determinant. This occurs due to the application of axis-aligned finite difference schemes to features that are obliquely oriented. In contrast, the multi-level representation (Pattern IV) yields zero second-order derivatives and a zero determinant at the center pixel, which seems to align with continuous theory. However, this is not due to any inherent isotropy in the discrete Hessian operator; rather, it is a numerical artifact resulting from the placement of intermediate pixel values (e.g., 0.5) along the diagonal transition. These values act as artificial smoothing, compensating for discrete anisotropy and neutralizing the computed curvature. Such precise and symmetric gray-level transitions are rare in practical image data, especially in natural scenes or noisy measurements. Therefore, while Pattern IV aligns with continuous curvature properties, it represents an idealized scenario not typical in real-world applications. The observed zero curvature should not be interpreted as evidence of isotropy in the discrete operator but rather as a result of specific numerical compensation.

\begin{table}[H]
    \centering
    \setlength{\tabcolsep}{12pt}
    \begin{tabular}{l|c|c|c|c|c}
    \toprule
    Pattern& $3 \times 3$ Nbr. & $v_{xx}$ &$v_{yy}$ & $v_{xy}$ & $\mathrm{det}\mathbf{H}$\\ \hline 
    (I)&(a) & -1 & 0 & 0 & 0 \\ \hline
    (III)&(b)& 0 & 0 & -1/4 & -1/16 \\ \hline
    (IV)&(c) & 0 & 0 & 0 & 0 \\
    \bottomrule
    \end{tabular}
    \caption{Discrete approximations of second-order image derivatives and the determinant of the Hessian matrix calculated for the center pixel of the $3 \times 3$ patterns (a), (b), and (c) shown in \cref{fig:all_patterns_comparison}. 
     }\label{tab:hessian_values}
\end{table}
Standard finite-difference-based Hessian approximations often exhibit anisotropic limitations: these methods tend to preferentially preserve edge features aligned with coordinate axes while suppressing those that are obliquely oriented. This axial bias introduces artifacts in curvature-driven models, typically manifesting as staircasing effects or unnatural smoothing along preferred directions. As demonstrated experimentally; see \cref{Effect of beta}(b), these issues become particularly pronounced when the curvature term dominates the energy functional, such as in our proposed model with only TNC regularization activated (\(\beta=0\)). Therefore, it is generally necessary to incorporate TV regularization terms \cite{liu2022operator}, which can effectively mitigate the anisotropic issues caused by finite-difference-based Hessian approximation methods.

\section*{Acknowledgments}
The work was supported by the National Natural Science Foundation of China NSFC 12071345. The authors would also like to thank the anonymous reviewers for their valuable suggestions and comments.

\bibliographystyle{siamplain}
\bibliography{references}
\end{document}